\documentclass{article}

\usepackage[preprint]{neurips_2026}

\usepackage[utf8]{inputenc} 
\usepackage[T1]{fontenc}    
\usepackage{hyperref}       
\usepackage{url}            
\usepackage{booktabs}       
\usepackage{amsfonts}       
\usepackage{nicefrac}       
\usepackage{microtype}      
\usepackage{xcolor}         
\usepackage{amsmath}
\usepackage{minted}
\usepackage{caption}
\usepackage{titletoc}
\usepackage{graphicx}
\usepackage{subcaption}
\titlecontents{section}
  [3em]                    
  {}                       
  {\contentslabel{2.3em}}  
  {}                       
  {\titlerule*[0.5pc]{.}\contentspage} 
\usepackage{wrapfig}
\usepackage{multirow}
\usepackage{enumitem}

\title{Large Language Models Lack Temporal Awareness \\of Medical Knowledge}

\author{%
  Zihan Guan$^{*,1}$, Qiao Jin$^{*,2}$, Guangzhi Xiong$^{1}$, Fangyuan Chen$^{3}$, Mengxuan Hu$^{1}$, \\
  \textbf{Qingyu Chen$^{4}$, Yifan Peng$^{5}$, Zhiyong Lu$^{\dag, 2}$, Anil Vullikanti$^{\dag, 1}$}  \\
  $^1$University of Virginia, $^2$National Institutes of Health \\
  $^3$Dana-Farber Cancer Institute, $^{4}$Yale University\\
  $^{5}$Weill Cornell Medicine \\
  $^*$ Equal-contribution, $^\dag$ Co-correspondence
}

\begin{document}

\maketitle

\begin{abstract}
The existing methods for evaluating the medical knowledge of Large Language Models (LLMs) are largely based on atemporal examination-style benchmarks, while in reality, medical knowledge is inherently dynamic and continuously evolves as new evidence emerges and treatments are approved. Consequently, evaluating medical knowledge without a temporal context may provide an incomplete assessment of whether LLMs can accurately reason about time-specific medical knowledge. Moreover, most medical data are historical, requiring the models not only to recall the correct knowledge, but also to know \textit{when} that knowledge is correct. To bridge the gap, we built TempoMed-Bench, the first-of-its-kind benchmark for evaluating the \textbf{temporal awareness} of the LLMs in the medical domain through evolving guideline knowledge. Based on the TempoMed-Bench, our evaluation analysis first reveals that LLMs lack temporal awareness in medical knowledge through the key findings: (1) model performance on up-to-date medical knowledge exhibits a gradual linear decline over time rather than a sharp knowledge-cutoff behavior, suggesting that parametric medical knowledge is not strictly bounded by knowledge cutoffs; (2) LLMs consistently struggle more with recalling outdated historical medical knowledge than with up-to-date recommendations: accuracy of historical knowledge is only 25.37\% - 53.89\% of up-to-date knowledge, indicating potential knowledge forgetting effects during training; and (3) LLMs often exhibit temporally inconsistent behaviors, where predictions fluctuate irregularly across neighboring years. We also show that the temporal awareness problem is a challenge that cannot be easily solved when integrated with agentic search tools (-3.15\% - 14.14\%).
This work highlights an important yet underexplored challenge and motivates future research on developing LLMs that can better encode time-specific medical knowledge\footnote{The code and data are available at \url{https://github.com/GuanZihan/TempoMed-Bench}}.
  
\end{abstract}

\section{Introduction}






Large language models (LLMs) have demonstrated remarkable performance in various medical tasks, including clinical question answering, diagnosis, and specialty-specific knowledge asssessment~\cite{barile2024diagnostic, rydzewski2024comparative, mihalache2023performance}. These capabilities are often due to their strong ability to encode large amounts of medical knowledge during pre-training and post-training~\cite{singhal2023large}. Because clinical applications are high-stakes, the LLMs' internal medical knowledge must be evaluated rigorously. 

\begin{wrapfigure}[14]{r}{6cm}
  \centering
  \includegraphics[width=\linewidth]{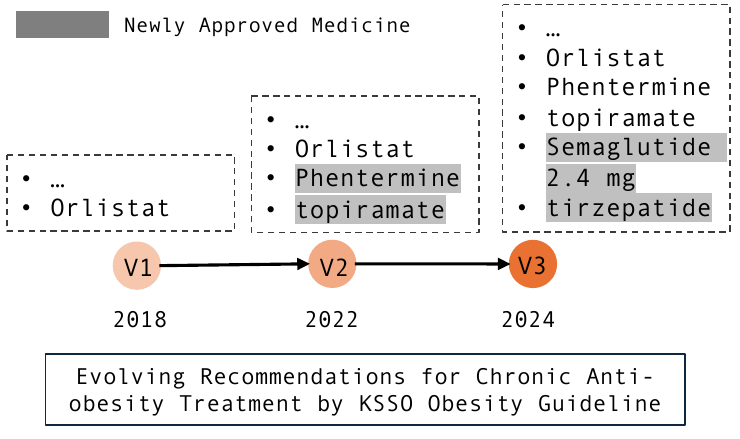}
  \caption{An example of the evolving medical knowledge. 
  } \label{fig:example_evolving_example_intro}
  \vspace{-2mm}
\end{wrapfigure}

Most existing evaluations, however, assess medical knowledge in an atemporal way. Medical Benchmarks~\cite{jin2021disease,pal2022medmcqa,zuo2025medxpertqa} typically test factual and reasoning ability using exam-style questions, but they overlook a critical dimension: \textbf{temporal validity}. Real-world medical knowledge is inherently dynamic and continuously evolves with new evidence, guidelines, and treatment approvals. For example, obesity treatment guidelines issued by the Korean Society for the Study of Obesity have changed over time as new anti-obesity medications became available~\cite{seo20192018, kim2023evaluation, bae20252024}. A treatment that was appropriate according to a 2018 guideline may no longer be recommended in a 2024 guideline (Figure~\ref{fig:example_evolving_example_intro}). Conversely, a current recommendation may not have been valid in an earlier clinical context. As a result, atemporal benchmarks may be insufficient to assess whether LLMs can reason about time-specific knowledge.

To address this limitation, we study a crucial yet underexplored problem through the lens of \textbf{temporal awareness}: the ability of LLMs to correctly recall and align medical knowledge with a specified point in time.
Temporal awareness is strongly grounded in real-world clinical practice: Much of medical data is inherently historical and longitudinal~\cite{goodwin2022timing, cui2025timer}, as they are recorded from past patient encounters and clinical trials. This requires the models not only to recall the correct knowledge, but also to know \textit{when} that knowledge is correct. 
This need arises in multiple scenarios. First, users may ask LLMs health-related questions whose correct answers depend on recently updated evidence. Second, clinicians and researchers may need to retrospectively interpret prior studies, historical recommendations, or past treatment decisions. 
In both cases, the model must distinguish between current and outdated knowledge.




To evaluate LLMs' temporal awareness in the medical domain, we introduce TempoMed-Bench, a benchmark built from evolving clinical guidelines. Guideline knowledge is continuously released and updated by official organizations, and can therefore be naturally  seen as snapshots of evolving medical knowledge. We collect over 20,000 guideline articles from the PubMed Central (PMC) database and organize them into 3,411 trajectories of guideline versions issued by the same organization on the same topic. For each trajectory, we extract differences between the up-to-date version and each of the outdated version, yielding 721 version-to-version difference pairs. Finally, for each pair, we construct a multiple-choice question (MCQ) to evaluate whether LLMs' knowledge aligns more closely with the current or outdated guidelines.

TempoMed-Bench aims to address four research questions. \textbf{RQ1}: Have LLMs already acquired up-to-date medical knowledge? \textbf{RQ2}: Can LLMs correctly recall historical (outdated) medical knowledge, and how consistent are their behaviors over time?
\textbf{RQ3}: Can LLMs maintain temporal consistency across different time periods? 
\textbf{RQ4}: Can agentic RAG systems mitigate these failures? 

We evaluate more than 10 frontier LLMs on TempoMed-Bench. Our evaluation results first reveal that LLMs lack temporal awareness in medical knowledge, which is supported by the key findings below.
In RQ1, we empirically observe that, across different guideline trajectories, strong models such as GPT-5 are not oracles in recalling all detailed up-to-date knowledge (Table~\ref{tab:current_targeted_model_distribution}); moreover, model performance on up-to-date medical knowledge exhibits a gradual linear decline over time (Figure~\ref{fig:year_grouped_accuracy}), rather than a sharp knowledge-cutoff behavior. In RQ2, we found that, in a given guideline trajectory, LLMs consistently struggle more with recalling historical medical knowledge than with up-to-date recommendations: accuracy of historical knowledge is only 25.37\% - 53.89\% of up-to-date knowledge (Figure~\ref{fig:compare_accuracy}), indicating potential knowledge forgetting effects during training. 
In RQ3, we show that LLMs exhibit temporally inconsistent behaviors, where predictions fluctuate irregularly across neighboring years without coherent temporal transitions. We also show that the temporal awareness problem is a non-trivial challenge that cannot be easily solved when integrated with popular agentic search tools, such as ToolUniverse~\cite{gao2025democratizing} and Biomni~\cite{huang2025biomni}, which only bring limited improvements (-3.15\% - 14.14\%; see Table~\ref{tab:accuracy_agentic_rag}).






To sum up, our contributions are:
(1) \textbf{Temporal-Aware Medical Knowledge Evaluation}. We propose temporal context as a necessary dimension for evaluating medical LLMs, motivated by the inherently evolving nature of real-world medical knowledge.
(2) \textbf{TempoMed-Bench}. We construct a large-scale collection of evolving clinical guideline trajectories and a benchmark of 721 temporally grounded medical MCQs.
(3) \textbf{Empirical Evidence of Limited Temporal Awareness in LLMs}. We empirically show that current LLMs lack temporal awareness of medical knowledge and that current agentic RAG systems only partially address the problem, through four research questions.


\section{Related Works}
\textbf{Medical Knowledge Evaluation.} Prior work on medical knowledge evaluation has largely concentrated on atemporal QA benchmarks~\cite{jin2022biomedical}, including MedQA~\cite{jin2021disease}, PubMedQA~\cite{jin2019pubmedqa}, and MedMCQA~\cite{pal2022medmcqa}, as well as broader benchmark collections such as MultiMedQA~\cite{singhal2023large} and MedHELM~\cite{bedi2026holistic}. More recent efforts have extended evaluation to explanation quality, clinical versatility, safety, and guideline adherence, as exemplified by MedExQA~\cite{kim2024medexqa}, MedS-Bench~\cite{wu2025towards}, and HealthBench~\cite{arora2025healthbench}. However, these evaluations still mostly assume that medical knowledge is static. In contrast, medical knowledge is inherently evolving, with clinically important changes frequently formalized through updated guidelines.

\textbf{Temporal Knowledge Evaluation.} Most knowledge in the real world evolves over time. Temporal awareness has recently been a popular dimension in evaluating the evolving knowledge of LLMs~\cite{lin2025temporal, wang2024large, chen2024temporal, zhu2025evolvebench, tang2025evowiki, nakshatri2025facts, mousavi2024dyknow}. However, the existing works only focus on the general knowledge (e.g., ``who served as the president of the United States from 2001 to 2009?'') from sources such as Wikidata~\cite{tang2025evowiki, zhu2025evolvebench, mousavi2024dyknow, chen2024temporal} or industry documents~\cite{nakshatri2025facts}. In this paper, we particularly focus on the temporal knowledge evaluation in the medical domain. The problem remains largely unexplored, mainly due to the intrinsic challenges of collecting and curating temporally evolving medical knowledge, which is often unstructured and distributed across various clinical trial reports and guideline papers. ~\cite{wu2025assessing} focuses on evaluating whether LLMs can endorse current and reject pseudo-outdated medical recommendations when some synthesized conflicting information is present in the context. Although we share a similar setting in evaluations with medical guidelines, the differences are also significant: (1) we focus on evaluating LLMs' temporal awareness in the medical domains, which is fundamentally different; (2) their outdated knowledge is constructed using manually created pseudo-outdated versions, while ours focuses on the real evolution of medical guidelines.

\section{TempoMed-Bench}

In this section, we introduce TempoMed-Bench, a benchmark for evaluating how LLMs handle evolving medical knowledge. Specifically, we construct a dataset grounded in clinical guideline documents, which naturally capture snapshots of medical knowledge at different points in time and are regularly revised to reflect new evidence. Based on these guidelines, we identify differences across versions and construct multiple-choice questions (MCQs) that explicitly probe models’ preferences between up-to-date and outdated medical knowledge.

In the following sections, we describe the dataset construction pipeline in detail and provide a comprehensive statistical analysis of the resulting benchmark.

\subsection{Dataset Preparation Pipeline}

\textbf{Guideline Collection \&  Filtering.} We collect the guideline papers from the PubMed Central (PMC)\footnote{\url{https://pmc.ncbi.nlm.nih.gov/}}  database, which contains over 7.5 million life science and biomedical papers of various types, including reviews, clinical studies, guidelines, and research articles. In this step, we aim to build a rough classifier to minimize the risk of missing guideline papers. To this end, we adopt a relaxed filtering strategy that combines metadata from PubMed with na\"ive keyword matching. Specifically, for each article, if the term ``guideline'' appears in the title or in the \texttt{<article-type>} tag, we classify it as a guideline paper. Figure~\ref{fig:pipeline_step1} visualizes the process.

The articles in the PMC database are organized into three subsets: \texttt{comm}, \texttt{noncomm}, and \texttt{other}. Using the above filtering process, we extract 13{,}002 papers from the \texttt{comm} subset, 8{,}790 papers from the \texttt{noncomm} subset, and 1{,}403 papers from the \texttt{other} subset, resulting in a total of 23{,}195 papers.

\begin{figure}
  \centering
  \includegraphics[width=\linewidth]{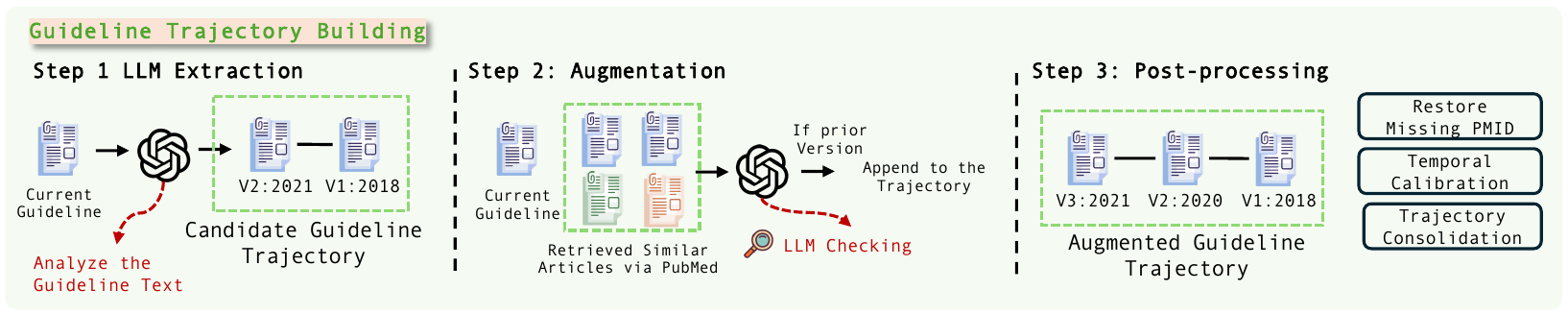}
  \caption{Overview of the guideline trajectory building pipeline. Step 1 uses an LLM to extract candidate prior versions from the current guideline. Step 2 augments the trajectory via retrieved similar articles and LLM-based verification. Step 3 refines the trajectory through post-processing.}\label{fig:pipeline_step2}
  \vspace{-5mm}
\end{figure}

\textbf{Guideline Trajectory Building.} The above process introduces many noisy papers that might not actually be a guideline paper. For example, the paper ``Tuberculosis in childhood: a systematic review of national and international guidelines''~\cite{berti2014tuberculosis} is classified as a guideline paper simply because `guideline' appears in the title. Therefore, a more fine-grained filter is required.

We adopt an LLM to analyze the article texts and determine whether the given article is a guideline paper or not, where a guideline paper is defined as a guideline issued by a professional society (e.g., the American College of Rheumatology). If the guideline checking is passed, then the LLM proceeds to analyze whether the guideline paper has cited any prior versions of the current guideline paper, where the prior guideline is formally defined as an official guidance document issued by the \textbf{same organization} before the year of the current guideline and focuses on the \textbf{same topic} as the current guideline. The overall prompt for the LLM is present in Listing~\ref{prompt:guideline_traj}. This process enables us to extract at least one prior guideline for 1{,}134 papers in the \texttt{comm} subset, 1{,}586 papers in the \texttt{noncomm} subset, and 200 papers in the \texttt{other} subset, resulting in a total of 2{,}920 guideline trajectories. We name these trajectories as candidate guideline trajectories.

We notice that the above candidate trajectory might not be complete either because the current guideline may not mention all the prior guidelines, or the LLMs might omit some by hallucination. Therefore, we apply necessary post-processing steps to revise and augment the trajectory data (with details shown in Appendix~\ref{appendix:post-processing}). After the post-processing, there is at least one prior guideline for 1,496 papers in the \texttt{comm} subset, 1,691 papers in the \texttt{noncomm} subset, and 224 papers in the \texttt{other} subset, resulting in a total of 3,411 guideline trajectories. We name these trajectories as \textbf{TempoMed-Traj}. The process is visualized in Figure~\ref{fig:pipeline_step2}.

\begin{wrapfigure}{r}{8cm}
  \centering
  \includegraphics[width=\linewidth]{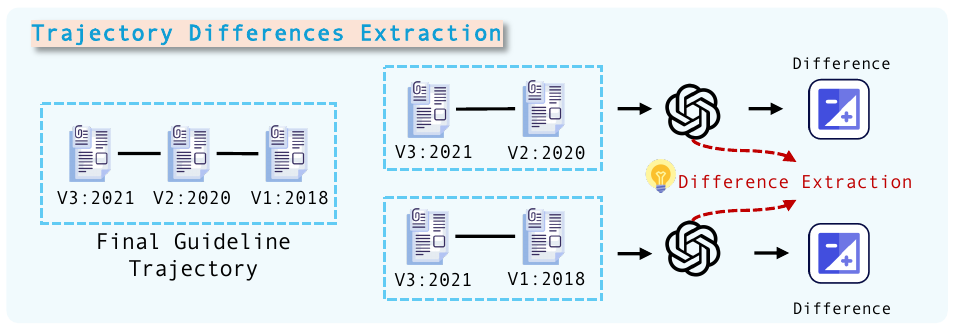}
  \caption{Extracting differences from a pair of up-to-date and outdated guidelines.} \label{fig:pipeline_step3}
  \vspace{-2mm}
\end{wrapfigure}

\textbf{Trajectory Difference Extraction.} Given the guideline trajectories, the next step is to extract the evolving changes between the up-to-date guideline and each of its previous outdated versions. To achieve this, we prompt LLMs to identify these differences, and the extracted changes are subsequently verified by a separate, strict verifier LLM. 
Listing~\ref{prompt:extract_difference} and Listing~\ref{prompt:verifier} present the prompt to the LLM. As seen, the extracted difference pair between each up-to-date guideline and outdated guideline must represent a direct head-to-head comparison and be meaningfully different. Then, a verifier LLM scores the extracted difference pair with detailed rubrics. 
Note that in this step, we only consider the trajectory where all the guideline nodes are in the guideline set we prepared in \{Guideline Collection \& Filtering\}, since otherwise we would not be guaranteed to get the full text of the paper. The process is visualized in Figure~\ref{fig:pipeline_step3}. We provide several examples of the extracted differences in the Appendix~\ref{appendix:extracted_differences_example}. 
This process results in 721 difference pairs; each is an extracted difference between an up-to-date guideline and an outdated guideline.



\textbf{MCQ Generation.} For each of the extracted differences, we construct a corresponding evaluation question that aims to test whether the LLMs' knowledge is closer to the up-to-date knowledge or to the outdated knowledge. To align with the traditional exam style, we design QAs according to the NBME (National Board of Medical Examiners) guideline\footnote{\url{https://www.nbme.org/sites/default/files/2021-02/NBME\_Item\%20Writing\%20Guide\_R_6.pdf}}. 
Specifically, for each extracted difference, we prompt the LLM as in Listing~\ref{prompt:qa_generation}, where the LLM is expected to generate the NBME-style question stem and option set for each QA instance. The option set contains five options, which contain an option that aligns with the knowledge in the up-to-date guideline, an option that aligns with the knowledge in the outdated guideline, and two clinically plausible distractor options that align with neither of the two. We also added a `I don't know' (IDK) option to investigate the LLM's abstention ability following~\cite{madhusudhan2025llms}. The process is visualized in Figure~\ref{fig:pipeline_step4}. We provide several examples of the constructed questions in the Appendix~\ref{appendix:mcqs}. This generates 721 MCQs in total, which we name as \textbf{TempoMed-Bench}.

\begin{wrapfigure}{r}{10cm}
  \centering
  \vspace{-5mm}
  \includegraphics[width=\linewidth]{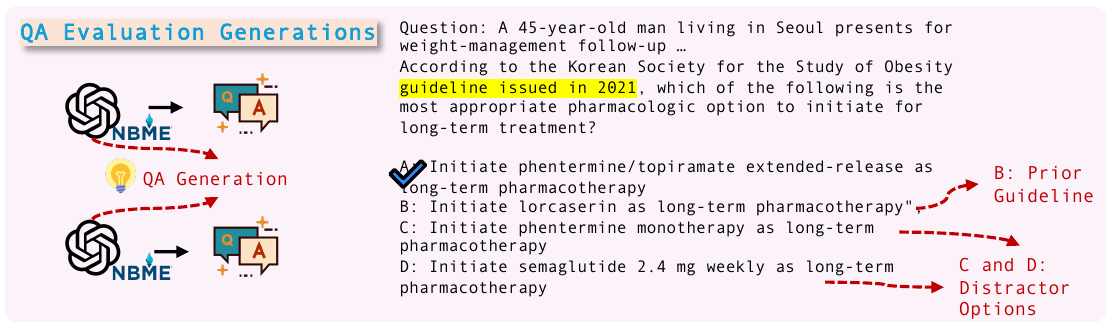}
  \caption{Constructing MCQs from extracted differences.} \label{fig:pipeline_step4}
  \vspace{-2mm}
\end{wrapfigure}


\begin{figure*}[!t]
      \centering
      \begin{subfigure}[t]{0.32\textwidth}
          \centering
          \includegraphics[width=\linewidth]{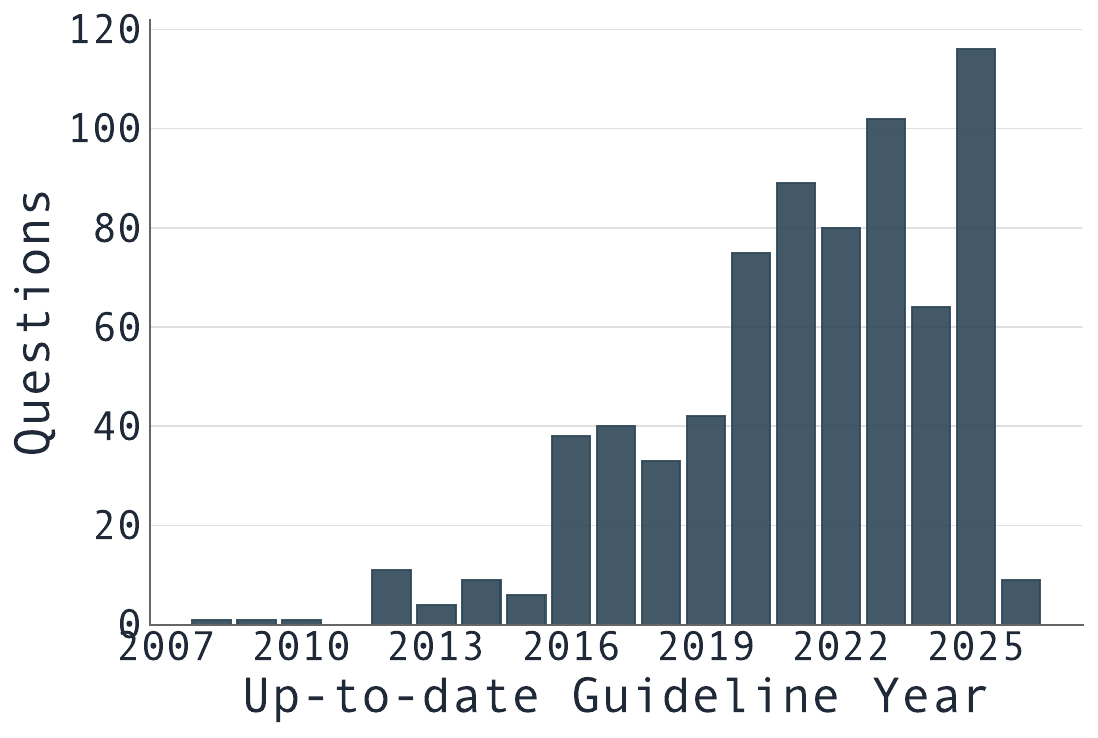}
          \caption{Up-to-date guideline year.}
          \label{fig:current_year_distribution}
      \end{subfigure}
      \hfill
      \begin{subfigure}[t]{0.32\textwidth}
          \centering
          \includegraphics[width=\linewidth]{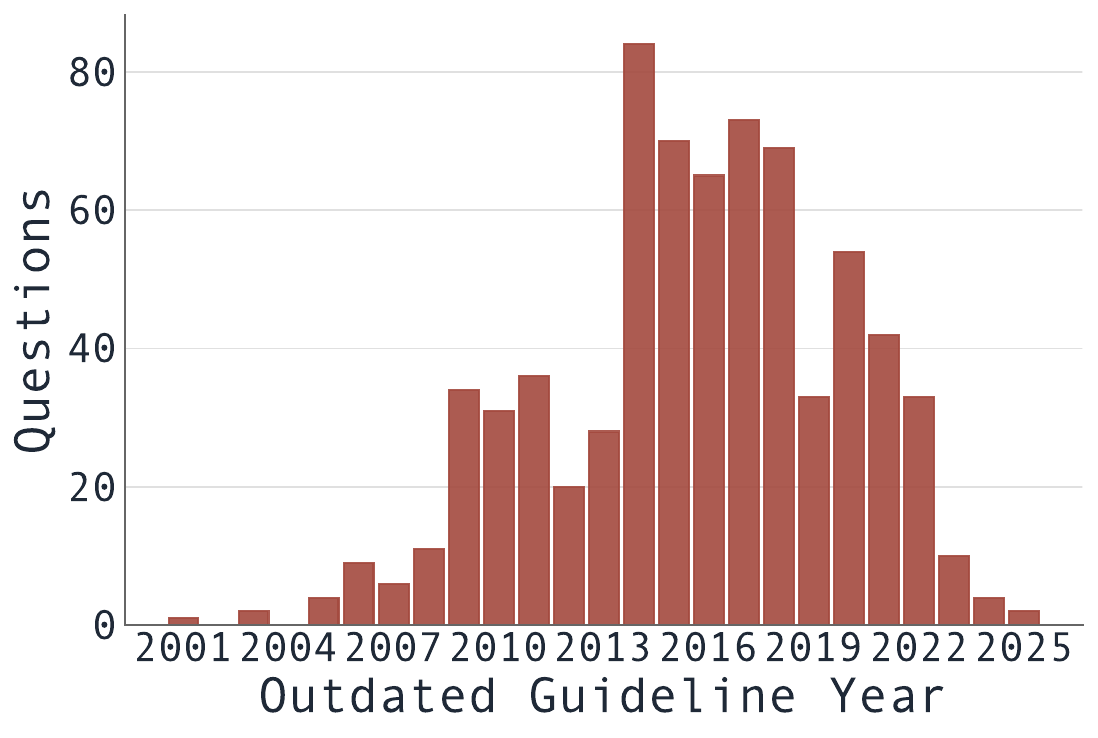}
          \caption{Outdated guideline year.}
          \label{fig:prior_year_distribution}
      \end{subfigure}
      \hfill
      \begin{subfigure}[t]{0.32\textwidth}
          \centering
          \includegraphics[width=\linewidth]{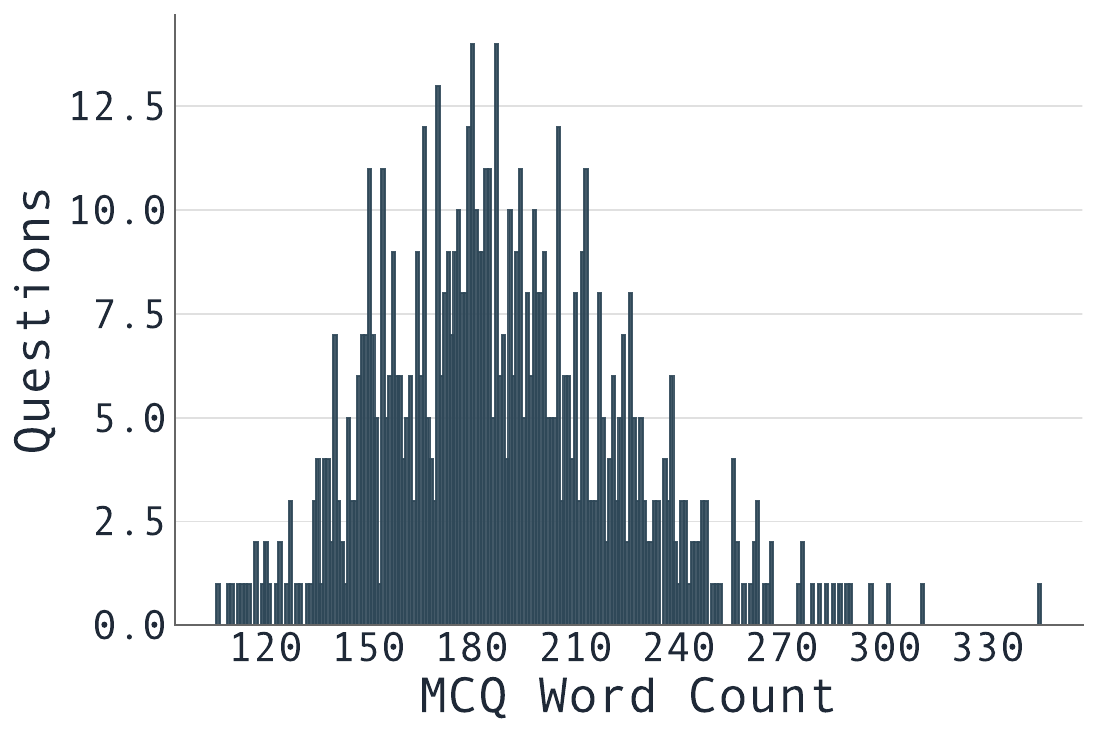}
          \caption{\# words in the Questions.} \label{fig:question_word_count_distribution}
      \end{subfigure}
      \caption{Statistical analysis of the MCQs in the TempoMed-Bench. Distribution of (a) up-to-date guideline years, (b) outdated guideline years, and (c) \# words in the MCQs.}
      \label{fig:mcq_analysis}
      \vspace{-5mm}
\end{figure*}

\subsection{Dataset Statistical Analysis}
\textbf{Analysis of TempoMed-Traj.} We provide visualizations for the following statistical analysis of the final post-processed trajectories. Figures~\ref{fig:trajectory_length}, ~\ref{fig:trajectory_timespan}, and ~\ref{fig:current_guideline_year} characterize the trajectory dataset from structural and temporal perspectives across the \texttt{comm}, \texttt{noncomm}, and \texttt{other} subsets. Figure~\ref{fig:trajectory_length} presents the distribution of trajectory lengths, measured by the number of guideline nodes in each trajectory, with single-node trajectories excluded. The figure implies that many of the trajectories contain only one prior version. Figure~\ref{fig:trajectory_timespan} presents the distribution of trajectory spans, defined as the difference between the earliest and latest guideline years in each trajectory. The figure indicates that most of the guideline year span is less than 10 years. Figure~\ref{fig:current_guideline_year} presents the distribution of current guideline years, capturing the recency profile of the most recent guideline in each trajectory. This shows that the up-to-date guideline year of the most extracted guideline trajectories is after 2010.

\textbf{Analysis of the TempoMed-Bench.} The final MCQ dataset comprises 721 multiple-choice questions, each constructed from a difference pair identified between an up-to-date guideline and an outdated guideline. Figure~\ref{fig:current_year_distribution} and Figure~\ref{fig:prior_year_distribution} illustrate the distributions of the up-to-date and outdated guideline years, respectively. The dataset spans updated guideline years from 2008 to 2026 and outdated guideline years from 2001 to 2025, covering a broad temporal range of medical knowledge evolution. In addition, Figure~\ref{fig:question_word_count_distribution} shows the distribution of question lengths, which approximates a Gaussian distribution.

\section{Evaluation Results}
Based on the TempoMed-Bench, we evaluate the temporal awareness of the LLMs through four research questions. 
\textbf{RQ1}: To what extent have LLMs acquired up-to-date medical knowledge? \textbf{RQ2}: Can LLMs correctly track and recall historical outdated medical knowledge? Beyond overall performance, we further analyze the temporal consistency of the model behaviors.
\textbf{RQ3}: Can LLMs maintain temporal consistency when recalling medical knowledge across different time periods? Finally, given the emergence of agentic retrieval-augmented generation (RAG) systems that can dynamically access external knowledge, we revisit the above temporal awareness problem under a retrieval-enhanced setting:
\textbf{RQ4}: How do agentic RAG systems affect LLMs’ ability to access up-to-date and time-specific medical knowledge?

\subsection{RQ1: Have LLMs already acquired up-to-date medical knowledge?}

\begin{table*}[t]
  \centering
  \caption{Evaluation results for different models on the TempoMed-Bench benchmark. Accuracy is computed over all QA instances. All models are instruction-tuned version by default.}
  \label{tab:current_targeted_model_distribution}
  \resizebox{\textwidth}{!}{
  \begin{tabular}{l|c|ccccc}
  \toprule
  Model & Accuracy (\%) & Up-to-date (\%) & Outdated (\%) & Distractor (\%) & Invalid (\%) & Unknown (\%) \\
  \midrule
\includegraphics[width=0.025\textwidth]{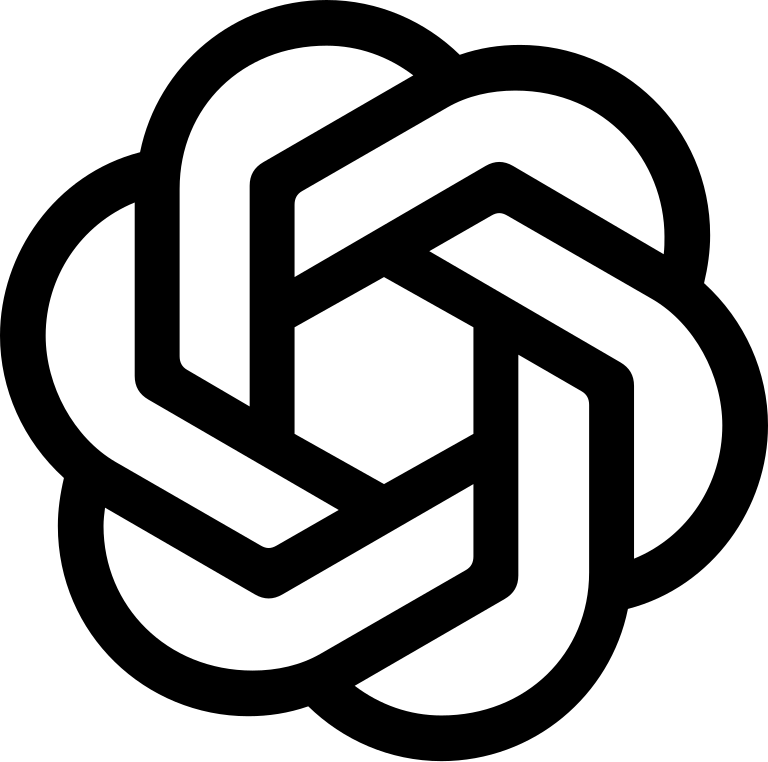} \texttt{GPT-5} & 70.69 & 70.69 & 15.05 & 2.87 & 8.89 & 2.50 \\
\includegraphics[width=0.025\textwidth]{icons/openai_logo.png}  \texttt{GPT-4.1} & 71.11 & 71.11 & 16.85 & 3.06 & 8.98 & 0.00 \\
\includegraphics[width=0.025\textwidth]{icons/openai_logo.png}  \texttt{GPT-4o} & 44.95 & 44.95 & 7.59 & 1.67 & 9.12 & 36.67 \\
\includegraphics[width=0.025\textwidth]{icons/openai_logo.png} \texttt{GPT-OSS-20B} & 54.05 & 54.05 & 18.49 & 6.33 & 6.47 & 14.66 \\
\includegraphics[width=0.025\textwidth]{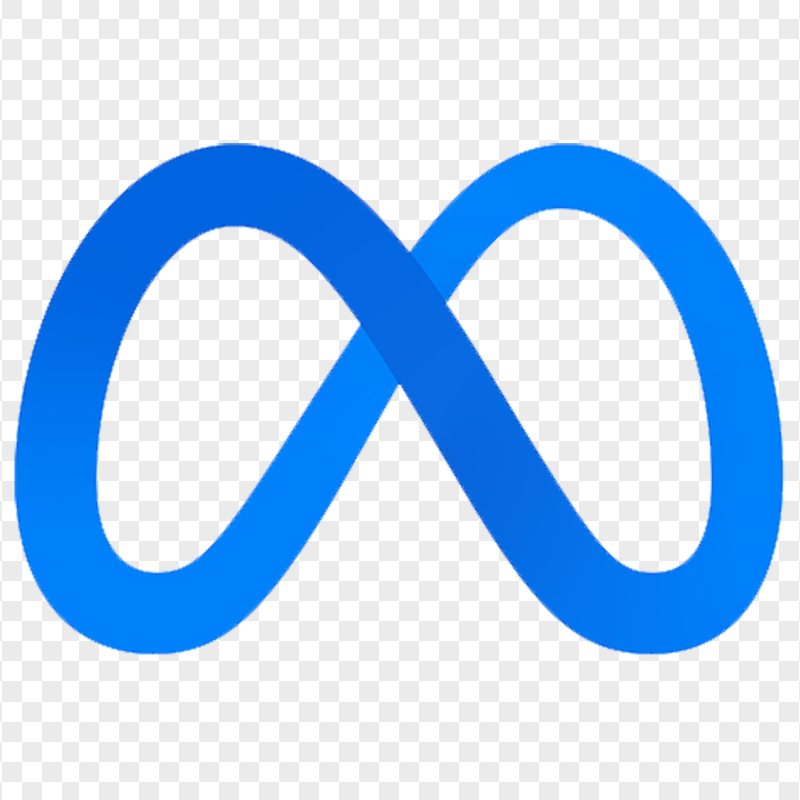} \texttt{Llama-3.1-8B} & 54.92 & 54.92 & 25.29 & 14.01 & 0.18 & 5.59 \\
\includegraphics[width=0.025\textwidth]{icons/meta_logo.png} \texttt{Llama-3.2-3B} & 53.63 & 53.63 & 27.14 & 15.63 & 0.05 & 3.56 \\
\includegraphics[width=0.025\textwidth]{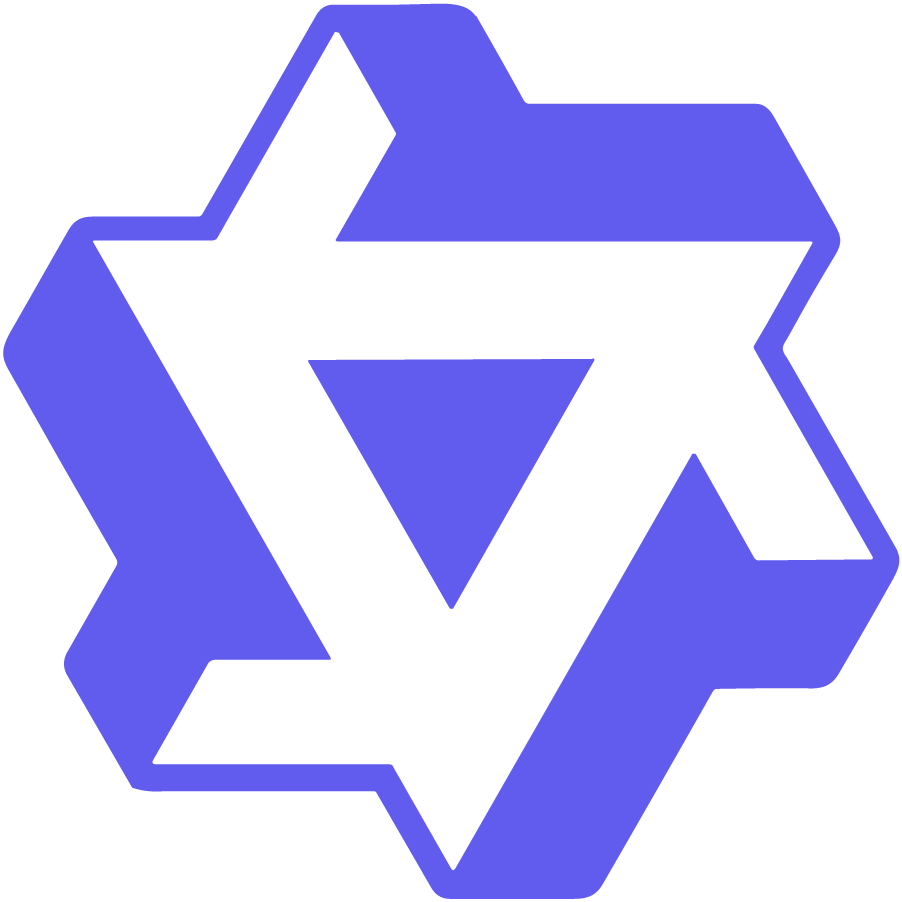} \texttt{Qwen2.5-14B} & 46.32 & 46.32 & 15.86 & 6.20 & 0.00 & 31.62 \\
\includegraphics[width=0.025\textwidth]{icons/qwen_logo.png} \texttt{Qwen2.5-7B} & 59.64 & 59.64 & 23.81 & 16.09 & 0.00 & 0.46 \\
\includegraphics[width=0.025\textwidth]{icons/qwen_logo.png} \texttt{Qwen3-4B} & 46.28 & 46.28 & 17.89 & 10.59 & 0.00 & 25.24 \\

\includegraphics[width=0.025\textwidth]{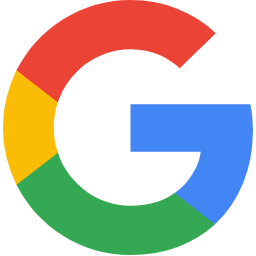} \texttt{MedGemma-27B} & 53.68 & 53.68 & 15.58 & 6.33 & 0.05 & 24.36 \\
\includegraphics[width=0.025\textwidth]{icons/google_logo.png} \texttt{MedGemma-1.5-4B} & 55.43 & 55.43 & 24.27 & 17.85 & 0.92 & 1.53 \\
\includegraphics[width=0.025\textwidth]{icons/google_logo.png} \texttt{MedGemma-4B} & 60.10 & 60.10 & 24.97 & 14.75 & 0.14 & 0.05 \\
\includegraphics[width=0.025\textwidth]{icons/google_logo.png} \texttt{Gemma-3-4B} & 58.16 & 58.16 & 22.65 & 18.82 & 0.00 & 0.37 \\
\includegraphics[width=0.025\textwidth]{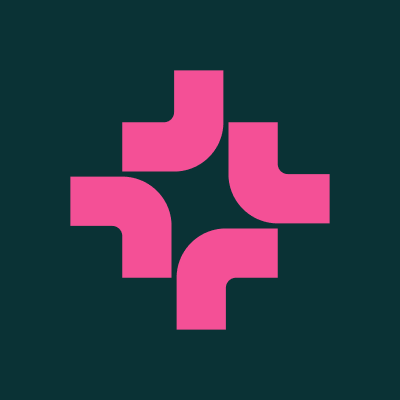} \texttt{OLMo-3-7B} & 57.51 & 57.51 & 22.98 & 14.98 & 0.00 & 4.53\\
  \bottomrule
  \end{tabular}}
\end{table*}

\begin{figure*}[t]
      \centering
      \begin{subfigure}[t]{0.32\textwidth}
          \centering
          \includegraphics[width=\linewidth]{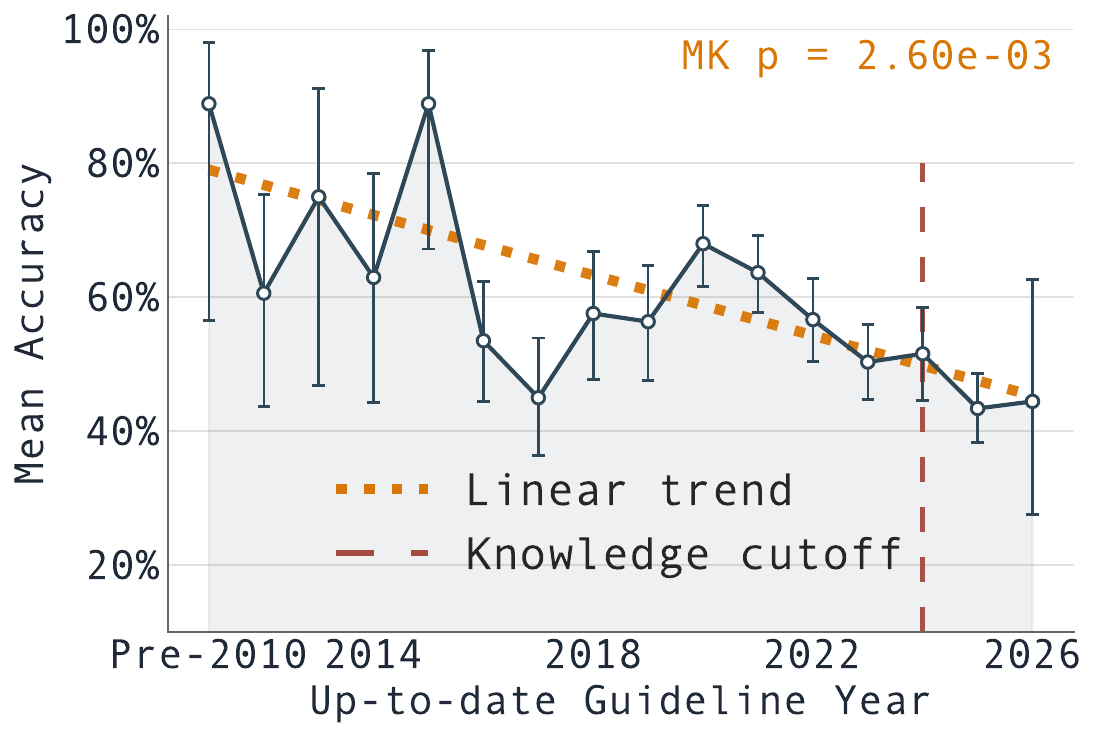}
          \caption{\texttt{Llama-3.1-8B}}
      \end{subfigure}
      \hfill
      \begin{subfigure}[t]{0.32\textwidth}
          \centering
          \includegraphics[width=\linewidth]{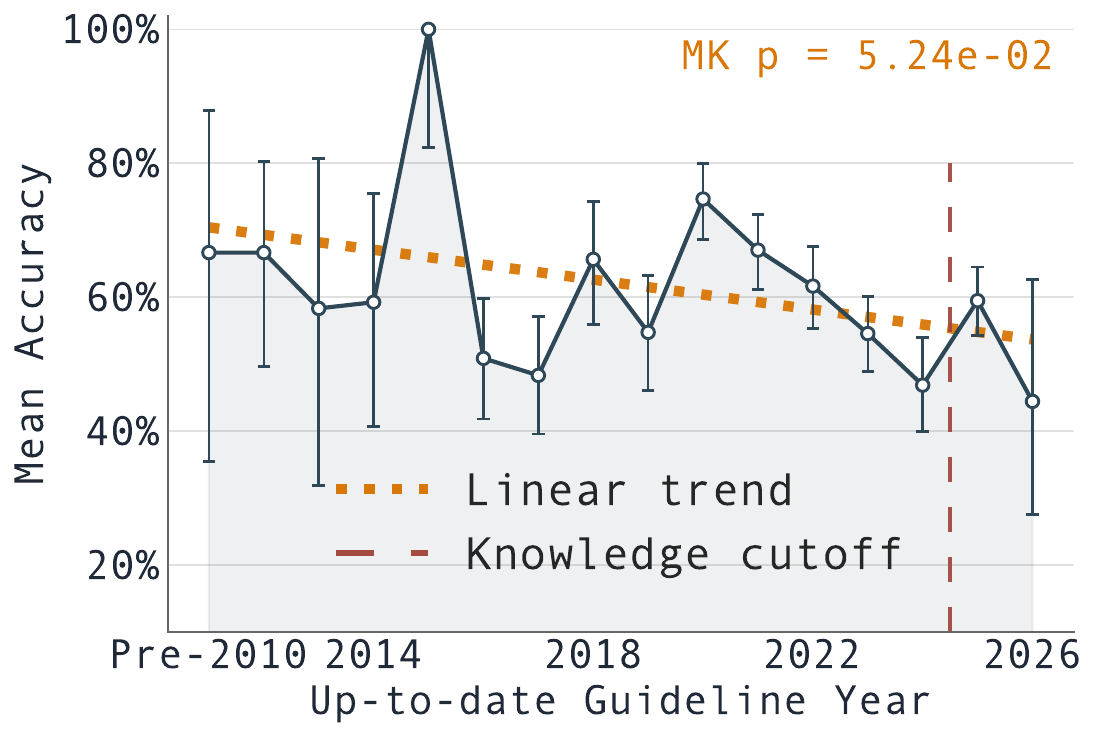}
          \caption{\texttt{Qwen-2.5-14B}}
      \end{subfigure}
      \hfill
      \begin{subfigure}[t]{0.32\textwidth}
          \centering
          \includegraphics[width=\linewidth]{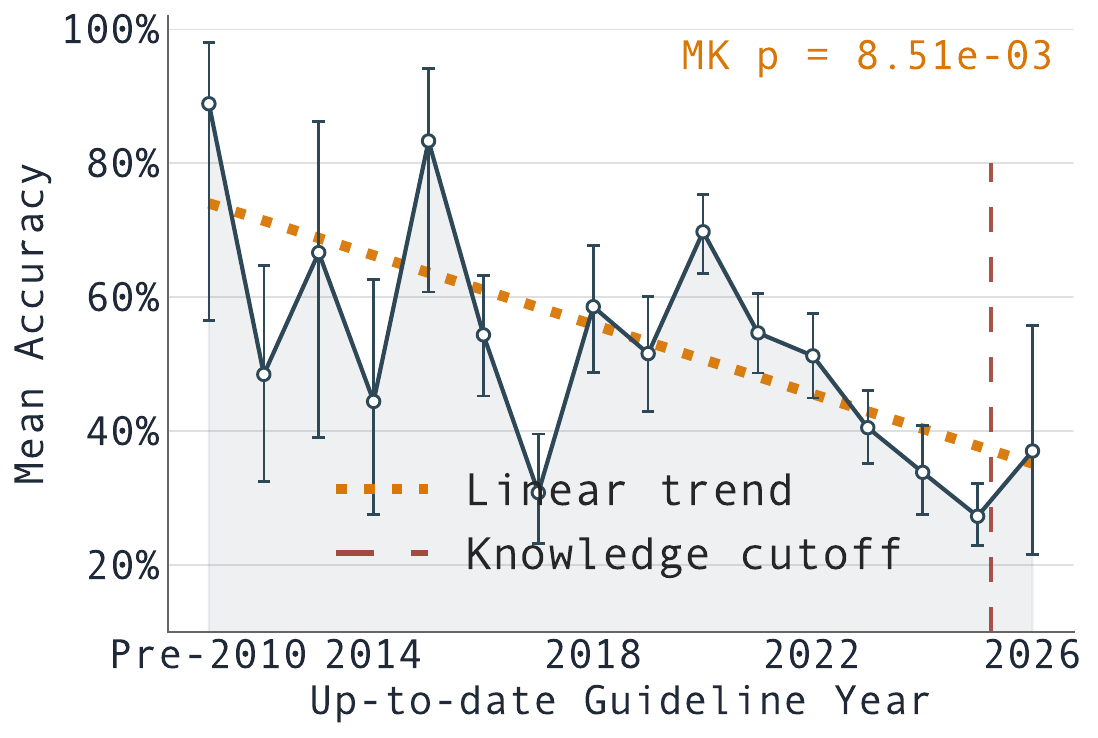}
          \caption{\texttt{Qwen-3-4B}}
      \end{subfigure}
      \hfill
      \begin{subfigure}[t]{0.32\textwidth}
          \centering
          \includegraphics[width=\linewidth]{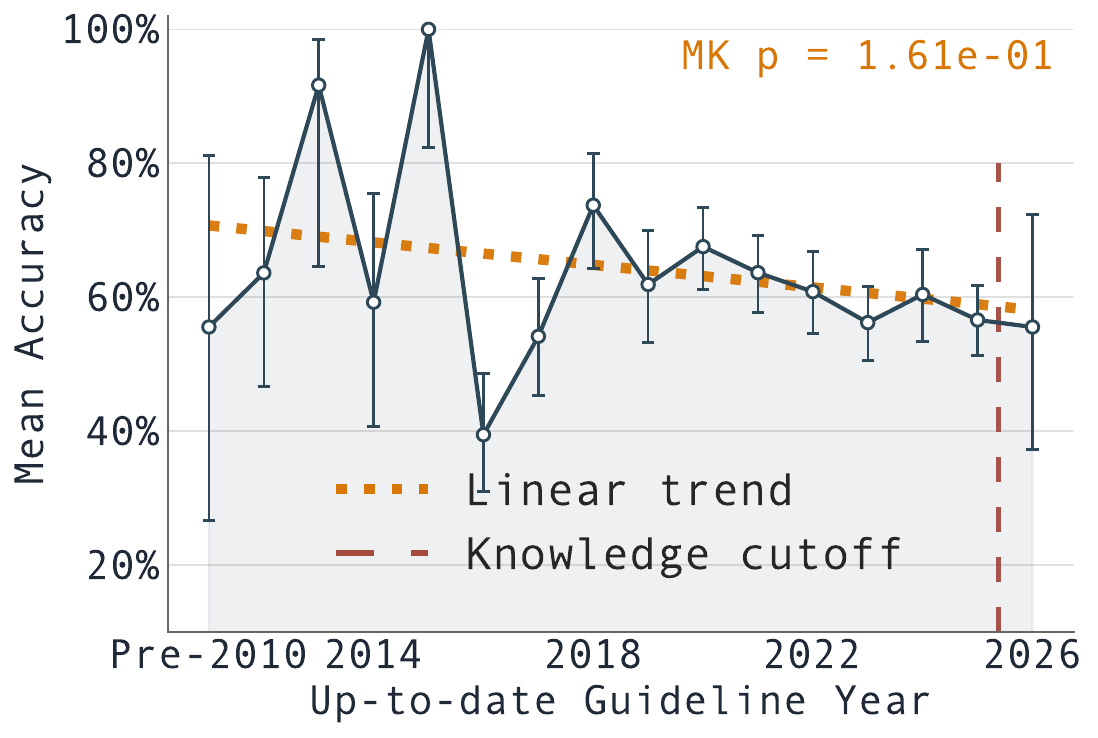}
          \caption{\texttt{Medgemma-4B}}
      \end{subfigure}
      \hfill
      \begin{subfigure}[t]{0.32\textwidth}
          \centering
          \includegraphics[width=\linewidth]{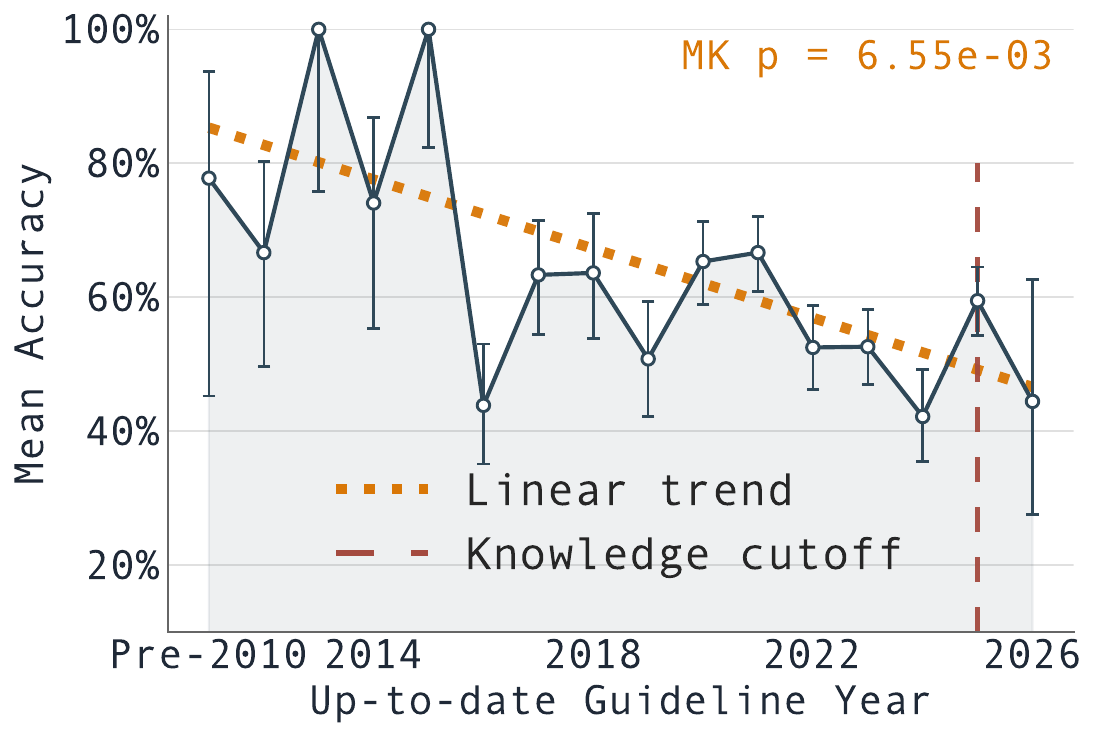}
          \caption{\texttt{Olmo-3-7B}}
      \end{subfigure}
      \hfill
      \begin{subfigure}[t]{0.32\textwidth}
          \centering
          \includegraphics[width=\linewidth]{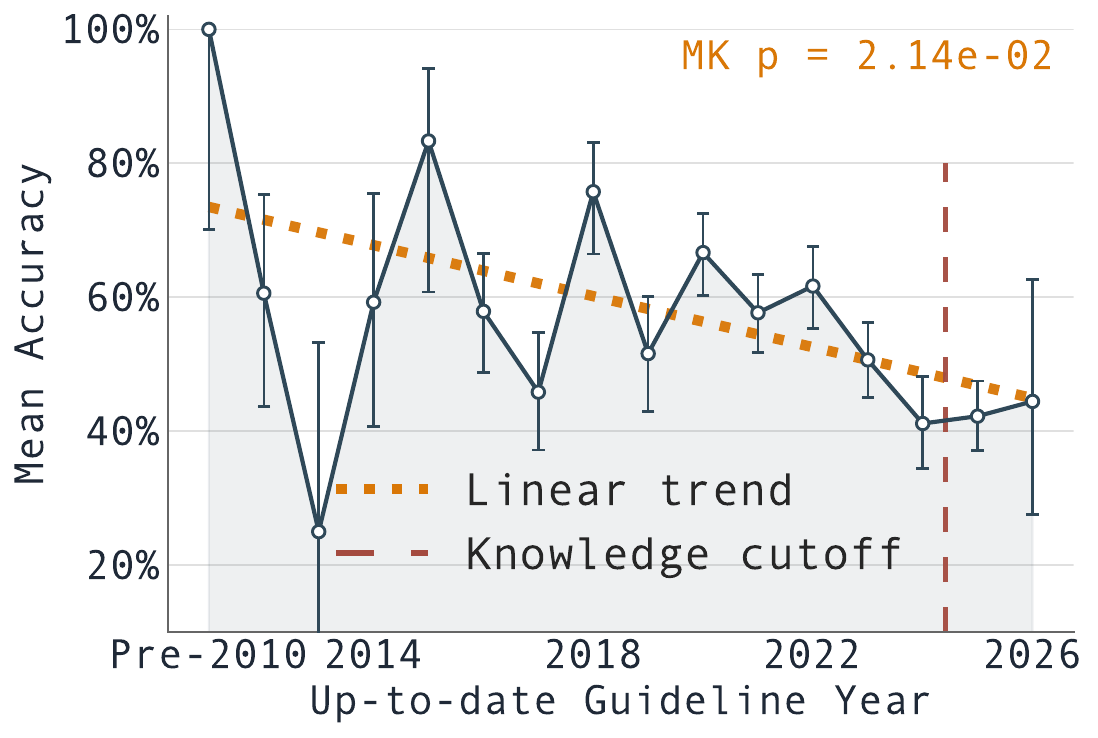}
          \caption{\texttt{GPT-OSS-20B}}
      \end{subfigure}
      \hfill
      \begin{subfigure}[t]{0.32\textwidth}
          \centering
          \includegraphics[width=\linewidth]{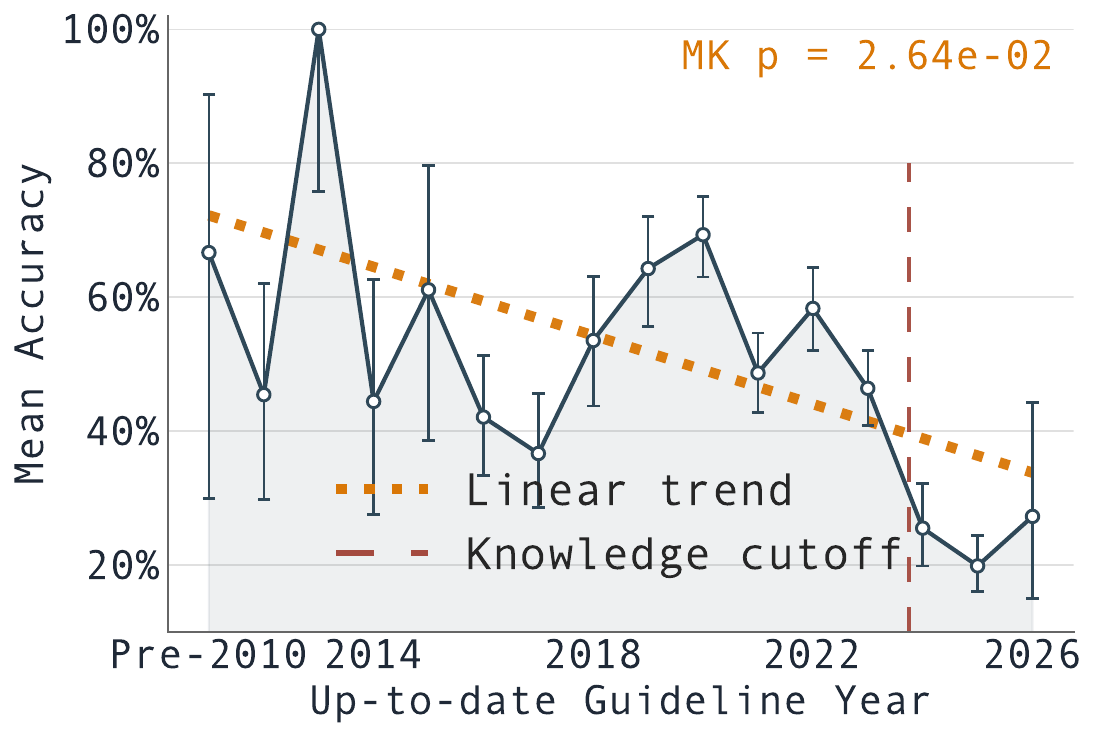}
          \caption{\texttt{GPT-4o}}
      \end{subfigure}
      \hfill
      \begin{subfigure}[t]{0.32\textwidth}
          \centering
          \includegraphics[width=\linewidth]{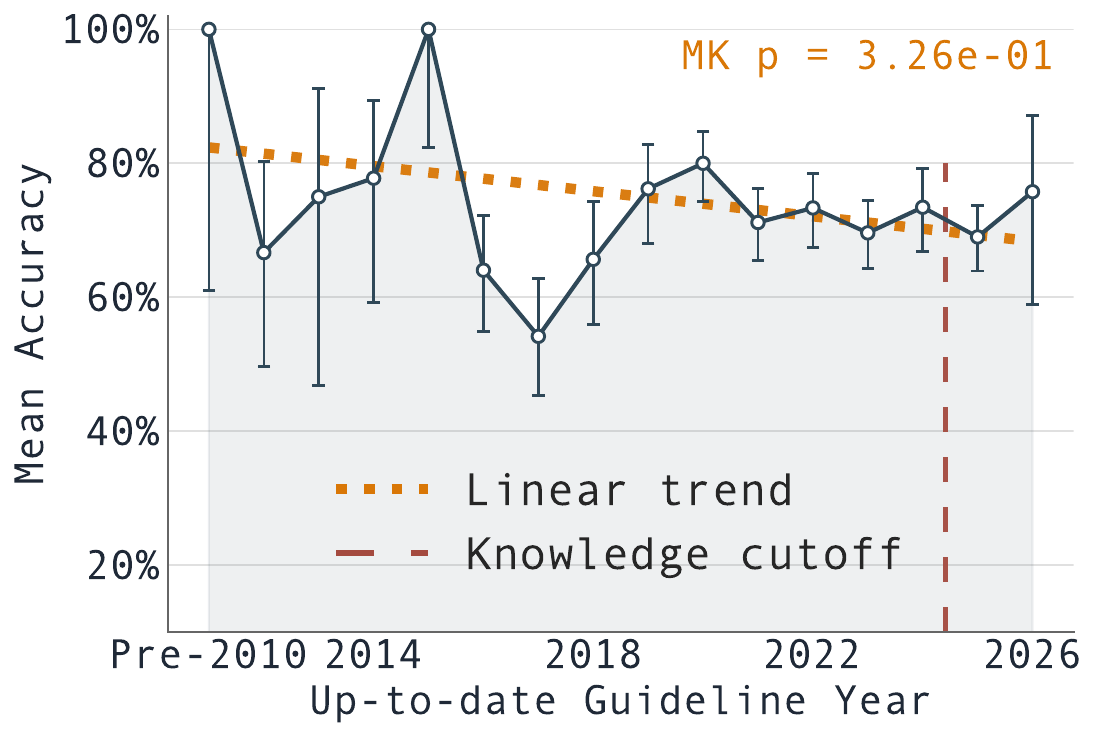}
          \caption{\texttt{GPT-4.1}}
      \end{subfigure}
      \hfill
      \begin{subfigure}[t]{0.32\textwidth}
          \centering
          \includegraphics[width=\linewidth]{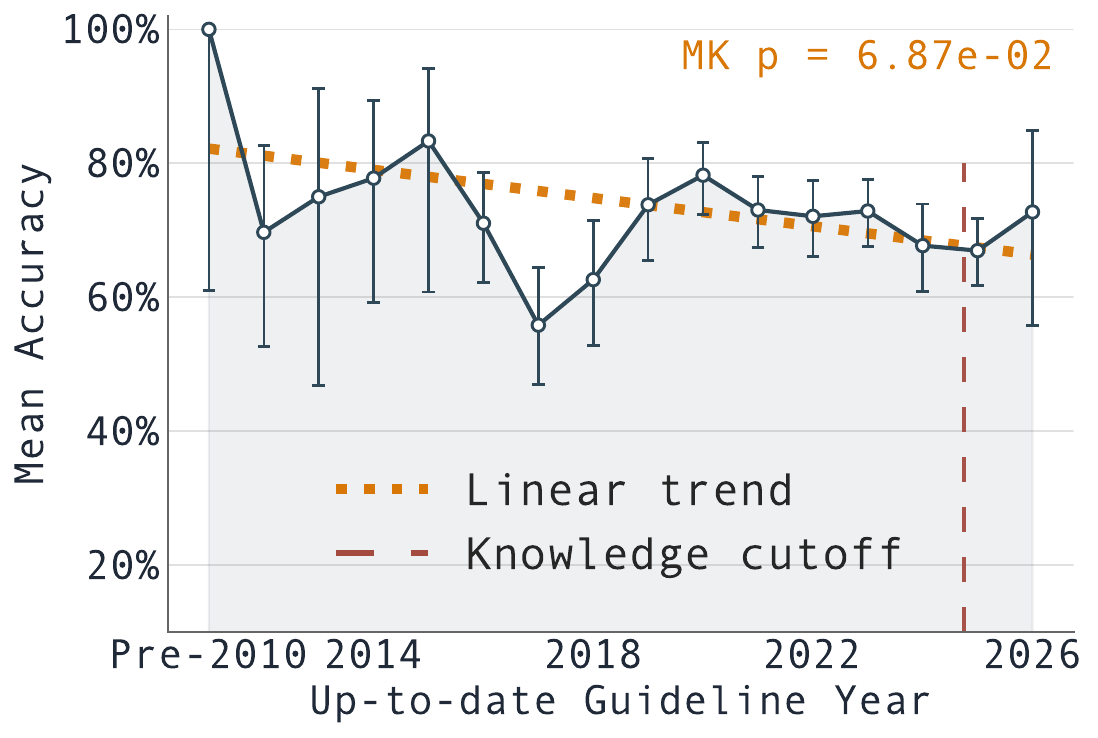}
          \caption{\texttt{GPT-5}}
      \end{subfigure}
      \caption{Average accuracy by the year of the up-to-date guideline. Points denote mean accuracy, and error bars indicate 95\% confidence intervals. The \textcolor{orange}{orange} dotted line shows a linear trend fitted to the mean values. The \textcolor{red!70!black}{red} vertical dashed line indicates the knowledge cutoff date of each model. 
      }
      \label{fig:year_grouped_accuracy}
\end{figure*}

\textbf{Experimental Setup.} To investigate this problem, we evaluate multiple LLMs on each question instance in the TempoMed-Bench. Each question is designed to assess whether a model’s response aligns with the most up-to-date medical guidelines in a trajectory, as opposed to outdated guidelines or distractor options that reflect incorrect or irrelevant recommendations. To ensure a robust evaluation, for each original question, we generate two additional variants by perturbing the positional order of the answer options. This augmentation allows us to reduce the influence of positional bias~\cite{zheng2023large, pezeshkpour2024large} and obtain a more reliable estimate of model performance (see Appendix~\ref{appendix:positional_perturbation} for details).

\textbf{Accuracy \& Prediction Distribution.} Table~\ref{tab:current_targeted_model_distribution} presents the evaluation results for different models on the TempoMed-Bench, where the Accuracy column represents the percentage of correctly answering the option corresponding to the up-to-date knowledge. In general, models with larger parameter size (e.g., \texttt{GPT-5}) exhibit strong encoding of the up-to-date knowledge compared to the smaller models such as \texttt{Qwen3-4B}. Moreover, when zooming in on the option distributions, we found that smaller models (e.g., \texttt{Qwen2.5-7B} and \texttt{Llama-3.1-8B}) tend to choose the distractor options more often than the larger models, partially suggesting that their parametric knowledge might be off the trajectory.

\textbf{Temporal Degradation on Up-to-date Medical Knowledge.} Figure~\ref{fig:year_grouped_accuracy} and~\ref{fig:more_results_current_accuracy} present accuracy grouped by the year of the corresponding up-to-date guideline. As shown, all models tend to capture more recent knowledge \textit{less effectively}, as evidenced by declining accuracy over the years. We further fit a trend line using the average accuracy values across years, and report the $p$-value from the Mann-Kendall test~\cite{mann1945nonparametric}. To analyze the trend, we mark each model’s knowledge cutoff date with a \textcolor{red!70!black}{red} dotted line. Interestingly, the performance degradation follows an approximately linear trend, rather than a sharp drop around the cutoff date. This suggests that the limitation is not solely due to the knowledge cutoff. One plausible explanation is that newer medical knowledge appears less frequently in pre-training corpora, whereas well-established guidelines from earlier years are more widely cited and thus more strongly encoded.

\subsection{RQ2: Can LLMs correctly track and recall historical outdated medical knowledge?}
\begin{figure*}[t]
      \centering
      \begin{subfigure}[t]{0.23\textwidth}
          \centering
          \includegraphics[width=\linewidth]{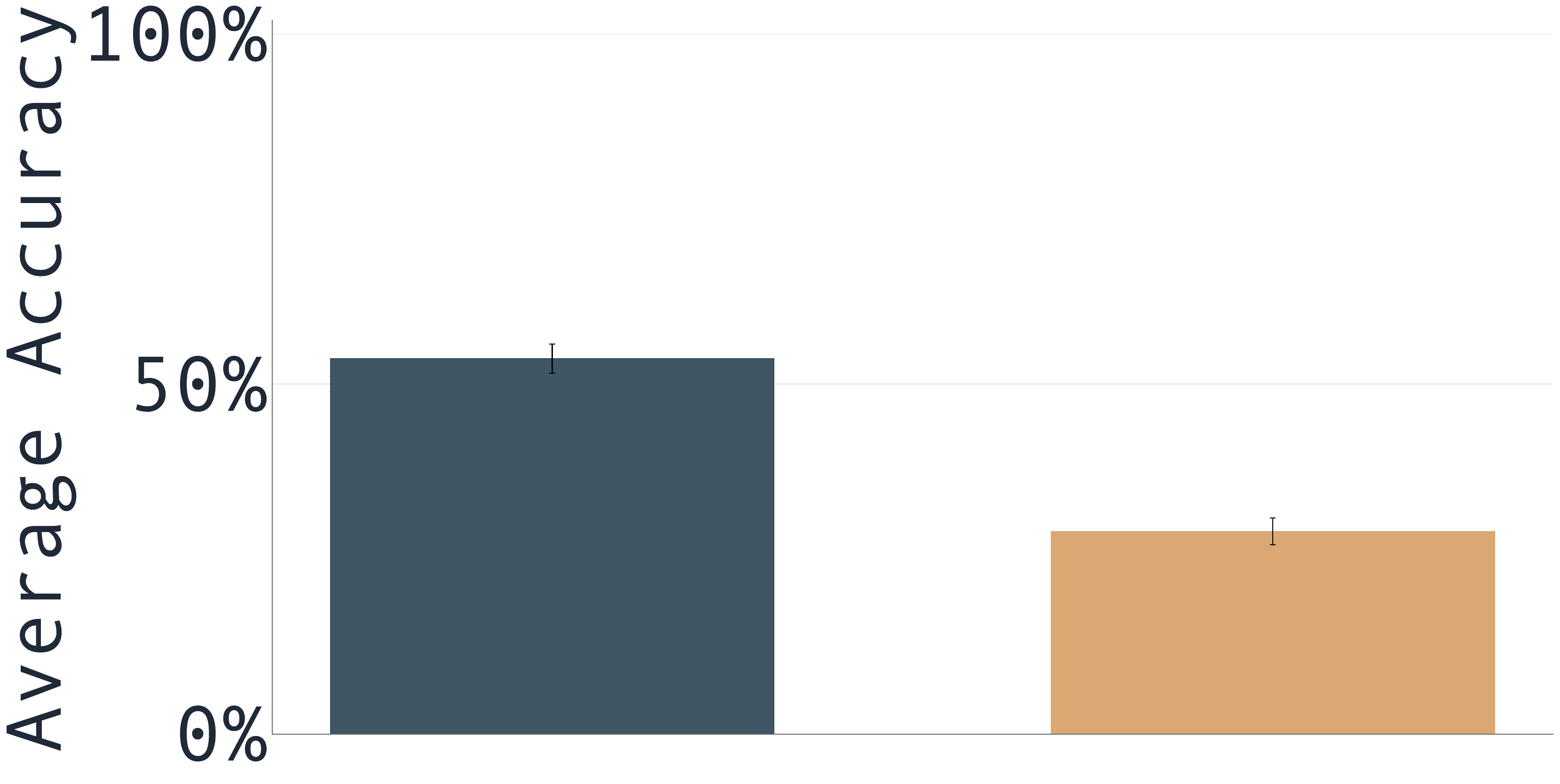}
          \caption{\texttt{Llama-3.1-8B}}
      \end{subfigure}
      \hfill
      \begin{subfigure}[t]{0.23\textwidth}
          \centering
          \includegraphics[width=\linewidth]{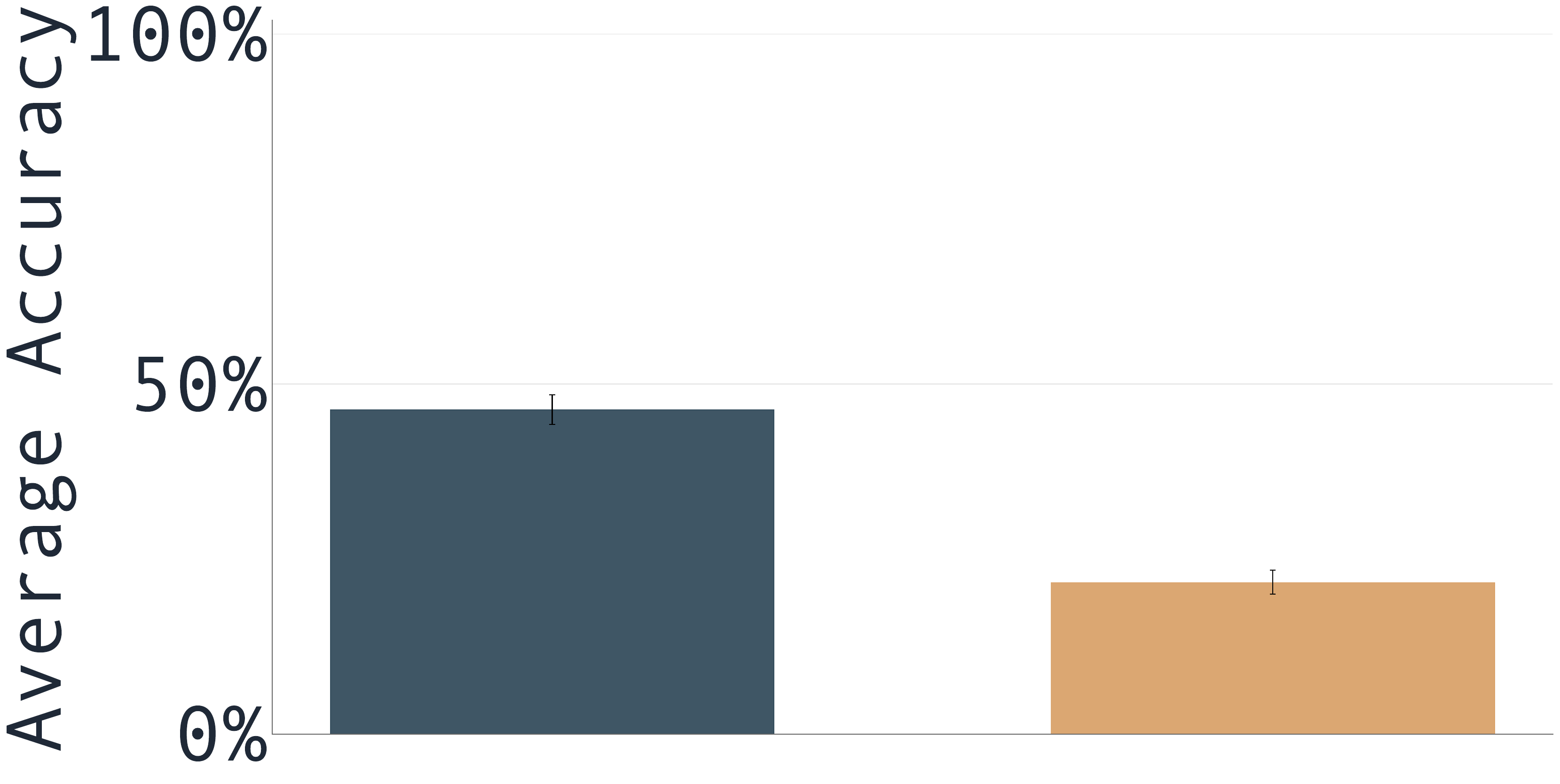}
          \caption{\texttt{Qwen 2.5-14B}}
      \end{subfigure}
      \hfill
      \begin{subfigure}[t]{0.23\textwidth}
          \centering
          \includegraphics[width=\linewidth]{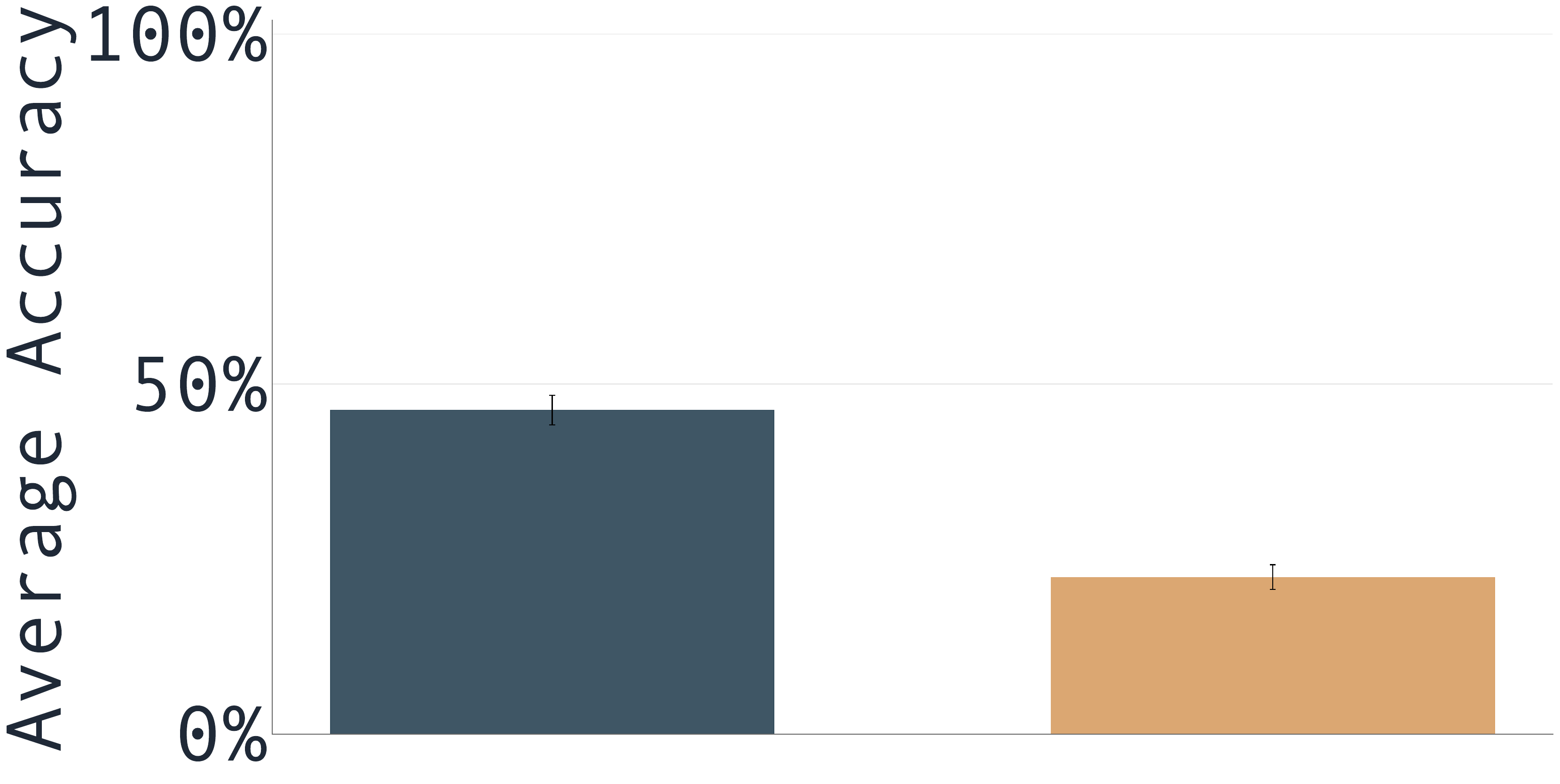}
          \caption{\texttt{Qwen3-4B}}
      \end{subfigure}
      \hfill
      \begin{subfigure}[t]{0.23\textwidth}
          \centering
          \includegraphics[width=\linewidth]{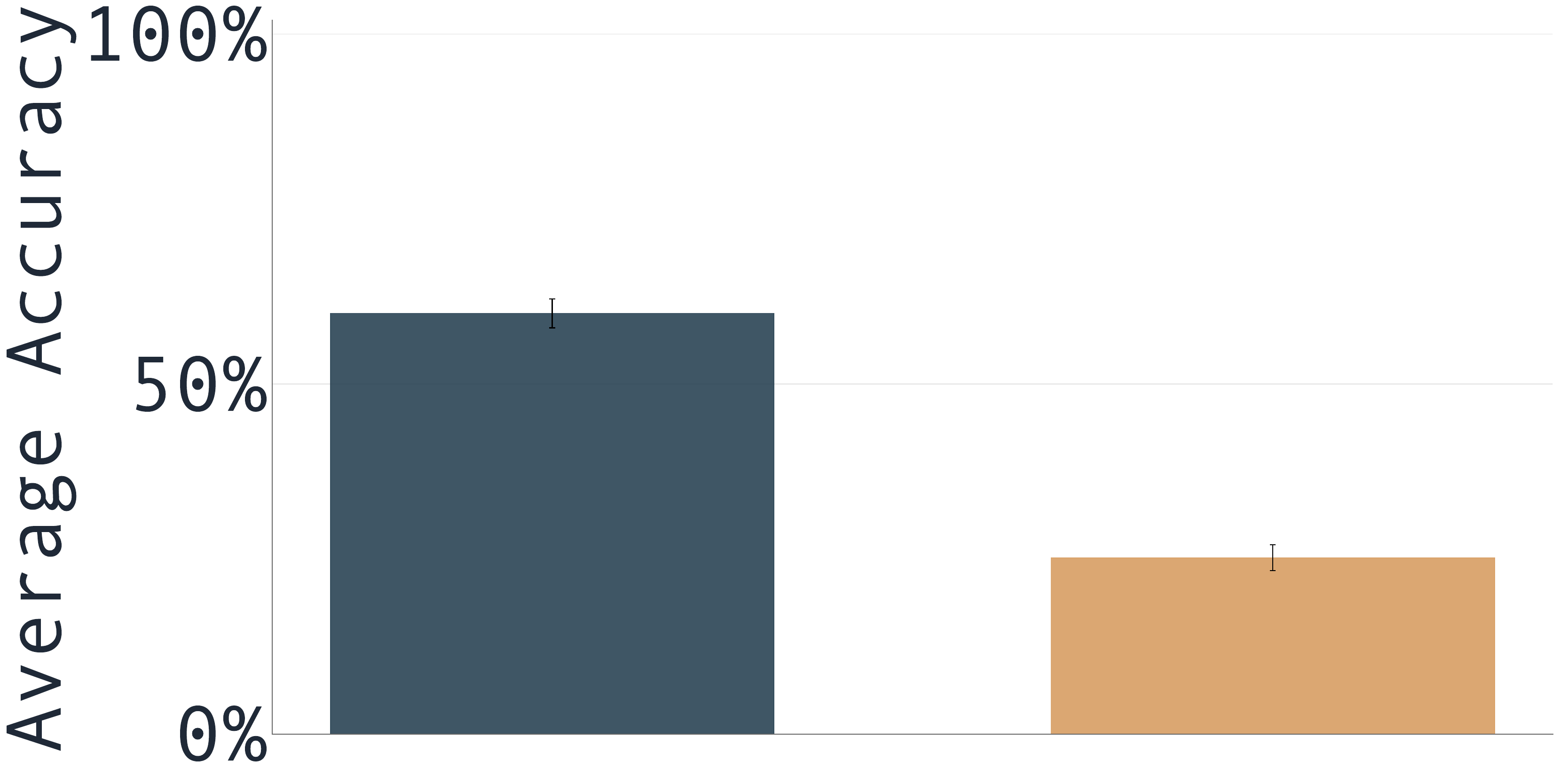}
          \caption{\texttt{Medgemma-4B}}
      \end{subfigure}
      \hfill
      \begin{subfigure}[t]{0.23\textwidth}
          \centering
          \includegraphics[width=\linewidth]{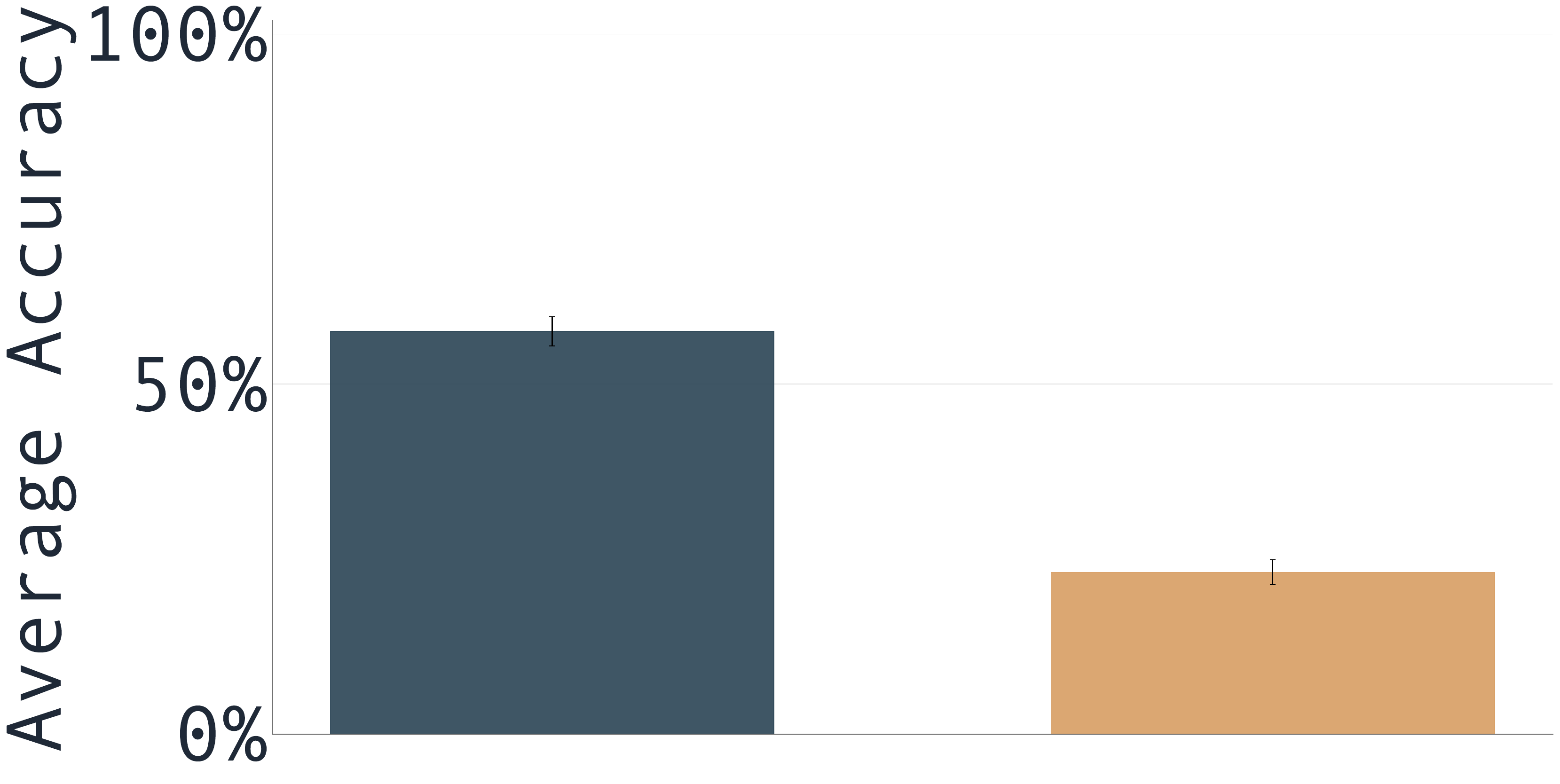}
          \caption{\texttt{Olmo-3-7B}}
      \end{subfigure}
      \hfill
      \begin{subfigure}[t]{0.23\textwidth}
          \centering
          \includegraphics[width=\linewidth]{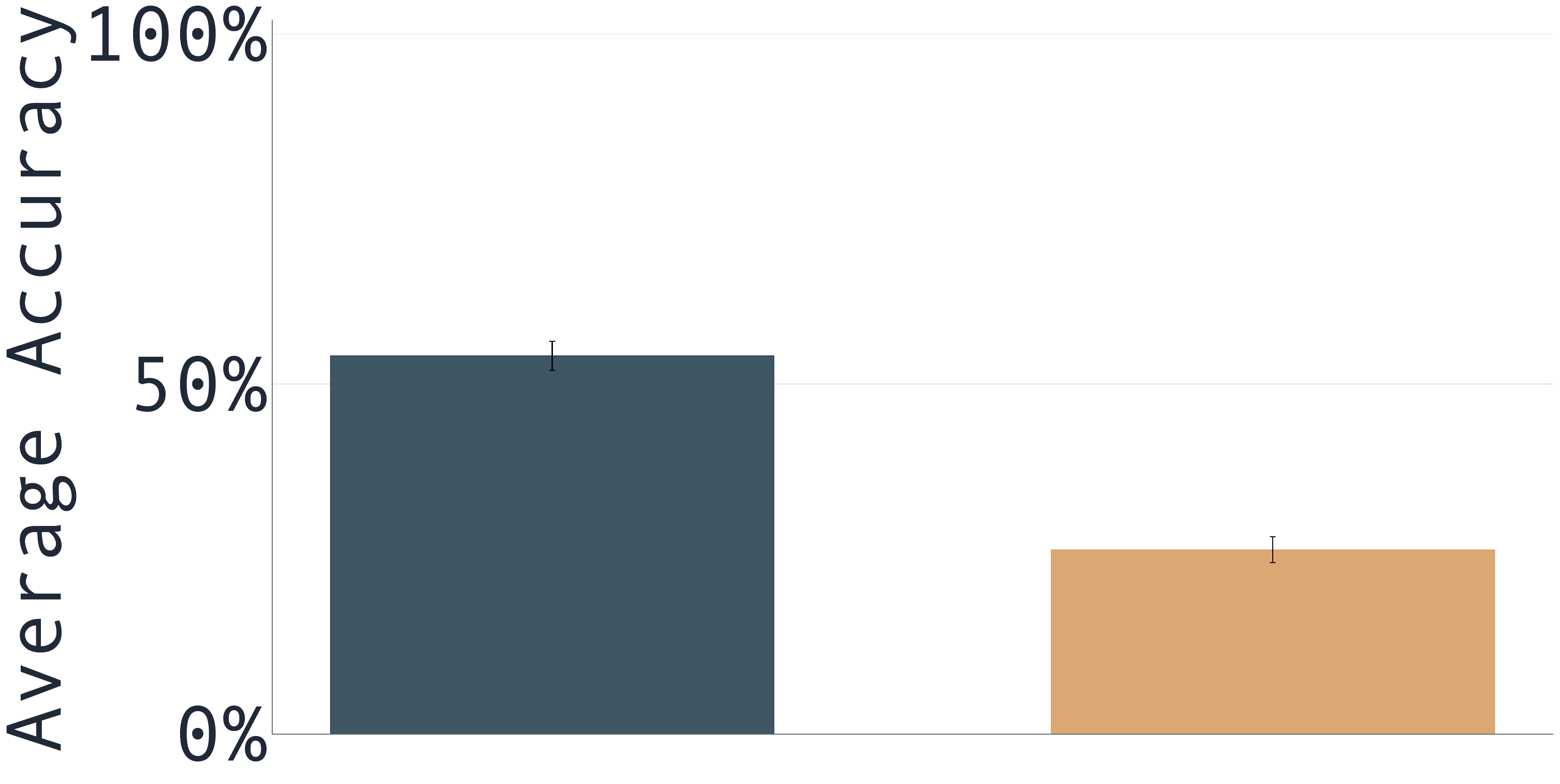}
          \caption{\texttt{GPT-OSS-20B}}
      \end{subfigure}
      \hfill
      \begin{subfigure}[t]{0.23\textwidth}
          \centering
          \includegraphics[width=\linewidth]{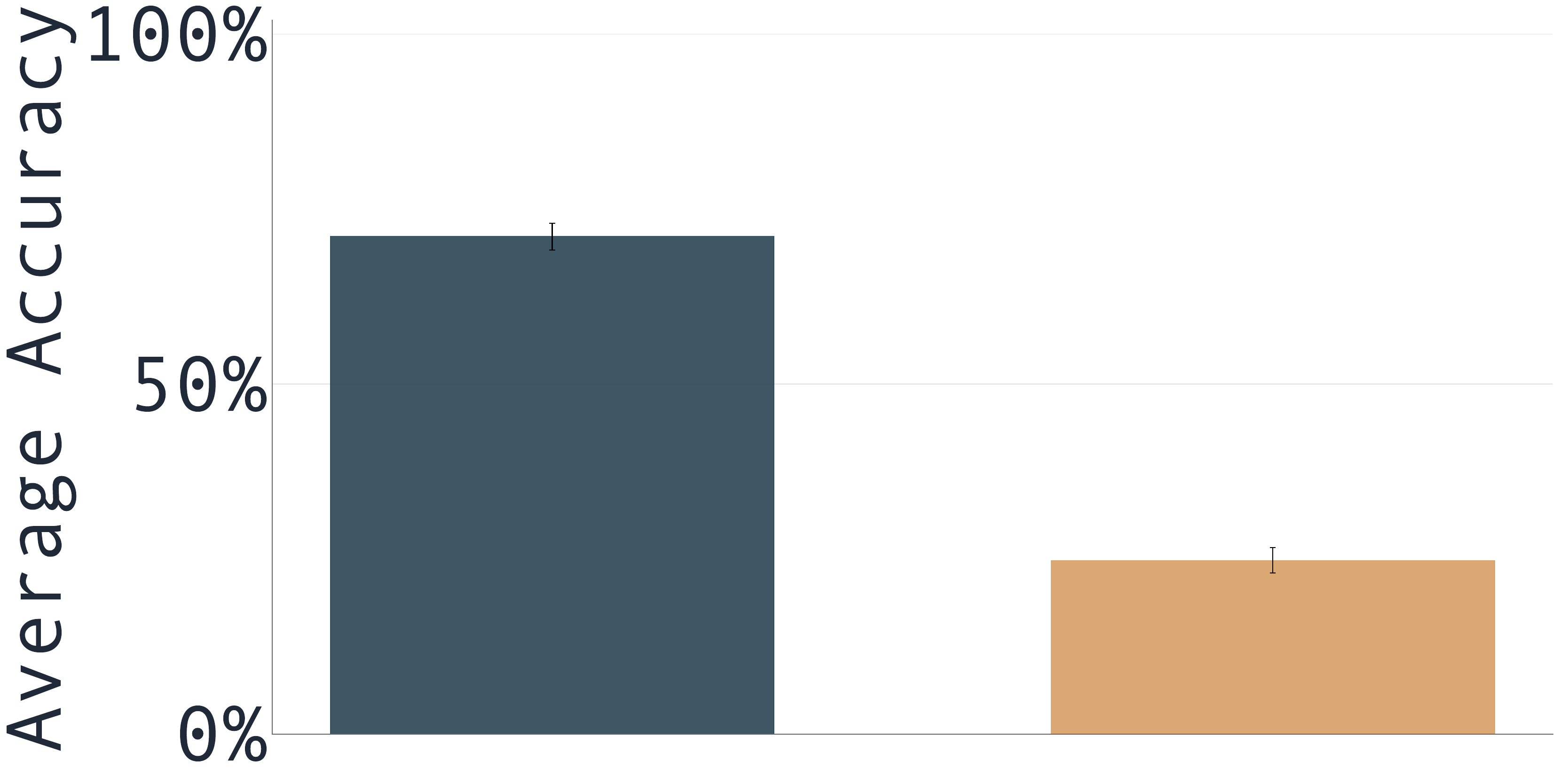}
          \caption{\texttt{GPT-4.1}}
      \end{subfigure}
      \hfill
      \begin{subfigure}[t]{0.23\textwidth}
          \centering
          \includegraphics[width=\linewidth]{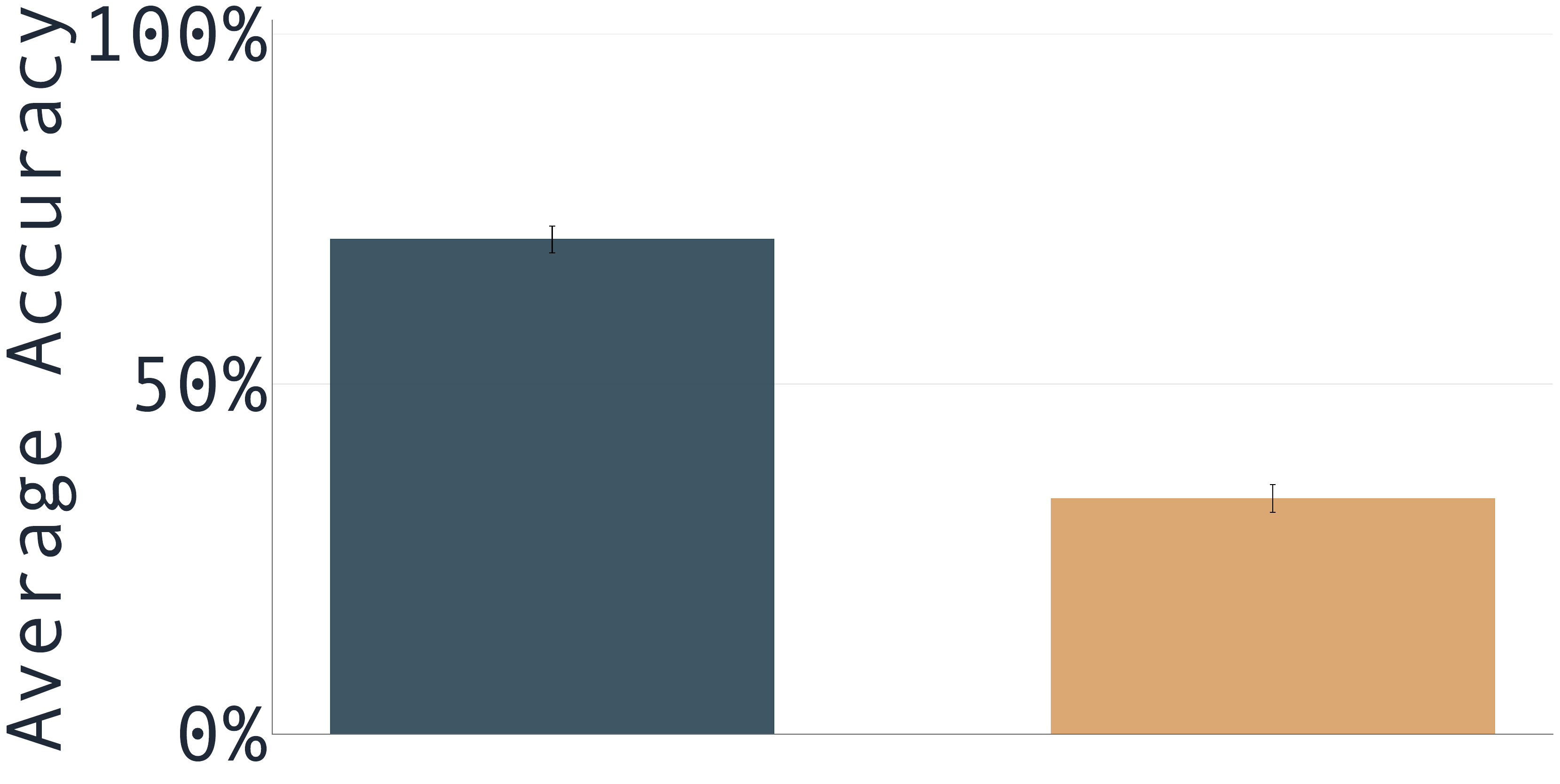}
          \caption{\texttt{GPT-5}}
      \end{subfigure}
      \caption{Accuracy comparison between up-to-date knowledge and historical knowledge. The error bars indicate 95\% confidence intervals. 
      }
      \label{fig:compare_accuracy}
      \vspace{-5mm}
\end{figure*}

\textbf{Experimental Setup.} To investigate the LLMs' capability in retrospective analysis, we evaluate whether LLMs can recall time-specific medical knowledge when asked to answer questions using historical versions of the guidelines. We conduct the experiment by changing the target year in the question stems from the \textit{year of the up-to-date guideline} to the corresponding \textit{historical guideline} with the template below. Appendix~\ref{appendix:example_outdated_questions} provides an example of how we construct such historical-knowledge-targeted questions.
\begin{minted}[breaklines, fontsize=\small, frame=lines, escapeinside=||]{text}
According to the xxx guideline issued in |\textcolor{red}{\{year of the historical guideline\}}|, which option do you think is correct?
\end{minted}

\textbf{Collapsed Accuracy on Outdated-targeted Questions.} Figure~\ref{fig:compare_accuracy} and Figure~\ref{fig:more_results_targeted_accuracy} visualize the comparison of accuracy over the up-to-date-targeted questions (i.e., original MCQ instances in the TempoMed-Bench) and the corresponding historical-knowledge-targeted questions. It is observed that there is clear collapsed accuracy on the historical knowledge: accuracy of historical knowledge is only 25.37\%-53.89\% of up-to-date knowledge, suggesting that the models encode up-to-date knowledge significantly better than the historical ones. This finding also reveals a potential limitation of existing pre-training and post-training stages: models may (un)intentionally forget historical knowledge.


\subsection{RQ3: Can LLMs maintain temporal consistency when recalling medical knowledge across different time periods?}

\textbf{Temporal Consistency.} We define \textit{temporal consistency} as a qualitative concept that characterizes whether an LLM exhibits coherent and stable behavioral patterns when recalling medical knowledge across time. Specifically, temporally consistent behavior does not necessarily require the model to provide factually correct answers for every year; rather, it requires the model’s predictions to evolve in a logically consistent manner over time. In contrast, temporally inconsistent behavior refers to irregular or contradictory prediction patterns that fluctuate unpredictably across neighboring years without coherent temporal transitions. Such behaviors suggest unstable or weakly grounded temporal knowledge representations rather than coherent medical reasoning, thereby undermining the trustworthiness of medical LLMs.

\textbf{Experimental Setup.} To evaluate temporal consistency in recalling the time-specific medical knowledge, we conduct a fine-grained analysis of LLM predictions using the time-specific question templates. Specifically, for each difference pair extracted in the TempoMed-Bench benchmark, we construct queries using the following template:
\begin{minted}[breaklines, fontsize=\small, frame=lines]{text}
According to the most recent guideline on or before {target year}, do you agree with the statement that {statement}?
\end{minted}
We vary the \{target year\} from 2000 to 2026 to simulate the temporal evolution of medical knowledge and evaluate whether models can correctly align their predictions with the appropriate guideline at each time point. We restrict our analysis to trajectories that contain exactly two guideline versions (i.e., one outdated and one updated version) to enable a controlled comparison of model behavior before and after a guideline change. The \{statement\} is constructed based on (1) the updated guideline and (2) the outdated guideline, allowing us to examine whether models can distinguish between up-to-date and outdated medical recommendations. Listing~\ref{listing:example_rq3} shows an example of such questions.

\begin{table}[!t]
    \centering
    \caption{Percentage distribution of the temporal consistency Patterns.}
    \label{tab:percentage_distribution}
    \resizebox{\textwidth}{!}{\begin{tabular}{l|c|cccccc}
    \toprule
        Model Name & \textit{Inconsistency} & \textit{All-True} &  \textit{All-False} & \textit{Only-Know-Latest} & \textit{Only-Know-Prior} & \textit{Wrong-Transition-Point} & \textit{Correct-Transition-Point}\\
        \midrule
        \includegraphics[width=0.025\textwidth]{icons/openai_logo.png} \texttt{GPT-5} & 78.18 & 0.00 & 0.00 & 12.73 & 0.00 & 5.45 & 3.64 \\
         \includegraphics[width=0.025\textwidth]{icons/openai_logo.png} \texttt{GPT-4.1} & 63.64 & 5.45 & 0.00 & 21.82 & 5.45 & 1.82 & 1.82 \\
         \includegraphics[width=0.025\textwidth]{icons/meta_logo.png} \texttt{Llama-3.1-8B} & 49.09 & 14.55 & 7.27 & 21.82 & 7.27 & 0.00 & 0.00  \\
         \includegraphics[width=0.025\textwidth]{icons/qwen_logo.png} \texttt{Qwen2.5-7B} &  58.18 & 9.09 & 5.45 & 14.55 & 12.73 & 0.00 & 0.00\\
         \includegraphics[width=0.025\textwidth]{icons/google_logo.png} \texttt{MedGemma-4B} & 34.55 & 52.73 & 1.82 & 9.09 & 1.82 & 0.00 & 0.00\\
    \bottomrule
    \end{tabular}}
\end{table}
\textbf{Diversified Consistency Patterns.} For each question instance, this analysis probes the model’s encoded knowledge across different years, resulting in a two-row heatmap where each column represents the varying years and each row represents the up-to-date statement and the outdated statement, respectively. Ideally, the model’s responses should reflect the temporal validity of the statements. For the up-to-date statement, the model should answer ``No'' before the up-to-date guideline year while answering ``Yes'' thereafter; Conversely, for the outdated statement, the model should answer ``Yes'' before the update and ``No'' afterwards. This results in a zig-zag transition pattern centered around the guideline update year (named as \textit{Correct-Transition-Point}). Empirically, however, this ideal pattern rarely occurs. Instead, we observe several iconic temporal consistency patterns in the probing heatmaps:

\begin{itemize}[leftmargin=*]
    \item \textit{All-True}: the model answers ``Yes'' for both statements across all years.
    \item \textit{All-False}: the model answers ``No'' for both statements across all years.
    \item \textit{Only-Know-Latest}: the model consistently answers ``Yes'' for the up-to-date statement and ``No'' for the outdated statement, regardless of the year.
    \item \textit{Only-Know-Prior}: the model consistently answers ``No'' for the up-to-date statement and ``Yes'' for the outdated statement, regardless of the year.
    \item \textit{Wrong-Transition-Point}: the zig-zag transition pattern appears, but the transition occurs at a year that deviates significantly from the true guideline update year.
\end{itemize}

All other heatmaps that do not belong to the above categories are deemed as \textit{Inconsistency}. We present examples of each error pattern in Appendix~\ref{appendix:error_pattern}. Each error pattern reveals a distinct limitation in how LLMs encode and utilize temporally evolving medical knowledge: \textit{All-True} indicates that the model exhibits a strong affirmative bias, accepting both outdated and up-to-date statements regardless of their temporal validity. \textit{All-False} suggests that the model systematically rejects both statements, indicating a lack of confidence or failure to recognize valid medical knowledge. \textit{Only-Know-Latest} shows that the model consistently favors the up-to-date statement across all years, ignoring temporal constraints. \textit{Only-Know-Prior} indicates that the model consistently relies on outdated knowledge, failing to incorporate more recent updates. \textit{Wrong-Transition-Point} reflects cases where the model captures the existence of a knowledge shift but misplaces its timing.

\textbf{Pattern Distributions.} To analyze each LLM's internal preference towards the temporal medical knowledge, we conduct a statistical analysis on the distributions of these patterns for each LLM model, with the results shown in Table~\ref{tab:percentage_distribution}. Out of all the error patterns, \textit{Only-Know-Latest} and \textit{All-True} are the top-2 major patterns, further indicating the conclusions in the previous sections: the LLMs might have encoded the latest knowledge but lack temporal awareness of the medical knowledge.

\subsection{RQ4: Will the problem be solved by using agentic RAG?}\label{sec:analysis_agentic}

\begin{wraptable}{r}{0.53\textwidth}
    \vspace{-3mm}
    \centering
    \caption{Accuracy on up-to-date and historical knowledge after integrating agentic RAG tools. 
    }
    \label{tab:accuracy_agentic_rag}
    \resizebox{0.53\textwidth}{!}{
    \begin{tabular}{cc|cc}
    \toprule
        Backend LLM & Agentic RAG & Up-to-date & Outdated \\
         \midrule
         \multirow{3}{*}{\texttt{GPT-4.1}} 
         & / & 78.1 & 27.35 \\
         & ToolUniverse & 76.9 {\color{red}(-1.2)} & 30.93 {\color{green!40!black}(+3.58)} \\
         & Biomni & 79.20 {\color{green!40!black}(+1.10)} & 29.56 {\color{green!40!black}(+2.21)}\\
         \midrule
         \multirow{3}{*}{\texttt{GPT-5}} 
         & / & 79.7 & 37.73 \\
         & ToolUniverse & 85.3 {\color{green!40!black}(+5.6)} & 51.87 {\color{green!40!black}(+14.14)} \\
         & Biomni & 76.55 {\color{red}(-3.15)} & 43.62{\color{green!40!black}(+5.89)} \\
         \bottomrule
    \end{tabular}}
    \vspace{-4mm}
\end{wraptable}

\textbf{Experimental Setup.} In the era of agentic AI, an intuitive solution to the above problems seems to be integrating the LLMs with retrieval tools (e.g., ToolUniverse~\cite{gao2025democratizing} and Biomni~\cite{huang2025biomni}), which can retrieve additional relevant guideline information (e.g., from PubMed and PMC) to the context. We therefore evaluate (1) biomni and (2) a single LLM integrated with ToolUniverse over the 721 question instances in the TempoMed-Bench. Appendix~\ref{appendix:agentic_search_tools} provides details for the implementations.

\textbf{Limited Improvements in Accuracy.} Table~\ref{tab:accuracy_agentic_rag} compares the performance before and after integrating the agentic RAG tools, under two different backend LLMs: \texttt{GPT-4.1} and \texttt{GPT-5}. As shown, the agentic RAG tools only bring limited improvements or even backfire in accuracy.

\textbf{Analysis of Limited Improvements.} Appendix~\ref{appendix:example_tooluniverse_model_answer} presents a representative failure case in which the model produces an incorrect answer despite successfully retrieving and incorporating the relevant guideline into the context. We attribute this failure primarily to conflicts among multiple pieces of retrieved information. Specifically, although the oracle guideline (e.g., PMID 37129716) is correctly retrieved via the `PMC\_search\_papers' tool, additional documents are introduced into the context through the `Web\_Search' tool. The presence of these potentially inconsistent sources increases the cognitive load on the model and may lead to interference between competing pieces of information, ultimately degrading prediction accuracy~\cite{amiraz2025distracting, shi2023large}.

\subsection{Limitations and Future Work}
While our study provides a first-of-its-kind dataset for evaluating the temporal medical knowledge of LLMs, several limitations remain and could be addressed in future work: (1) We apply strict criteria when extracting guideline differences, resulting in a relatively limited dataset containing only 721 instances; (2) Although we observe limited improvements when using off-the-shelf agentic RAG tools, we have not yet explored how to further improve performance through techniques such as fine-tuning or more advanced agentic systems; (3) Due to limited full-text availability in the PubMed database (only the PMC provides full-text access), our trajectory dataset remains incomplete.

\section{Conclusion}
In this paper, we introduce TempoMed-Bench, the first benchmark for evaluating the temporal awareness of LLMs in the medical domain. Our empirical results demonstrate that even state-of-the-art models exhibit limited temporal awareness when handling evolving medical knowledge. Furthermore, integrating agentic RAG systems does not fully resolve this limitation. We hope this work highlights an important yet underexplored challenge and motivates future research on developing LLMs that can more effectively encode and reason over time-specific medical knowledge.

\section*{Acknowledgments}
This research was partly supported by the Intramural Research Program of the National Institutes of Health (NIH). The contributions of the NIH author(s) are considered Works of the United States Government. This research was also partially supported by the NIH Pathway to Independence Award K99LM014903 (Q.J.). The findings and conclusions presented in this paper are those of the author(s) and do not necessarily reflect the views of the NIH or the U.S. Department of Health and Human Services.
Zihan Guan and Anil Vullikanti were partly supported by NSF grants CCF-1918656 and CNS-2317193.

\clearpage
\bibliography{references}
\bibliographystyle{plain}

\clearpage
\appendix
\startcontents[appendix]
\printcontents[appendix]{l}{1}{\section*{\centering Appendix Contents}}

\clearpage

\section{Data Collection Pipeline}

\begin{wrapfigure}{r}{6cm}
  \centering
  \vspace{-3mm}
  \includegraphics[width=\linewidth]{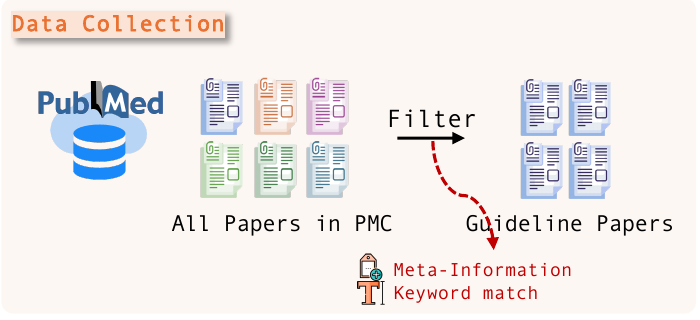}
  \caption{Filtering guideline papers from PubMed Central.} \label{fig:pipeline_step1}
  \vspace{-3mm}
\end{wrapfigure}

\subsection{Post-processing Guideline Trajectories}\label{appendix:post-processing}
We notice that the above candidate trajectory might not be complete either because the current guideline may not mention all the prior guidelines, or the LLMs might omit some by hallucination. Therefore, we use another source of information to reliably augment the guideline trajectory. Specifically, PubMed has provided the \{Similar Articles\} for a given paper, which contains similar articles ranked by the related citation algorithm\footnote{\url{https://pubmed.ncbi.nlm.nih.gov/help/\#computation-of-similar-articles}}. This source is complementary to the text information that we used in the previous step, but not a dominating one. For example, in the 1{,}586 trajectories of \texttt{comm} subset, 1{,}505 trajectories of them contain at least one prior guideline paper that cannot be identified in the retrieved \{Similar Articles\}. The overall prompt is present in Listing~\ref{prompt:check_prior}.

Each paper in the PMC database has a unique PMID and a unique PMCID. We also notice that some extracted trajectories might unexpectedly have some missing PMIDs due to the generation errors of the LLM. We carefully restore them with the recorded PMCID. Moreover, we also filter the trajectories whose current guideline is essentially a prior guideline of another trajectory, to avoid the guideline trajectory starting with some outdated guideline.

After the augmentation, there is at least one prior guideline for 1,496 papers in the \texttt{comm} subset, 1,691 papers in the \texttt{noncomm} subset, and 224 papers in the \texttt{other} subset, resulting in a total of 3,411 guideline trajectories.

\begin{figure*}[t]
      \centering
      \begin{subfigure}[t]{0.32\textwidth}
          \centering
          \includegraphics[width=\linewidth]{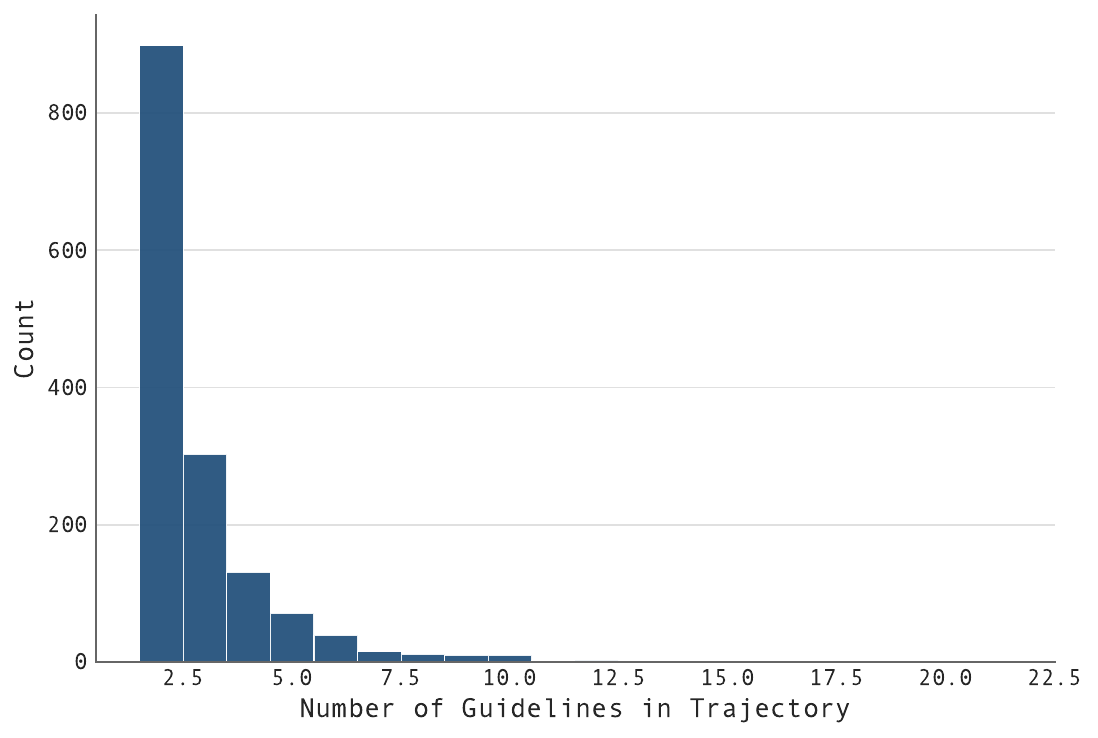}
          \caption{\texttt{comm} subset}
          \label{fig:trajectory_length_comm_plot}
      \end{subfigure}
      \hfill
      \begin{subfigure}[t]{0.32\textwidth}
          \centering
          \includegraphics[width=\linewidth]{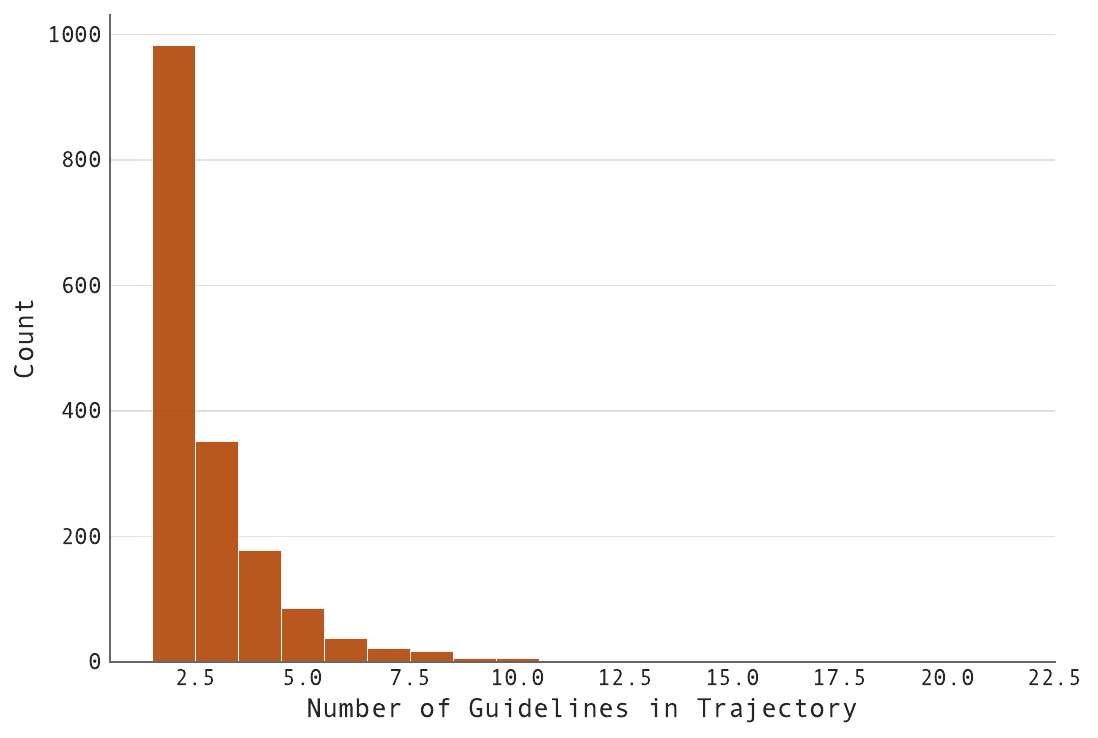}
          \caption{\texttt{noncomm} subset}
          \label{fig:trajectory_length_noncomm_plot}
      \end{subfigure}
      \hfill
      \begin{subfigure}[t]{0.32\textwidth}
          \centering
          \includegraphics[width=\linewidth]{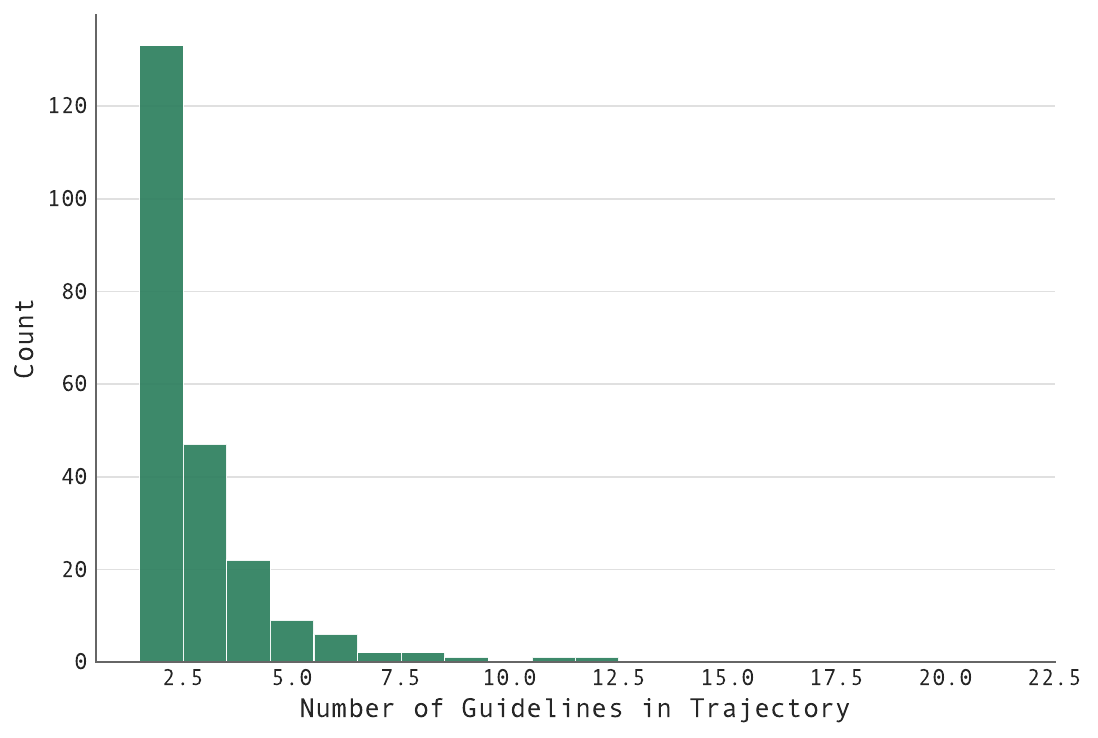}
          \caption{\texttt{other} subset}
          \label{fig:trajectory_length_other_plot}
      \end{subfigure}
      \caption{Statistical analysis of the guideline trajectories in the TempoMed-Traj. Distribution of trajectory lengths across the three subsets, excluding trajectories with only one node.}
      \label{fig:trajectory_length}
\end{figure*}

\begin{figure*}[t]
      \centering
      \begin{subfigure}[t]{0.32\textwidth}
          \centering
          \includegraphics[width=\linewidth]{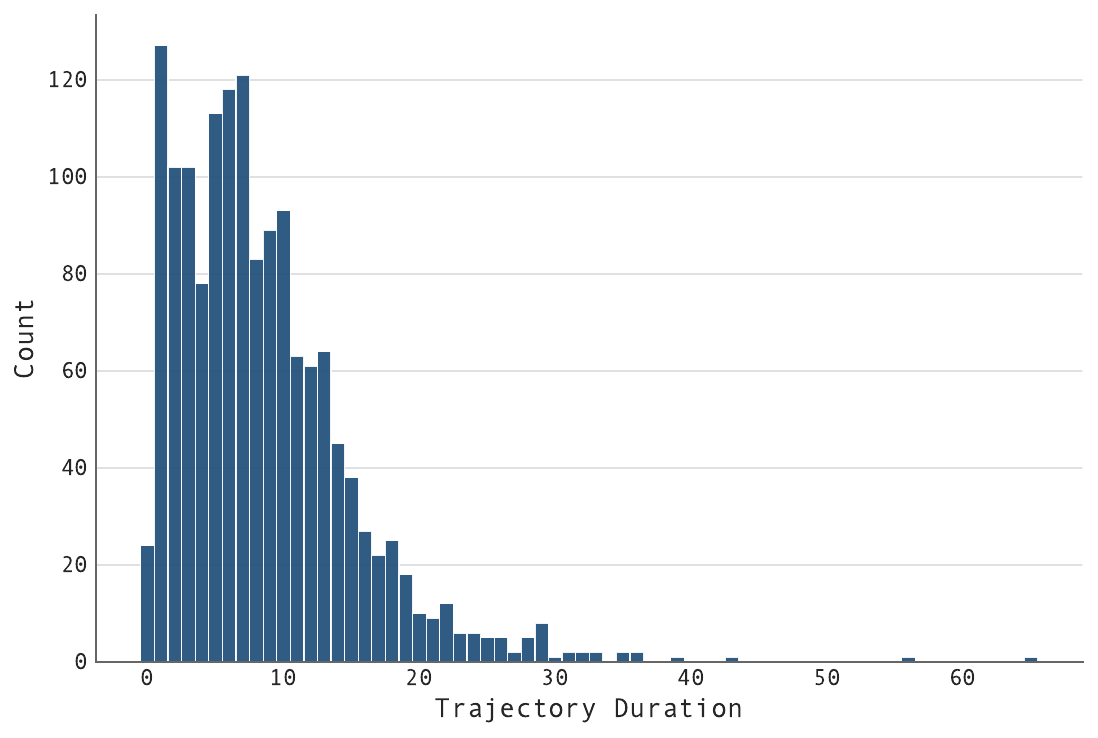}
          \caption{\texttt{comm} subset}
          \label{fig:trajectory_timespan_comm_plot}
      \end{subfigure}
      \hfill
      \begin{subfigure}[t]{0.32\textwidth}
          \centering
          \includegraphics[width=\linewidth]{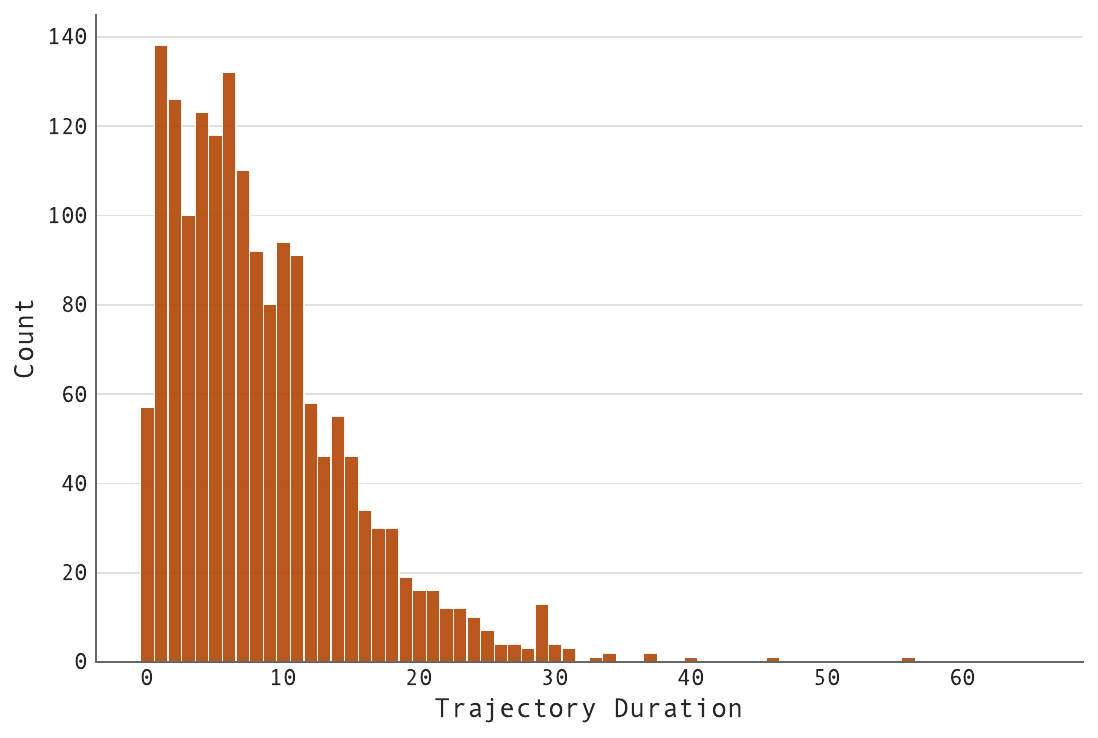}
          \caption{\texttt{noncomm} subset}
          \label{fig:trajectory_timespan_noncomm_plot}
      \end{subfigure}
      \hfill
      \begin{subfigure}[t]{0.32\textwidth}
          \centering
          \includegraphics[width=\linewidth]{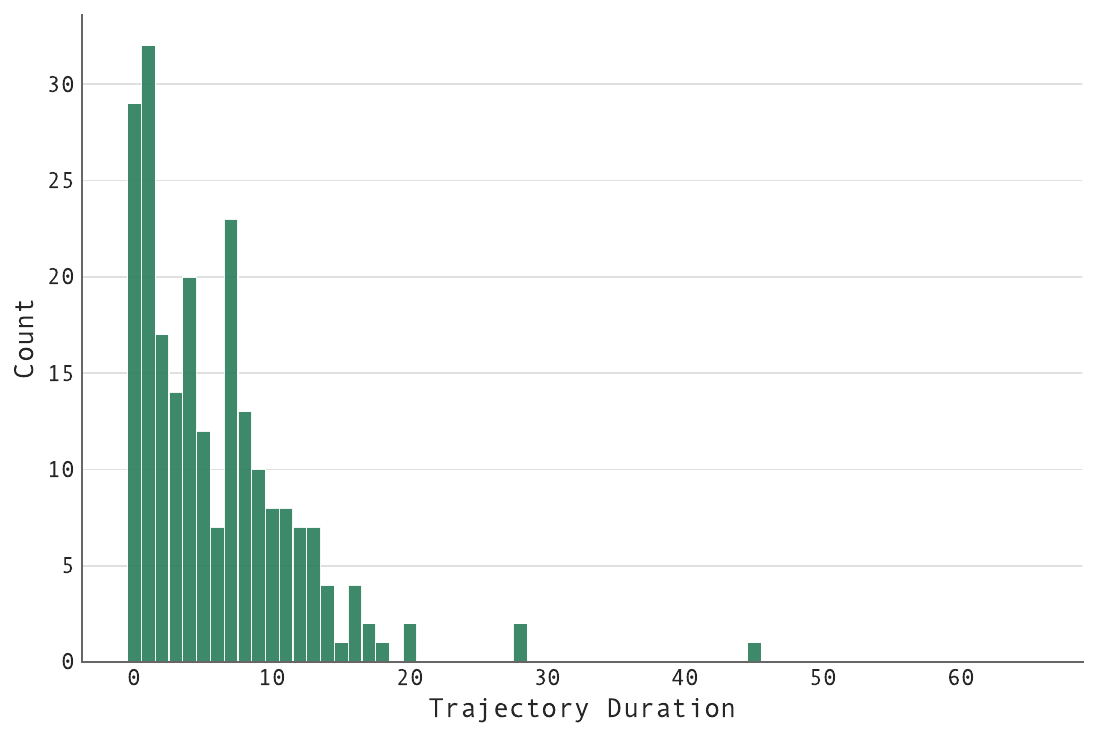}
          \caption{\texttt{other} subset}
          \label{fig:trajectory_timespan_other_plot}
      \end{subfigure}
      \caption{Statistical analysis of the guideline trajectories in the TempoMed-Traj. Distribution of the guideline year span in the three subsets, excluding trajectories with only one node.}
      \label{fig:trajectory_timespan}
\end{figure*}

\begin{figure*}[t]
      \centering
      \begin{subfigure}[t]{0.32\textwidth}
          \centering
          \includegraphics[width=\linewidth]{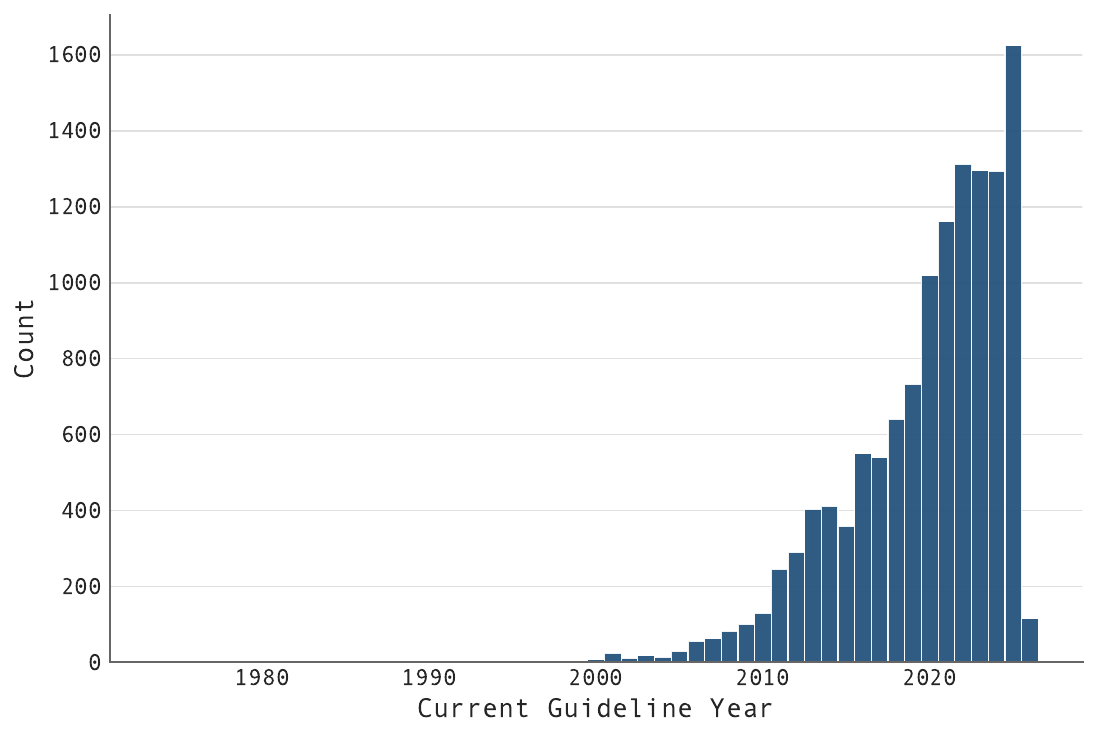}
          \caption{\texttt{comm} subset}
          \label{fig:current_guideline_year_comm_plot}
      \end{subfigure}
      \hfill
      \begin{subfigure}[t]{0.32\textwidth}
          \centering
          \includegraphics[width=\linewidth]{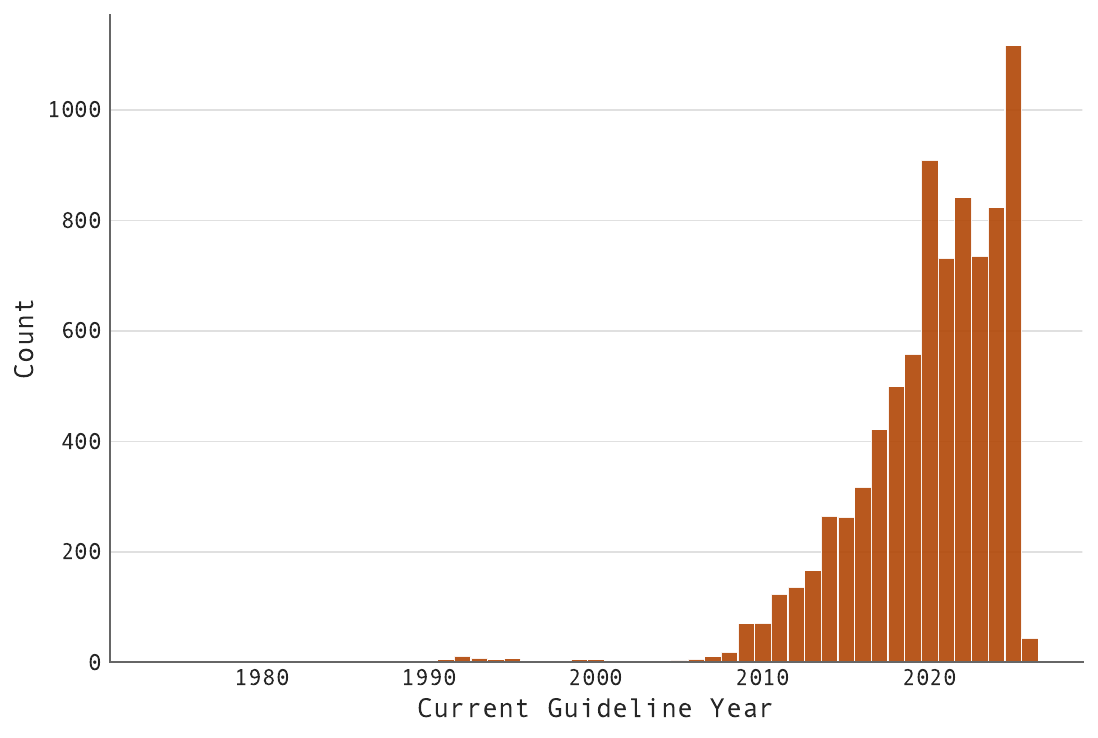}
          \caption{\texttt{noncomm} subset}
          \label{fig:current_guideline_year_noncomm_plot}
      \end{subfigure}
      \hfill
      \begin{subfigure}[t]{0.32\textwidth}
          \centering
          \includegraphics[width=\linewidth]{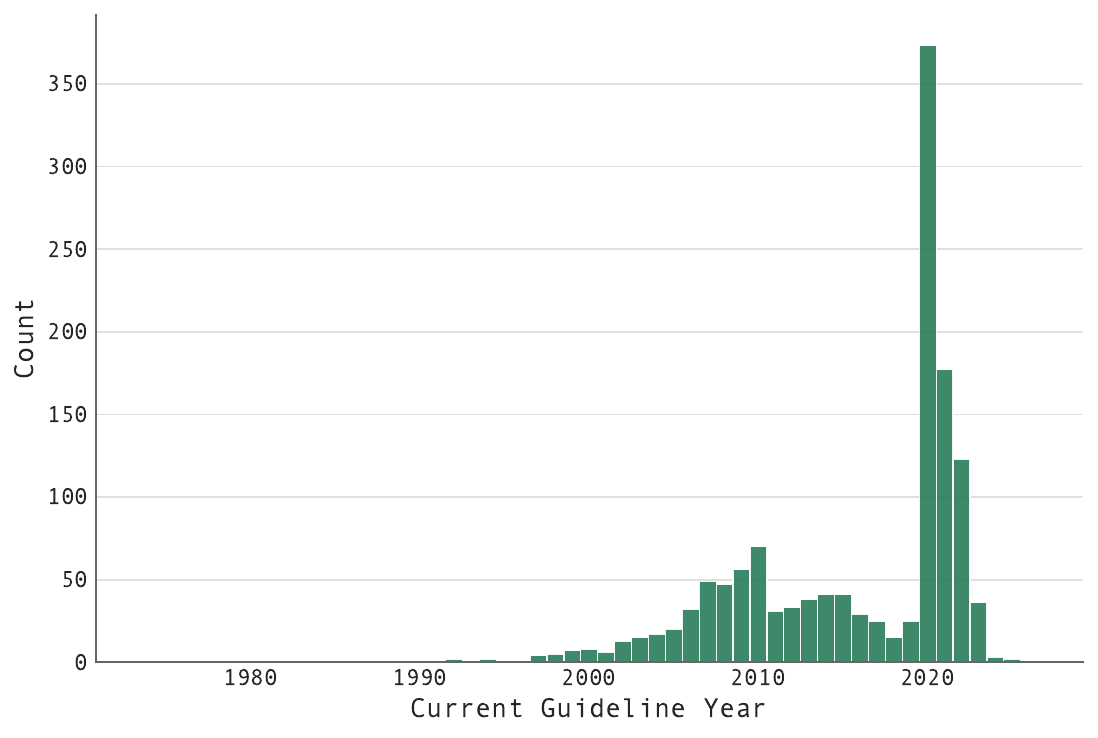}
          \caption{\texttt{other} subset}
          \label{fig:current_guideline_year_other_plot}
      \end{subfigure}
      \caption{Statistical analysis of the guideline trajectories in the TempoMed-Traj. Distribution of the current guideline year in the three subsets.}
      \label{fig:current_guideline_year}
\end{figure*}

\begin{figure}
    \centering
    \includegraphics[width=\linewidth]{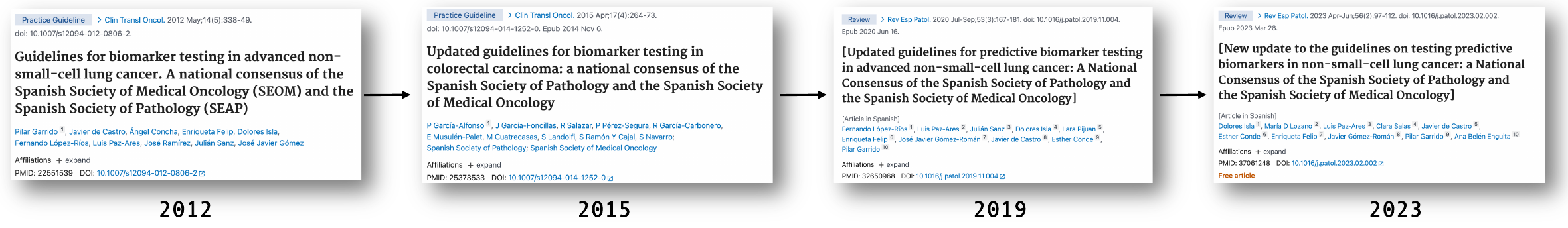}
    \caption{A typical example of the guideline trajectory.}
    \label{fig:trajectory-example}
\end{figure}

We provide visualizations for the following statistical analysis of the final post-processed trajectories. Figures~\ref{fig:trajectory_length}, ~\ref{fig:trajectory_timespan}, and ~\ref{fig:current_guideline_year} characterize the trajectory dataset from structural and temporal perspectives across the \texttt{comm}, \texttt{noncomm}, and \texttt{other} subsets. Figure~\ref{fig:trajectory_length} presents the distribution of trajectory lengths, measured by the number of guideline nodes in each trajectory, with single-node trajectories excluded. Figure~\ref{fig:trajectory_timespan} presents the distribution of trajectory spans, defined as the difference between the earliest and latest guideline years in each trajectory. Figure~\ref{fig:current_guideline_year} presents the distribution of current guideline years, capturing the recency profile of the most recent guideline in each trajectory. Within each figure, the three subfigures correspond to the \texttt{comm}, \texttt{noncomm}, and \texttt{other} subsets. Figure~\ref{fig:trajectory-example} presents a typical example of the guideline trajectory in our dataset. As shown, the initial version was released in 2012 and underwent several updates in 2015, 2019, and 2023. They are all issued by the Spanish Society of Pathology and the Spanish Society of Medical Oncology and focus on the same topic of polycystic ovarian syndrome.

\subsection{Extracting Differences from the Guideline Trajectories}

Given the guideline trajectories, the next step is to extract the evolving changes between the up-to-date guideline and each of its previous outdated versions. To achieve this, we prompt LLMs to identify these differences, and the extracted changes are subsequently verified by a separate, strict verifier LLM. 
Listing~\ref{prompt:extract_difference} and Listing~\ref{prompt:verifier} present the prompt to the LLM. As seen, the extracted difference pair between each up-to-date guideline and outdated guideline must represent a direct head-to-head comparison and be meaningfully different. Then, a verifier LLM scores the extracted difference pair with detailed rubrics.

Note that in this step, we only consider the trajectory where all the guideline nodes are in the guideline set we filtered from the PMC database, since otherwise, we would not be guaranteed to get the body text of the paper. 

\subsection{Generating QAs from the Extracted Differences}
For each of the extracted differences, we construct a corresponding evaluation question that aims to test whether the LLMs' knowledge is closer to the up-to-date knowledge or to the outdated knowledge. To align with the traditional exam style, we design QAs according to the NBME (National Board of Medical Examiners) guideline\footnote{\url{https://www.nbme.org/sites/default/files/2021-02/NBME\_Item\%20Writing\%20Guide\_R_6.pdf}}, where a typical sample of the NBME-style question is shown below

\begin{minted}[breaklines, fontsize=\small, frame=lines]{text}
VIGNETTE:
A 32-year-old man comes to the office because of a 4-day history of progressive weakness in his
extremities. He has been healthy except for an upper respiratory tract infection 10 days ago. His
temperature is 37.8°C (100.0°F), pulse is 94/min, respirations are 42/min and shallow, and blood pressure
is 130/80 mm Hg. He has symmetric weakness of both sides of the face and the proximal and distal muscles
of the extremities. Sensation is intact. No deep tendon reflexes can be elicited. Babinski sign is absent.

LEAD-IN:
Which of the following is the most likely diagnosis?

OPTION SET:
A. Acute disseminated encephalomyelitis
B. Guillain-Barré syndrome*
C. Myasthenia gravis
D. Poliomyelitis
E. Polymyositis
\end{minted}

Specifically, for each extracted difference, we prompt the LLM as in Listing~\ref{prompt:qa_generation}, where the LLM is expected to generate the NBME-style question stem (including vignette and lead-in) and option set for each QA instance. 

The option set contains five options, which contain an option that aligns with the knowledge in the up-to-date guideline, an option that aligns with the knowledge in the outdated guideline, and two distractor options that align with neither of the two. We also added a `I do not know' option.

\section{Prompts}
\begin{minted}[breaklines, fontsize=\small, frame=lines]{text}
Based on the given article, identify **all prior clinical guidelines** issued by the **same organization** that address the **same topic** as the current guideline discussed in the article.

A prior guideline is defined as an official guidance document issued by the **same organization** before the year of the current guidance and focuses on the **same topic** as the current guideline.

You are encouraged to identify a comprehensive list of such prior guidelines.

For each identified prior guideline:

- Extract the year of publication
- Extract the PMID (if available)
- Extract the issuing organization
- Extract the title
- Provide the reason why this is a prior guideline

## Process:
0. Determine whether the paper itself is a clinical guideline issued by a professional society (e.g., the American College of Rheumatology).
    - If yes, proceed with the steps below and extract prior guidelines.
    - If no, do not record any prior guidelines and set the prior guidelines as an empty list.
1. Examine the *Introduction*, *Related Works* or *Literature Review* sections to identify documents cited as prior guidelines that **meet the above criteria**;
2. Record the `year`, `PMID`, `Organization`, and `title` of each previous guideline.
3. Explain why each document is identified as a prior guideline that precedes the current one (e.g., explicit citation, chronological ordering, or stated replacement).

## Important Notes:
- If multiple prior guidelines exist, include all of them.
- If a field is not available, return `None` for that field. For example, if the old guideline is from the official website with no PMIDs, you can just use PMID: None
- **DO NOT generate or hallucinate PubMed IDs.**
- If no prior guidelines are identified, return an empty list.
- Do **not** classify a document as a prior guideline if it is merely a translation of the current guideline with no substantive content changes (e.g., Chinese or Japanese versions).
```json
{{
  "Topic": "Systemic Lupus Erythematosus",
  "Title": "2025 American College of Rheumatology (ACR) Guideline for the Treatment of Systemic Lupus Erythematosus",
  "year_of_current_guidance": 2025,
  "Organization": ["American College of Rheumatology"],
  "PMID": 41182321
  "prior_guidelines": [
    {{
      "year": 1999,
      "PMID": 10513791,
      "Organization": "American College of Rheumatology",
      "title": "Guidelines for referral and management of systemic lupus erythematosus in adults. American College of Rheumatology Ad Hoc noncommittee on Systemic Lupus Erythematosus Guidelines"
      "reason": "The paper explicitly mentions that 'Guidelines for management of SLE in adults were lastpublished by the American College of Rheumatology in 1999[8].' and [8] corresponds to PMID 10513791."
    }}
  ]
}}
```
\end{minted}
\captionof{listing}{Prompt for extracting guideline trajectories}
\label{prompt:guideline_traj}

\begin{minted}[breaklines, fontsize=\small, frame=lines]{text}
Determine whether the candidate paper is a PRIOR VERSION of the current guideline paper.

Return JSON only with these keys:
is_prior_version: true/false
confidence: integer 0-100
same_topic: true/false
same_or_related_organization: true/false
is_guideline_like: true/false
reason: short string

Decision rule for true:
- candidate is older than current guideline
- candidate is on the same clinical topic
- candidate appears to be a guideline/consensus/statement/recommendation/practice guidance paper
- candidate is likely an earlier version or closely preceding edition from the same or clearly related issuing organization
- do not mark true for general reviews, background studies, or unrelated society papers

Current guideline PMID: {record.get('PMID')}
Current guideline title: {normalize_text(record.get('Title'))}
Current guideline year: {record.get('year_of_current_guidance')}
Current guideline topic: {normalize_text(record.get('Topic'))}
Current guideline organizations: {current_orgs}
Known existing prior guidelines:
{existing_priors}

{candidate_block}
\end{minted}
\captionof{listing}{Prompt for checking similar papers retrieved from the PubMed.}
\label{prompt:check_prior}

\begin{minted}[breaklines, fontsize=\small, frame=lines]{text}
Compare the CURRENT guideline with the PRIOR guideline.

Topic: {topic}
Organization: {organization}

=============================CURRENT GUIDELINE XML=============================
{current_xml}

=============================PRIOR GUIDELINE XML=============================
{prior_xml}

=============================Instructions=============================
1. Please Extract:
    - The key clinical recommendation in the CURRENT guideline.
    - The corresponding recommendation in the PRIOR guideline.
2. For each extracted change, the topic must be **identical** between the prior and current recommendations (i.e., a strict head-to-head comparison).
    - For example, the prior recommendation and the current recommendation are different treatments based on the **SAME disease**
3. Focus only on recommendations that have changed substantially, for example:
    - The prior guideline recommends therapy A for disease C, whereas the current guideline recommends therapy B for the **SAME disease** C based on new clinical evidence. 
    - Each pair of `current_recommendation` and `prior_recommendation` must represent a direct **head-to-head comparison** and be **meaningfully different**.
4. **MUST exclude minor or incremental updates**, including but not limited to:
    - The prior guideline lacked specific guidance, and the current guideline merely adds clarification or detail.
    - Both guidelines recommend the same intervention, with the current guideline only expanding on context, rationale, or implementation details.
5. If the prior and current guidelines address different clinical topics, return an empty list.
6. Each generated `current_recommendation` and `prior_recommendation` must be self-contained, precise, and clinically interpretable, explicitly stating any necessary background conditions or patient populations. Do not use vague words and try to be more specific and detailed.
7. Your recommendation text in `current_recommendation` and `prior_recommendation` MUST be faithful to the corresponding raw texts. DO NOT fabricate any new clinical terms or conditions.
8. Identify as many differences as possible.
\end{minted}
\captionof{listing}{Prompt for extracting differences from the guideline trajectory.}
\label{prompt:extract_difference}

\begin{minted}[breaklines, fontsize=\small, frame=lines]{text}
### Assignment

You are evaluating whether the **prior** and **current** clinical guideline recommendations shown below exhibit a **strict head-to-head change**.

Your task is to determine whether the current recommendation truly replaces, updates, or meaningfully modifies the prior recommendation **for the same explicit clinical decision unit**, and assign a score accordingly.

---

## SCORING

- GOOD = 1
- OK = 0.5
- BAD = 0

---

## STRICT HEAD-TO-HEAD REQUIREMENTS

### 1. SAME EXPLICIT CLINICAL DECISION UNIT (MANDATORY)

The two recommendations must address the **same explicitly stated clinical unit**, including:

- the same disease or condition (as explicitly named)
- the same patient population
- the same disease stage / severity / control status
- the same clinical decision point  
  (e.g., first-line therapy, step-up therapy, elective surgery after non-operative management)

IMPORTANT:
- Clinical reasoning, medical ontology, or pathophysiologic hierarchy MUST NOT be used to infer equivalence.
- The clinical decision unit in the two recommendations must be explicitly the same. Wording may differ slightly ONLY if the two recommendations refer to the same explicitly stated disease, population, and decision point, but they must refer to the same decision unit as named in the text.
- **Overlap, subset, or superset relationships do NOT qualify as the same unit**.
- If the wording in the prior and the current recommendation are **NOT the exactly the SAME** and you think they are refering to the SAME clinical decision unit, YOU MUST CLEARLY EXPLAIN WHY in your `rationale` with a starting hashtag <reason>!!!

A broader condition in the prior guideline CANNOT be treated as implicitly covering a more specific subtype or complication introduced only in the current guideline.

Example (NOT head-to-head):
- prior: “complicated diverticulitis”
- current: “pelvic abscess treated with percutaneous drainage”

Even if the subtype is clinically part of the broader category, this is considered a **decision-unit mismatch**.

---

### 2. POPULATION ALIGNMENT (TEXT-EXPLICIT, NOT INFERRED)

- Population descriptions **MUST not differ in wording**, otherwise should be rejected (BAD: 0.0).
- Any stratification used in the **current** recommendation (e.g., uncontrolled disease, refractory cases, abscess subtype, treatment pathway, eligibility subgroup) MUST also be explicitly present in the prior recommendation.
- You may NOT assume that the prior recommendation applies to a subgroup unless it is explicitly mentioned.

#### Allowed Exception: Refinement of Enumerated Categories

If the prior recommendation **explicitly enumerates multiple categories**, procedures, or settings (e.g., THA, TKA, HFS; mild/moderate/severe disease), the current recommendation MAY provide more specific guidance for those same enumerated items.

This is considered a valid refinement ONLY if no new category, subtype, or decision axis is introduced.

---

### 3. REFINEMENT VS. FRAMEWORK CHANGE

Refinement is allowed ONLY along axes that **already exist** in the prior guideline.

- Valid refinement:
  - prior explicitly lists A, B, C
  - current provides more specific guidance for A and/or B

- Invalid refinement (framework change):
  - current introduces new disease subtypes, complications, treatment pathways, or eligibility gates
  - these elements were NOT explicitly defined in the prior recommendation
  - This type of refinement should be **directly rejected** (Bad: 0.0)

Framework changes are NOT head-to-head comparisons, even if clinically reasonable.

---

### 4. TRUE CHANGE IN GUIDANCE

To qualify as a meaningful change:

- The current recommendation must replace, restrict, expand, or clearly supersede the prior guidance for the same decision unit.

The following do NOT qualify as substantive change:
- increased certainty without strategy change
- stronger wording without new clinical action
- background explanation or rationale
- administrative or non-clinical updates

---

### 5. CLARITY REQUIREMENT

Both the prior and current recommendations must be clear and specific enough to support comparison.

Vague language (e.g., “controversial”, “case-by-case”, “may be considered” without context) weakens the comparison.

---

## SCORING GUIDELINES

### GOOD (1.0)

Assign GOOD if and only if ALL of the following are met:

- same explicit clinical decision unit
- no population or stratification mismatch
- a clear, substantive replacement or modification of guidance
- both recommendations are clear and specific

Typical pattern:
- prior A → current B (A is explicitly replaced, restricted, or superseded)

---

### OK (0.5)

Assign OK if ALL of the following are met:

- same explicit clinical decision unit
- no new disease subtype or decision axis introduced
- BUT:
  - the change is modest, optional, or asymmetric
  - OR the current guideline largely affirms the prior recommendation and adds limited caveats or special situations
  - OR one of the recommendations lacks sufficient specificity

Typical pattern:
- prior A remains valid
- current adds special cases, optional adjustments, or clarifications

---

### BAD (0.0)

Assign BAD if ANY of the following apply:

- different disease, population, stage, or decision unit
- new stratification, subtype, or decision axis introduced only in the current guideline
- refinement occurs along an axis not present in the prior guideline
- no meaningful change in guidance
- one or both recommendations are too vague to compare

---

## IMPORTANT NOTE

- Be conservative and annotation-faithful.
- Do NOT reward clinical plausibility, inferred hierarchy, or medical reasoning beyond what is explicitly stated in the text.
- Only explicit, text-level alignment qualifies as a head-to-head comparison.

## IMPORTANT CLARIFICATION ON SUBSTANTIVE CHANGE:

- A change that only removes or alters relative preference, recommendation strength, or wording (e.g., “may be preferred” vs “no stated preference”), WITHOUT introducing or excluding any clinical action, SHOULD NOT be considered a substantive change.

Such changes should be scored as OK (0.5), not GOOD (1).
\end{minted}
\captionof{listing}{Prompt for verifying extracted differences from the guideline trajectory.}
\label{prompt:verifier}

\begin{minted}[breaklines, fontsize=\small, frame=lines]{text}

You are an expert medical exam writer trained in NBME item-writing guidelines.
You are generating evaluation questions about CHANGES between clinical guidelines.

==============================
TASK
==============================
Create ONE high-quality multiple-choice question that evaluates knowledge of a CHANGE between two clinical guidelines.

Answer option requirements:
- Choice A MUST represent the CURRENT guideline recommendation.
- Choice B MUST represent the PRIOR guideline recommendation.
- Choices C and D must represent medically plausible alternatives that are NOT the current or prior recommendation.

Difficulty requirements:
- The question must be HARD and CLEAR.
- It must require understanding what changed between guidelines.
- It must describe a realistic clinical scenario.
- The correct answer should not be obvious without knowing the guideline update.

Guideline citation requirement:
- The question must explicitly cite the guideline source and year.
Example: 'According to the Society for Immunotherapy of Cancer guideline issued in 2023...'

Important restriction:
- Do NOT mention the words 'current' or 'prior' in the question text.
- Those labels are only used internally for answer construction.

==============================
NBME ITEM WRITING RULES
==============================
1. Test APPLICATION of knowledge, not recall of isolated facts.

2. Use a clinical vignette structured as:
   - patient demographics
   - chief complaint
   - relevant history
   - physical examination
   - laboratory or imaging findings

3. Write a CLOSED and FOCUSED lead-in question.

Acceptable lead-ins include:
- Which of the following is the most appropriate next step in management?
- Which of the following is recommended according to the guideline?
- Which of the following interventions should be initiated?

Avoid vague lead-ins such as:
- Which statement is true?
- What is associated with this condition?

4. Create EXACTLY FOUR answer options (A–D).

Rules for options:
- Only ONE option is the best answer.
- All options must be HOMOGENEOUS (same category such as treatments, tests, or diagnoses).
- Distractors must be medically plausible.
- Options must have similar length and grammatical structure.

5. Avoid technical flaws:
- Do NOT use 'all of the above' or 'none of the above'.
- Avoid vague terms such as 'often', 'usually', 'frequently'.
- Avoid absolute terms such as 'always' or 'never'.
- Avoid grammatical cues that reveal the correct answer.
- Avoid repeating distinctive words from the vignette in only one option.

6. Apply the COVER-THE-OPTIONS rule:
A knowledgeable reader should be able to infer the correct answer after reading the vignette and lead-in before seeing the options.

7. Ensure the question tests clinical reasoning and guideline application.

Important:
Do not encode the guideline change into the patient scenario unless the guideline explicitly conditions the recommendation on those features.
\end{minted}
\captionof{listing}{Prompt for generating NBME-style QA questions from the extracted difference.}
\label{prompt:qa_generation}

\section{Examples}
\subsection{Examples of the Extracted Guideline Trajectory and Differences}\label{appendix:extracted_differences_example}
Each difference file is formulated as a dictionary object. The field `clinical\_focus' provides a short description of the clinical topic or decision point under comparison. The fields `current\_recommendation' and `prior\_recommendation' contain normalized summaries of the recommendation in the current and prior guidelines, respectively, while `current\_recommendation\_raw\_text' and `prior\_recommendation\_raw\_text' preserve the original supporting passages from the source documents. The field `change\_summary' summarizes how the recommendation evolved across guideline versions. Finally, `score' records the score from the verifier LLM (0=Not Good, 0.5=Okay but not good, and 1=Good), and `rationale' provides a justification for why the paired recommendations are considered a valid and meaningful change instance. This format allows both structured downstream processing and direct inspection of the underlying textual evidence.

\subsubsection{PMC12583794 vs. PMC10088549}
Here we show an example of the extracted differences from the paper PMC12583794 (2024 KSSO Guideline) and the paper PMC10088549 (2022 KSSO Guideline). 

\begin{minted}[breaklines, fontsize=\small, frame=lines]{text}
"clinical_focus": "Diagnostic criteria for adult obesity (requirement for confirmation of excess adiposity beyond BMI and evaluation for 'clinical obesity')",
"current_recommendation": "To diagnose obesity, confirm excess adiposity using at least one anthropometric criterion (e.g., waist circumference) or direct measurement of body fat in addition to BMI; thereafter, evaluate for clinical obesity—excess adiposity with organ dysfunction, metabolic abnormalities, or functional impairments—regardless of BMI.",
"current_recommendation_raw_text": "The diagnosis of obesity requires the confirmation of excess adiposity using at least one anthropometric criterion (e.g., waist circumference) or direct measurement of body fat, in addition to BMI. … Individuals with confirmed obesity should subsequently be evaluated for clinical obesity, which is defined as excess adiposity accompanied by organ dysfunction, metabolic abnormalities, or functional impairments, regardless of BMI.",
"prior_recommendation": "In adults, diagnosis of obesity is based on BMI; the adult criterion for obesity in Korea is BMI >=25 kg/m2.",
"prior_recommendation_raw_text": "Diagnostic criteria for obesity, abdominal obesity, and metabolic syndrome remain the same as in the previous version (Table 2). Diagnosis of obesity based on body mass index (BMI) applies regardless of age in adults. … R1-1-2. Considering the risk of obesity-related comorbidities, the criterion for adult obesity in Korea is a BMI of 25 kg/m^2 or higher (IIa, B).",
"change_summary": "The current guideline moves from a BMI-only diagnostic approach to a two-step process requiring confirmation of excess adiposity in addition to BMI and the subsequent assessment of 'clinical obesity' independent of BMI; the prior guideline relied on BMI thresholds alone (>=25 kg/m2) to diagnose adult obesity.",
"score": 1.0,
"rationale": "#reason Both texts address the same explicit clinical decision unit—diagnostic criteria for adult obesity within the same guideline context. The prior specifies adult obesity diagnosis based solely on BMI (criterion BMI >=25 kg/m² in Korea). The current explicitly changes this by requiring confirmation of excess adiposity with an anthropometric or direct body-fat measure in addition to BMI, thereby replacing the BMI-only approach. This is a clear, substantive modification of the same decision unit. The added step to evaluate for “clinical obesity” is a subsequent assessment and does not alter the core unit being compared (diagnosis criteria), which is directly updated."
\end{minted}

\subsubsection{PMC12855235 vs. PMC8986677}
Here we show an example of the extracted differences from the paper PMC12855235 (2025 SEOM-GEMCAD-TTD Guideline) and the paper PMC8986677 (2021 SEOM-GEMCAD-TTD Guideline). 

\begin{minted}[breaklines, fontsize=\small, frame=lines]{text}
"clinical_focus": "Use of watch-and-wait (non-operative management) after a well-documented clinical complete response (cCR) following neoadjuvant therapy in localized rectal cancer",
"current_recommendation": "For individuals with a well-documented clinical complete response after neoadjuvant therapy, a watch-and-wait strategy should be presented as a therapeutic option because oncologic outcomes are comparable to surgery.",
"current_recommendation_raw_text": "Based on current evidence, for individuals with well-documented cCR, W\u00026W yields comparable oncological outcomes to surgery and should be presented as a therapeutic option",
"prior_recommendation": "Non-operative management after neoadjuvant therapy could not be formally recommended; watch-and-wait should be reserved for prospective clinical trials and select individuals after multidisciplinary evaluation.",
"prior_recommendation_raw_text": "Till now, no evidence from randomized trials is available to confirm both the long-term oncological outcomes and the superiority of organ preservation in terms of QoL. Currently, a recommendation for non-operative management after neoadjuvant treatment cannot formally be proposed and the W\u00026W strategy may be reserved for prospective clinical trials and individually selected patients after a multidisciplinary evaluation of response.",
"change_summary": "The 2025 guideline endorses watch-and-wait as a valid option for well-documented cCR, whereas the 2021 guideline advised restricting watch-and-wait to clinical trials or highly selected cases without a formal recommendation.",
"score": 1.0,
"rationale": "#reason Both recommendations address the same explicit clinical decision unit: watch-and-wait (non-operative management) for localized rectal cancer patients with a well-documented clinical complete response after neoadjuvant therapy. The prior stance restricted W&W to prospective trials and individually selected patients after multidisciplinary evaluation (effectively not formally recommending it), whereas the current recommendation advises that W&W should be presented as a therapeutic option given comparable oncologic outcomes to surgery. This is a clear, substantive change from restricted/trial-only use to broader offering within the same population and decision point."
\end{minted}

\subsection{Examples of the Constructed MCQs}\label{appendix:mcqs}
\begin{minted}[breaklines, fontsize=\small, frame=lines]{text}
A radiation oncology department is replacing paper-based preprocedure safety checklists used for external beam radiotherapy and high–dose-rate (HDR) brachytherapy. The team wants better multidisciplinary communication and faster audits of checklist use. They also plan to auto-populate treatment parameters from the oncology information system and trigger out-of-tolerance alerts for HDR source strength and independent MU checks. The physicist leading the project notes that adding automation will change failure modes and that checklist fatigue has been a problem with static lists. According to the American Association of Physicists in Medicine guideline issued in 2023, which of the following implementation strategies is most appropriate?

A. Implement an EMR-integrated electronic checklist with interlocks, formal documentation, and rapid audit capability; design from a sociotechnical/human–computer interaction perspective and deploy intelligent dynamic items with targeted automation (auto-populated fields, out-of-tolerance alerts) validated by formal risk analysis and maintained through routine updates.

B. Implement a basic electronic checklist primarily to enable interlocks and quick audits; pilot test to identify pitfalls, avoid deep EMR integration to prevent tethering during time-critical procedures, and consider portable devices to improve usability.

C. Retain paper checklists supplemented by verbal time-outs; perform quarterly retrospective audits, and avoid interlocks or automation to preserve workflow flexibility during time-sensitive cases.

D. Deploy a fully automated hard-stop system that auto-populates all items and removes human verification; omit formal risk analysis and ongoing updates because automation will minimize residual error risk.

========================
Note:
(Ground-truth: Option A)
(Outdated guideline: Option B)
(Distractor Options: Option C and D)
\end{minted}

\section{Experimental Setups}

\subsection{QA Evaluation}
\textbf{Models.} For all the experiments, we use 10 open-source models, including \texttt{medgemma-4b-it}~\cite{sellergren2025medgemma},  \texttt{Llama-3.1-8B-Instruct}~\cite{grattafiori2024llama}, \texttt{Llama-3.2-3B-Instruct}~\cite{grattafiori2024llama}, \texttt{Qwen3-4B-Instruct}~\cite{yang2025qwen3}, \texttt{gpt-oss-20b}~\cite{agarwal2025gpt}, \texttt{Olmo-3-7B-Instruct}~\cite{olmo2025olmo}, \texttt{Qwen-2.5-14B-Instruct}~\cite{qwen2025qwen25technicalreport}, \texttt{Qwen2.5-7B}~\cite{qwen2025qwen25technicalreport}, \texttt{MedGemma-1.5-4B-IT}~\cite{sellergren2026medgemma}, \texttt{Gemma-3-4B-IT}~\cite{team2025kimi}, and 3 closed-source models, including \texttt{GPT-4o}, \texttt{GPT-4.1}, and \texttt{GPT-5}.

\textbf{Hyper-parameters.} Except for \texttt{GPT-5}, we use a temperature of 0 across all the experiments. All the experiments are conducted on a server with an A100 GPU.

\subsection{Positional Perturbation}\label{appendix:positional_perturbation}
We applied positional perturbations to each 5-option MCQ at the evaluation stage. Starting from the original item, we generated perturbed variants under three schemes. 
\begin{itemize}[leftmargin=*]
    \item \textit{Reorder-only} permutes the displayed option order while preserving the original label--content bindings
    \item \textit{Label Shuffling} keeps the displayed slots fixed but permutes the option contents across positions, thereby changing the location of the correct, outdated, and distractor options.
    \item \textit{Alternate Labels} preserves the option contents but replaces the canonical option labels with an alternative label set (e.g., X, Y, Z, W).
\end{itemize}

\subsection{Agentic Search Tools}\label{appendix:agentic_search_tools}
\paragraph{ToolUniverse} ToolUniverse~\cite{gao2025democratizing} is a platform that contains a comprehensive list of agentic tools in the biomedical domain, such as PubMedSearch, a tool enabling the LLM agent to search with the PubMed dataset\footnote{\url{https://pubmed.ncbi.nlm.nih.gov/}}; FDASearch, a tool enabling the LLM agent to search with the OpenFDA\footnote{\url{https://open.fda.gov/}}. Following recommendations by the ToolUniverse official document, we integrate the ToolUniverse into a single LLM (e.g., \texttt{GPT-4.1} and \texttt{GPT-5}) by augmenting the context messages with the retrieved information with the tools\footnote{\url{https://zitniklab.hms.harvard.edu/ToolUniverse/guide/chatgpt\_api.html}}.

\paragraph{Biomni} Biomni~\cite{huang2025biomni} is a general-purpose biomedical AI agent designed to autonomously execute a wide range of research tasks across diverse biomedical subfields. We also use \texttt{GPT-4.1} and \texttt{GPT-5} as the default backend LLMs.

\section{Examples of Question Instances}
\subsection{Examples of the Historical-knowledge-targeted-questions}\label{appendix:example_outdated_questions}

Here, we present a historical-knowledge-targeted question together with its corresponding up-to-date-targeted question.

\textbf{Up-to-date-targeted Questions (correct answer in green color)}

\begin{minted}[breaklines, fontsize=\small, frame=lines, escapeinside=||]{text}
A 45-year-old Korean woman presents for an annual health examination. She reports gaining 6 kg over the last 3 years. She is 160 cm tall and weighs 62 kg (BMI 24.2 kg/m2). Blood pressure is 148/92 mm Hg. Waist circumference is 87 cm. Laboratory studies show HbA1c 7.6%, triglycerides 260 mg/dL, HDL cholesterol 39 mg/dL, ALT 72 U/L; thyroid-stimulating hormone is normal. Abdominal ultrasonography demonstrates hepatic steatosis without fibrosis. She asks whether she meets criteria for obesity.\n\nAccording to the Korean Society for the Study of Obesity guideline |\textcolor{red}{issued in 2024}|, which of the following is the most appropriate diagnostic approach for this patient?

|\textcolor{green}{A. Use the elevated waist circumference to confirm excess adiposity in addition to BMI and diagnose clinical obesity because of diabetes and hepatic steatosis.}|

B. Refrain from diagnosing obesity because BMI is <25 kg/m2; classify her as non-obese and address comorbidities separately.

C. Diagnose obesity based solely on waist circumference, disregarding BMI and comorbidities.

D. Obtain dual-energy X-ray absorptiometry to measure body fat percentage and defer any obesity diagnosis until BMI reaches >=25 kg/m2

E. I do not know the answer.
\end{minted}

\textbf{Corresponding Historical-knowledge-targeted Questions (correct answer in green color)}

\begin{minted}[breaklines, fontsize=\small, frame=lines, escapeinside=||]{text}
A 45-year-old Korean woman presents for an annual health examination. She reports gaining 6 kg over the last 3 years. She is 160 cm tall and weighs 62 kg (BMI 24.2 kg/m2). Blood pressure is 148/92 mm Hg. Waist circumference is 87 cm. Laboratory studies show HbA1c 7.6%, triglycerides 260 mg/dL, HDL cholesterol 39 mg/dL, ALT 72 U/L; thyroid-stimulating hormone is normal. Abdominal ultrasonography demonstrates hepatic steatosis without fibrosis. She asks whether she meets criteria for obesity.\n\nAccording to the Korean Society for the Study of Obesity guideline |\textcolor{red}{issued in 2022}|, which of the following is the most appropriate diagnostic approach for this patient?

A. Use the elevated waist circumference to confirm excess adiposity in addition to BMI and diagnose clinical obesity because of diabetes and hepatic steatosis.

|\textcolor{green}{B. Refrain from diagnosing obesity because BMI is <25 kg/m2; classify her as non-obese and address comorbidities separately.}|

C. Diagnose obesity based solely on waist circumference, disregarding BMI and comorbidities.

D. Obtain dual-energy X-ray absorptiometry to measure body fat percentage and defer any obesity diagnosis until BMI reaches >=25 kg/m2

E. I do not know the answer.
\end{minted}

\subsection{Examples of the Questions for RQ3}

\begin{minted}[breaklines, fontsize=\small, frame=lines]{text}
According to the most recent guideline on Lung Cancer Screening published by the China lung cancer early detection and treatment expert group (CLCEDTEG) on or before 2000, do you agree that LDCT lung cancer screening should be routinely offered to adults aged 50–74 years with at least 10 years of domestic coal use in Xunwei City, even if they do not meet smoking pack-year criteria?

A. Yes
B. No
C. I do not know
\end{minted}
\captionof{listing}{Example of the Questions for RQ3}
\label{listing:example_rq3}

\begin{figure*}[!h]
      \centering
      \begin{subfigure}[t]{0.32\textwidth}
          \centering
          \includegraphics[width=\linewidth]{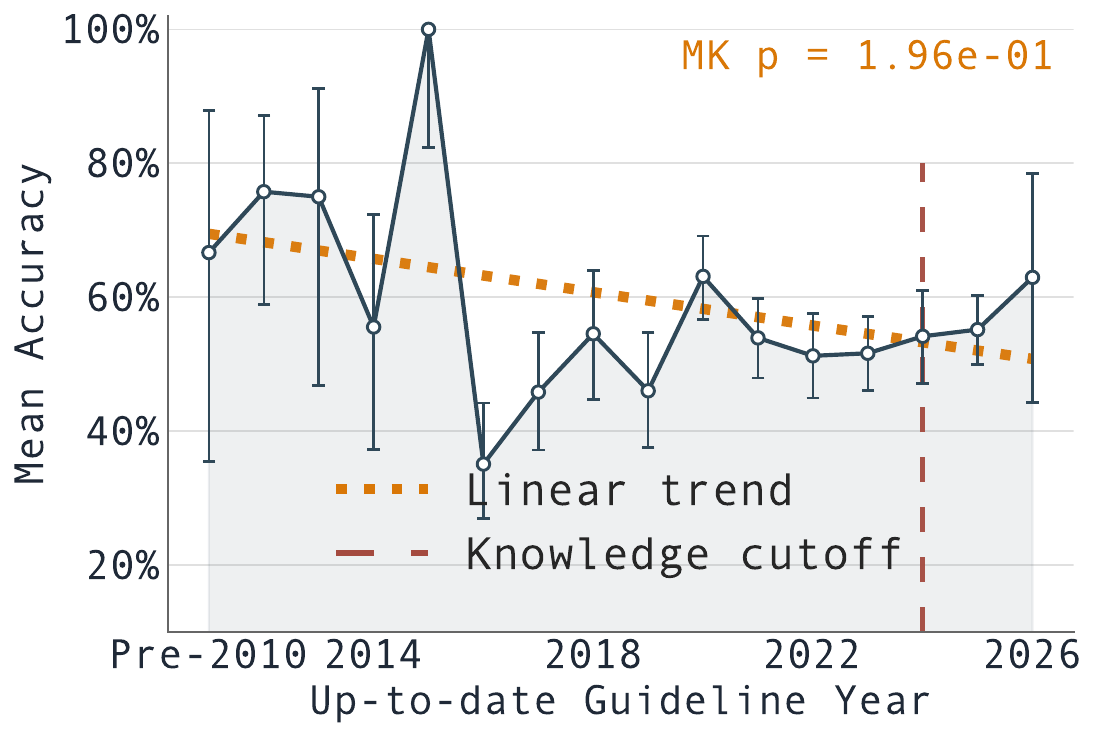}
          \caption{\texttt{Llama-3.2-3B}}
      \end{subfigure}
      \hfill
      \begin{subfigure}[t]{0.32\textwidth}
          \centering
          \includegraphics[width=\linewidth]{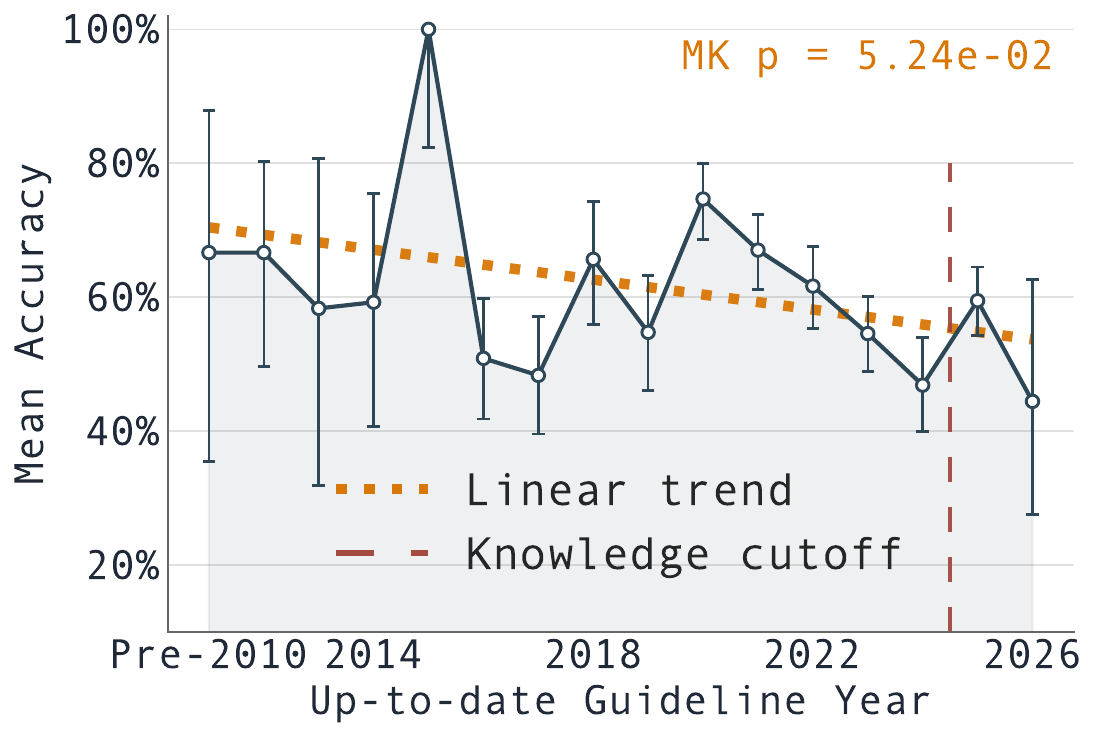}
          \caption{\texttt{Qwen-2.5-7B}}
      \end{subfigure}
      \hfill
      \begin{subfigure}[t]{0.32\textwidth}
          \centering
          \includegraphics[width=\linewidth]{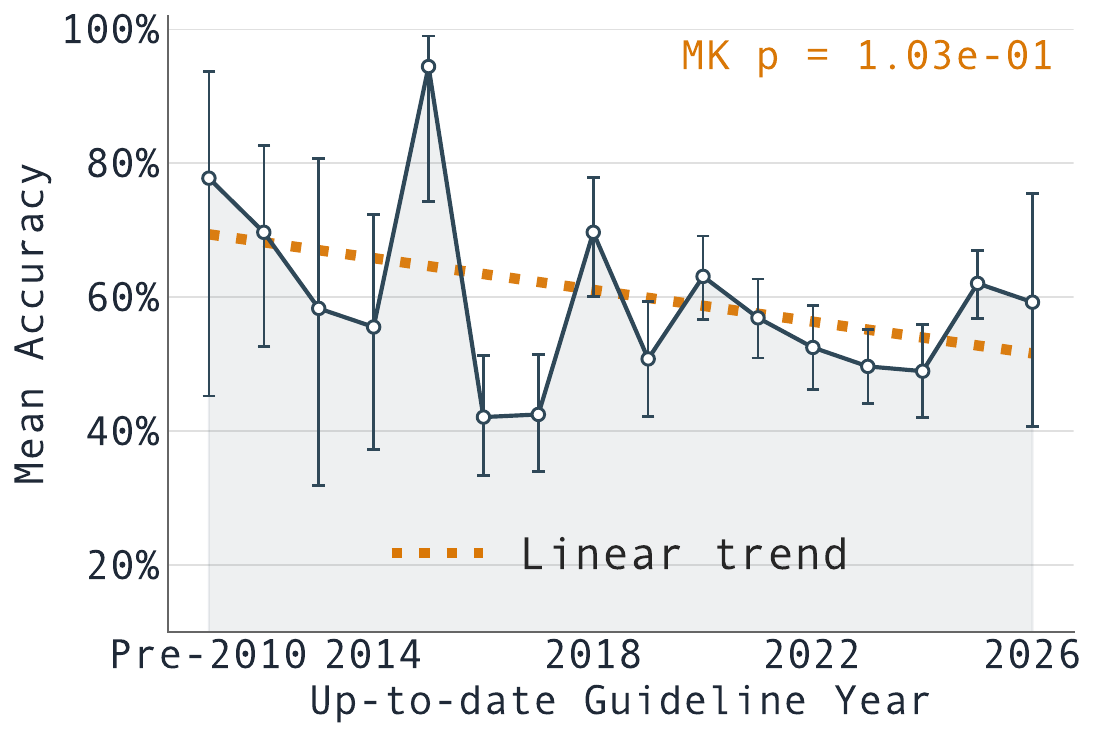}
          \caption{\texttt{Medgemma-1.5-4B-IT}}
      \end{subfigure}
      \hfill
      \begin{subfigure}[t]{0.32\textwidth}
          \centering
          \includegraphics[width=\linewidth]{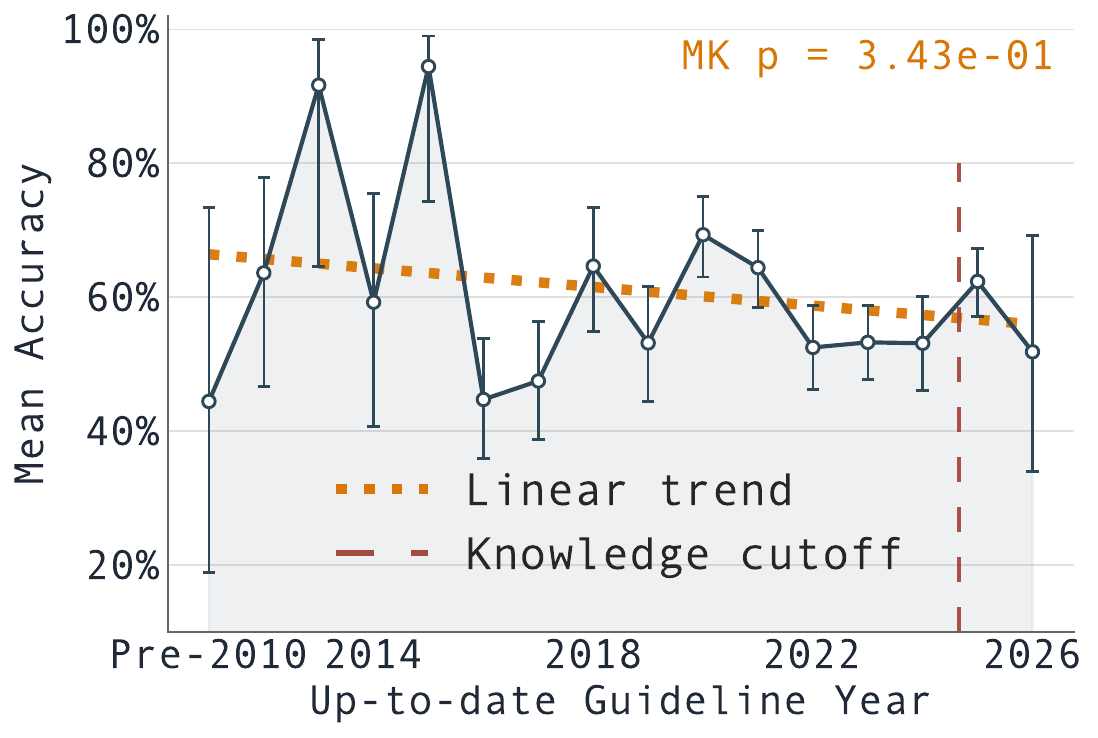}
          \caption{\texttt{Gemma-3-4B}}
      \end{subfigure}
      \caption{More results on average accuracy by the year of the up-to-date guideline.}
      \label{fig:more_results_current_accuracy}
\end{figure*}

\begin{figure*}[t]
      \centering
      \begin{subfigure}[t]{0.23\textwidth}
          \centering
          \includegraphics[width=\linewidth]{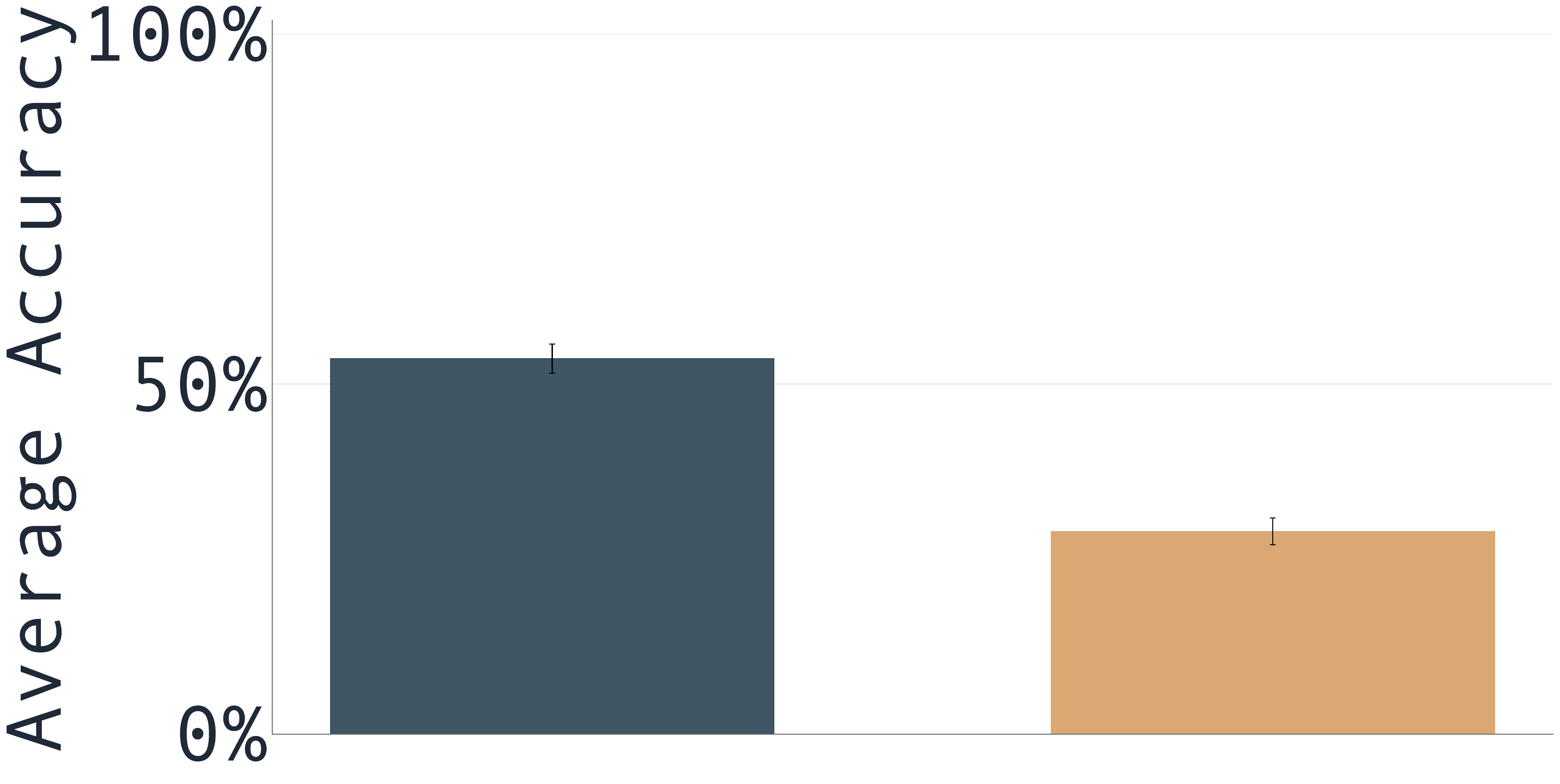}
          \caption{\texttt{Llama-3.2-3B}}
      \end{subfigure}
      \hfill
      \begin{subfigure}[t]{0.23\textwidth}
          \centering
          \includegraphics[width=\linewidth]{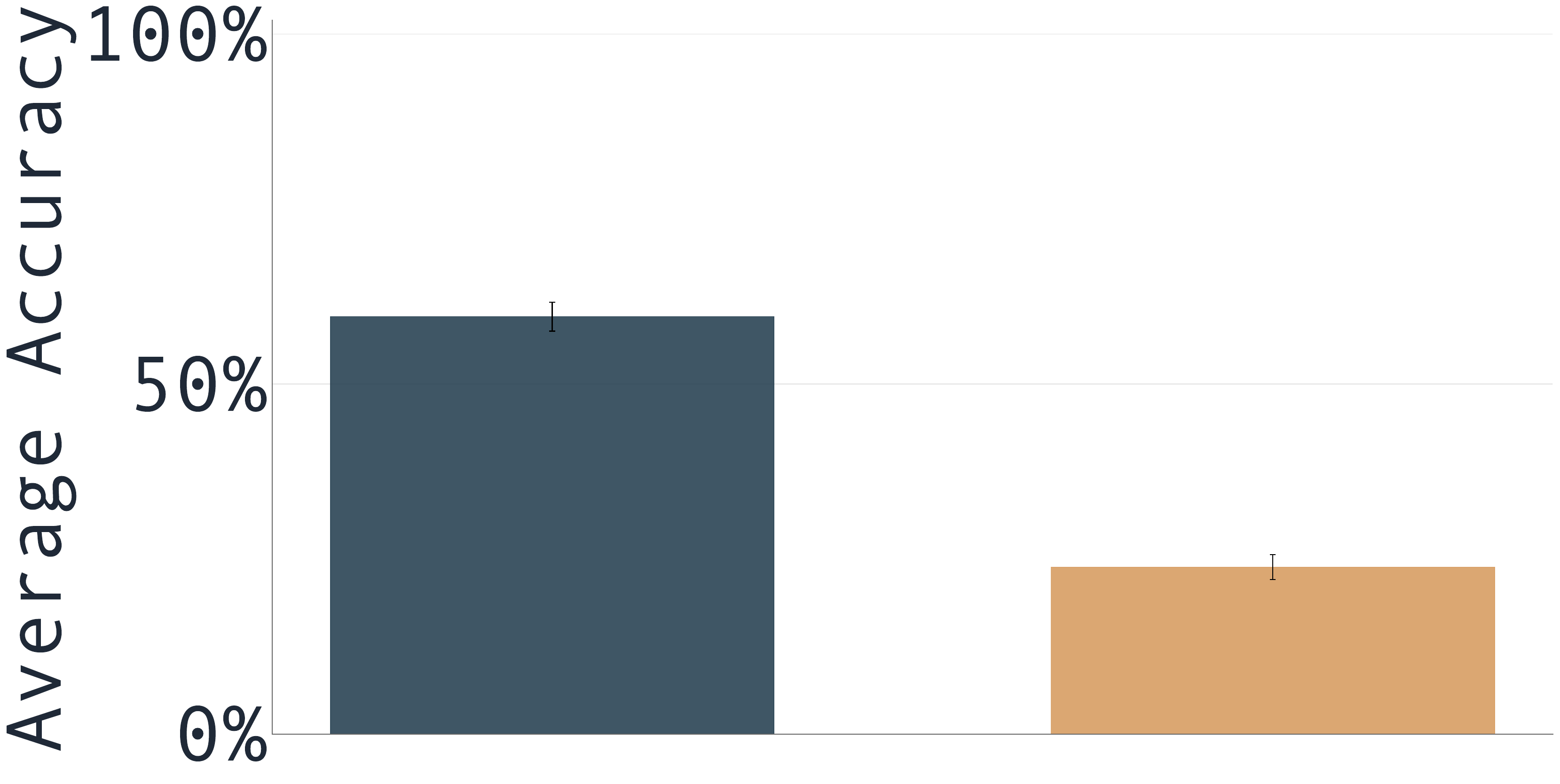}
          \caption{\texttt{Qwen-2.5-7B}}
      \end{subfigure}
      \hfill
      \begin{subfigure}[t]{0.23\textwidth}
          \centering
          \includegraphics[width=\linewidth]{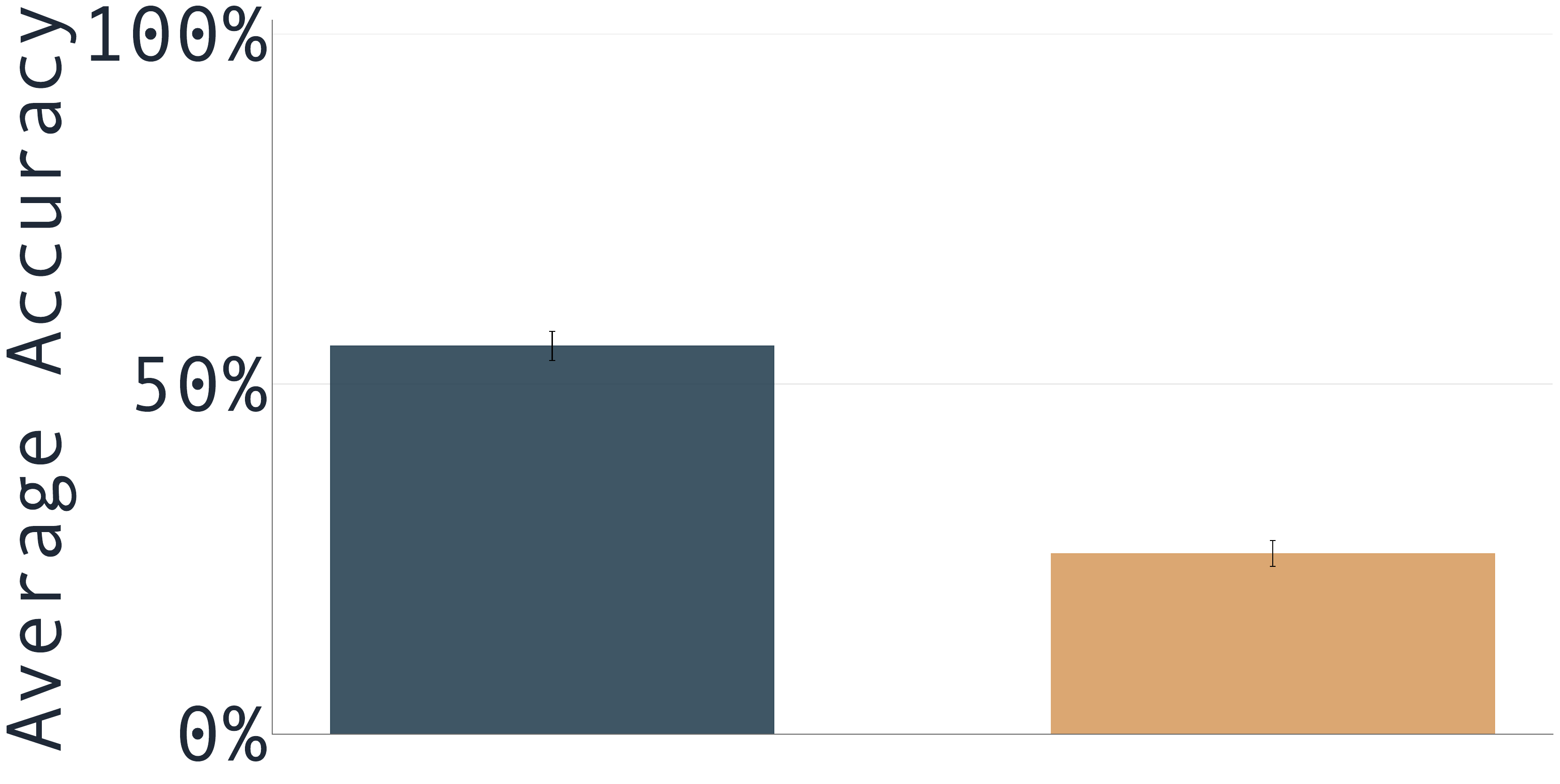}
          \caption{\texttt{Medgemma-1.5-4B}}
      \end{subfigure}
      \hfill
      \begin{subfigure}[t]{0.23\textwidth}
          \centering
          \includegraphics[width=\linewidth]{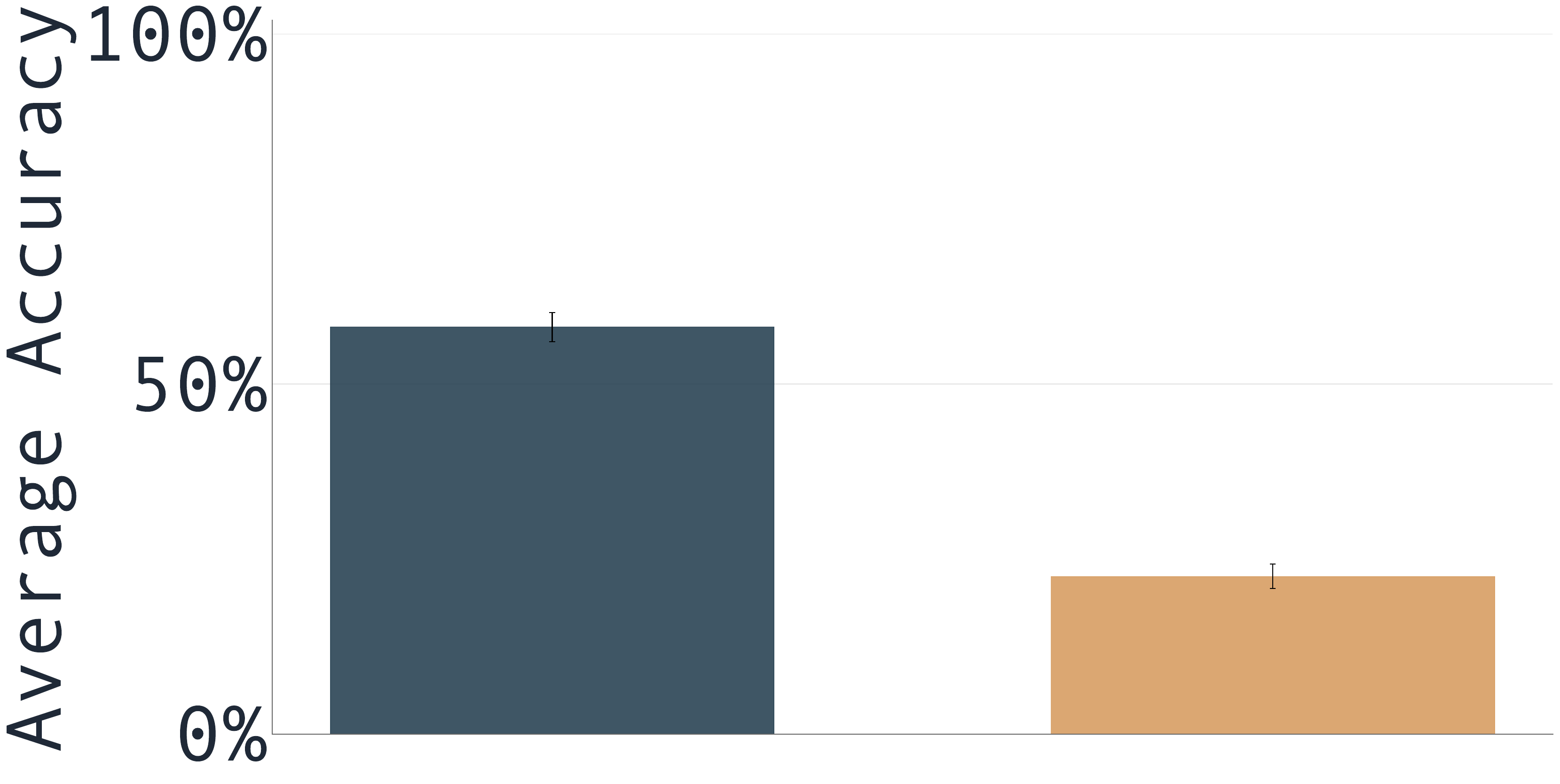}
          \caption{\texttt{Gemma-3-4B}}
      \end{subfigure}
    \caption{More results on accuracy comparison between up-to-date-targeted and historical-knowledge-targeted questions.}
    \label{fig:more_results_targeted_accuracy}
\end{figure*}

\subsection{Example of the Model Answers}
\subsubsection{Examples of the Temporal Consistency Patterns in RQ3} \label{appendix:error_pattern}

\begin{figure}[!t]
    \centering
    \includegraphics[width=1.5\linewidth, angle=-90]{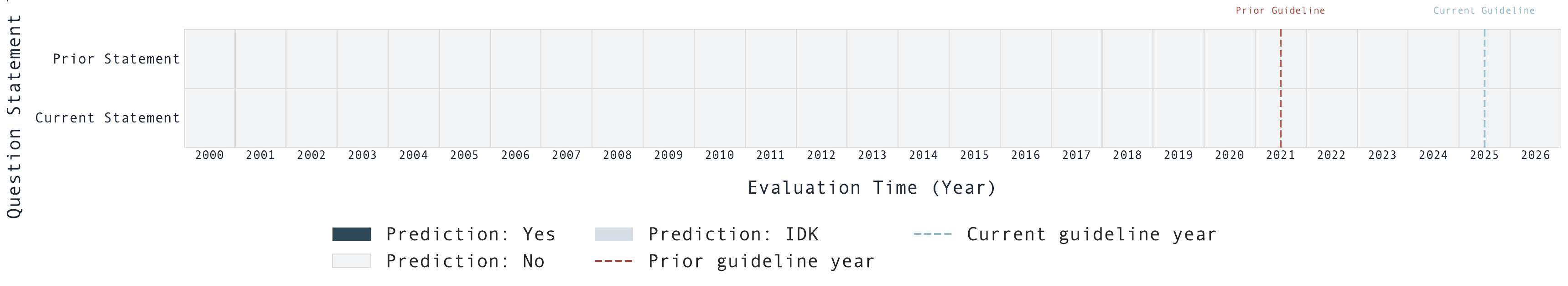}
    \caption{\textit{All-No} Pattern.}
    \label{fig:All-No}
\end{figure}

\begin{figure}[!t]
    \centering
    \includegraphics[width=1.5\linewidth, angle=-90]{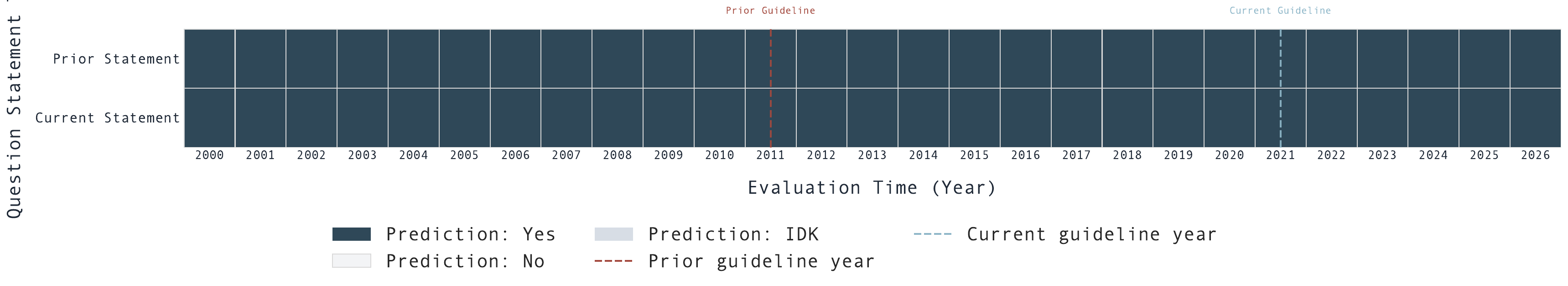}
    \caption{\textit{All-Yes} Pattern.}
    \label{fig:All-Yes}
\end{figure}

\begin{figure}[!t]
    \centering
    \includegraphics[width=1.5\textwidth, angle=-90]{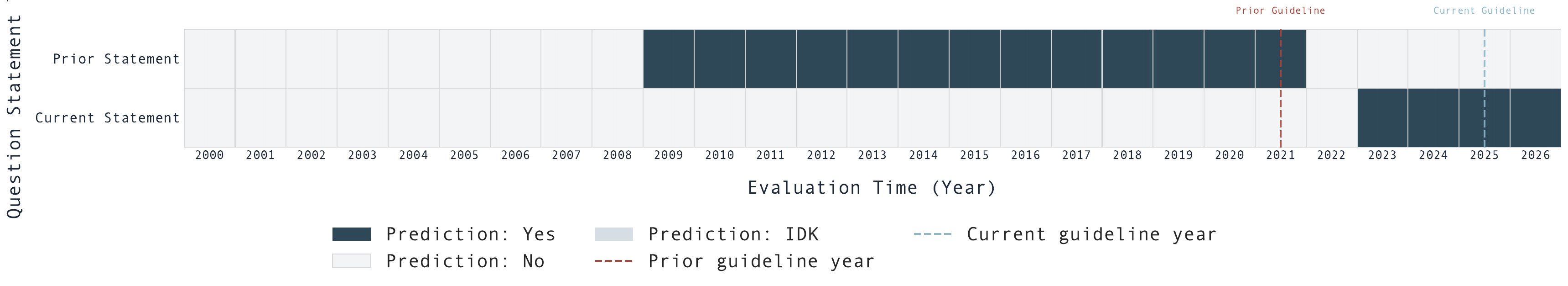}
    \caption{\textit{Wrong-Transition-Point} Pattern.}
    \label{fig:Wrong-Transition-Point}
\end{figure}

The following figures (Figure~\ref{fig:Only-Know-Current}, Figure~\ref{fig:All-No}, Figure~\ref{fig:All-Yes}, Figure~\ref{fig:Wrong-Transition-Point}) show the heatmaps for different temporal consistency patterns.

\begin{figure}[!t]
    \centering
    \includegraphics[width=1.5\linewidth, angle=-90]{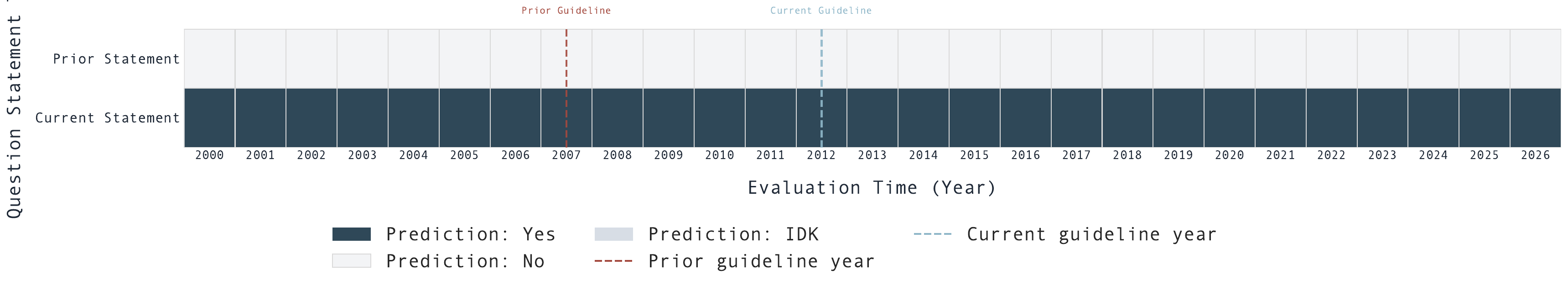}
    \caption{\textit{Only-Know-Current} Pattern.}
    \label{fig:Only-Know-Current}
\end{figure}

\clearpage
\subsubsection{Examples of Tool Trace After Using the ToolTrace}\label{appendix:example_tooluniverse_model_answer}
\begin{minted}[breaklines, fontsize=\tiny, frame=lines]{text}
"tool_trace": [
  {
    "name": "PubMed_Guidelines_Search",
    "arguments": {
      "query": "GIST post-adjuvant surveillance SEOM GEIS guideline 2022",
      "limit": 5
    },
    "result_preview": "[]"
  },
  {
    "name": "web_search",
    "arguments": {
      "query": "SEOM GEIS 2022 guideline GIST post-adjuvant surveillance recommendations",
      "max_results": 10
    },
    "result_preview": "{'status': 'success', 'data': {'status': 'success', 'query': 'SEOM GEIS 2022 guideline GIST post-adjuvant surveillance recommendations', 'search_type': 'general', 'total_results': 9, 'results': [{'title': 'PDF SEOM-GEIS clinical guideline for gastrointestinal stromal tumors (2022)', 'url': 'https://www.grupogeis.org/documentos/guias-geis/Pub-Int+_Area-GIST_Guias-GIST-SEOM-GEIS_2023.pdf', 'snippet': 'These guidelines are elaborated by the conjoint eort of the Spanish Society of Medical Oncology (SEOM) and the Spanish Sarcoma Research Group (GEIS) and provide a multidisciplinary and updated consensus for the diagnosis and treatment of GIST patients.', 'rank': 1}, {'title': 'SEOM-GEIS clinical guideline for gastrointestinal stromal tumors (2022)', 'url': 'https://pubmed.ncbi.nlm.nih.gov/37129716/', 'snippet': 'These guidelines are elaborated by the conjoint effort of the Spanish Society of Medical Oncology (SEOM) and the Spanish Sarcoma Research Group (GEIS) and provide a multidisciplinary and updated consensus for the diagnosis and treatment of GIST patients.', 'rank': 2}, {'title': 'SEOM-GEIS clinical guideline for gastrointestinal stromal tumors (2022)', 'url': 'https://www.researchgate.net/publication/370470414_SEOM-GEIS_clinical_guideline_for_gastrointestinal_stromal_tumors_2022', 'snippet': 'These guidelines are elaborated by the conjoint effort of the Spanish Society of Medical Oncology (SEOM) and the Spanish Sarcoma Research Group (GEIS) and provide a multidisciplinary and updated ...', 'rank': 3}, {'title': 'GEIS guidelines - GEIS Group', 'url': 'https://grupogeis.org/en/scientific-activity/guides-and-nomograms/geis-guidelines', 'snippet': 'GEIS Guidelines of Uncommon Tumors (2021) - Part 2 GEIS-SEHOP Guidelines for the Treatment of Rhabdomyosarcoma (2021) GEIS Osteosarcoma Guidelines (2017) SEOM Clinical Guideline of management of soft-tissue sarcoma (2016) SEAP-SEOM Guidelines for pathological and molecular diagnosis in GIST (2016) 2015 STS Guidelines Adult Osteosarcoma (2011)', 'rank': 4}, {'title': 'Guías Clínicas SEOM - SEOM: Sociedad Española de Oncología Médica © 2019', 'url': 'https://seom.org/publicaciones/guias-clinicas/105418-guias-clinicas-seom', 'snippet': 'SEOM edita anualmente Guías Clínicas realizadas en colaboración con los Grupos Cooperativos Nacionales de investigación en cáncer, Secciones SEOM y otras Sociedades Científicas.', 'rank': 5}, {'title': 'Guía clínica SEOM-GEIS para los tumores del estroma gastrointestinal (2022)', 'url': 'https://colectivogist.wordpress.com/2023/05/04/guia-clinica-seom-geis-para-los-tumores-del-estroma-gastrointestinal-2022/comment-page-1/', 'snippet': 'traducción: colectivogist Resumen El tumor del estroma gastrointestinal (GIST) es la neoplasia maligna de origen mesenquimal más frecuente y un modelo paradigmático para el éxito del desarrollo racional de terapias dirigidas contra el cáncer. La introducción de inhibidores de la tirosina quinasa con actividad contra KIT/PDGFRA tanto en estadios localizados como avanzados ha mejorado ...', 'rank': 6}, {'title': '2023 GEIS Guidelines for gastrointestinal stromal tumors', 'url': 'https://journals.sagepub.com/doi/10.1177/17588359231192388', 'snippet': 'Gastrointestinal stromal tumor (GIST) is the most common malignant neoplasm of mesenchymal origin. GIST spans a wide clinical spectrum that ranges from tumors with essentially no metastatic potential to malignant and life-threatening spread diseases. Gain-of-function mutations in KIT or PDGFRA receptor tyrosine kinases are the crucial drivers of most GISTs, responsible for tumor initiation and ...', 'rank': 7}, {'title': 'ESMO Clinical Practice Guidelines: Sarcoma and GIST', 'url': 'https://www.esmo.org/guidelines/esmo-clinical-practice-guidelines-sarcoma-and-gist', 'snippet': 'Latest enhanced and revised set of guidelines The ESMO Clinical Practice Guidelines on Sarcoma and GIST are the result of a consensus conference held every two years which brings together a large group of European experts.', 'rank': 8}, {'title': 'SEOM-GEIS clinical guideline for gastrointestinal stromal tumors (2022)', 'url': 'https://scientiasalut.gencat.cat/handle/11351/10233?show=full&locale-attribute=en', 'snippet': 'This item appears in the following Collection (s) HVH - Articles científics [4476] VHIO - Articles científics [1250] Show simple item record', 'rank': 9}], 'backend_used': 'duckduckgo', 'attempted_backends': ['duckduckgo']}}"
  },
  {
    "name": "PMC_search_papers",
    "arguments": {
      "query": "SEOM GEIS 2022 guideline GIST post-adjuvant surveillance",
      "limit": 5,
      "include_abstract": true
    },
    "result_preview": "[{'title': 'Understanding Gastric GIST: From Pathophysiology to Personalized Treatment.', 'abstract': 'Background: Gastric gastrointestinal stromal tumors (GISTs) represent a subset of gastrointestinal tumors predominantly found in the stomach. Despite their rarity, these tumors carry significant implications for patient health and management. GISTs are potentially malignant tumors with unpredictable progression. They originate from the interstitial cells of Cajal, which are positioned between the intramural neurons and the smooth muscle cells of the digestive tract. These tumors are characterized primarily by mutations in the c-Kit gene, as well as other mutations such as those in the platelet-derived growth factor receptor alpha (PDGFRA) gene. Methods: Our comprehensive search across five databases initially yielded 2976 articles. After eliminating 197 duplicates, we screened the titles and abstracts of 2779 articles, excluding 2692 for not meeting the inclusion criteria. During the full-text screening, 16 more articles were excluded. Ultimately, 71 papers met the inclusion criteria and were included in our analysis. Results: Due to differences in study designs, inclusion criteria for patients, and reported outcomes, a meta-analysis was not conducted. The accurate diagnosis of GIST is established through histopathological examination and immunohistochemistry. Histopathologically, GISTs are classified into three main types: spindle cell, epithelioid, and mixed. The therapeutic management of GIST involves surgery, endoscopic treatment, and chemotherapy. Conclusions: The prognosis for GIST patients depends on various factors, including risk category, disease stage, applied treatments, and recurrence post-treatment. A significant recent advancement comes from artificial intelligence, which can be increasingly involved in both the diagnosis and treatment of this tumor.', 'authors': ['Moga DF', 'Vlădoiu G', 'Frățilă AM', 'Dan AA', 'Popa D', 'Oprea V'], 'year': '2024', 'pmc_id': 'PMC11277334', 'pmid': '39064037', 'doi': '10.3390/jcm13143997', 'doi_url': 'https://doi.org/10.3390/jcm13143997', 'url': 'https://www.ncbi.nlm.nih.gov/pmc/articles/PMC11277334/', 'pdf_url': 'https://www.ncbi.nlm.nih.gov/pmc/articles/PMC11277334/pdf/', 'venue': 'J Clin Med', 'open_access': True, 'source': 'PMC', 'article_type': None, 'citations': 0, 'data_quality': {'has_authors': True, 'has_abstract': True, 'has_year': True, 'has_pmid': True, 'has_doi': True, 'has_url': True, 'has_venue': True}, 'abstract_source': 'PubMed'}]"
  },
  {
    "name": "PubMed_get_article",
    "arguments": {
      "pmid": "37129716"
    },
    "result_preview": "{'status': 'success', 'data': {'pmid': '37129716', 'title': 'SEOM-GEIS clinical guideline for gastrointestinal stromal tumors (2022).', 'abstract': 'Gastrointestinal stromal tumor (GIST) is the most common malignant neoplasm of mesenchymal origin, and a paradigmatic model for a successful rational development of targeted therapies in cancer. The introduction of tyrosine kinase inhibitors with activity against KIT/PDGFRA in both localized and advanced stages has remarkably improved the survival in a disease formerly deemed resistant to all systemic therapies. These guidelines are elaborated by the conjoint effort of the Spanish Society of Medical Oncology (SEOM) and the Spanish Sarcoma Research Group (GEIS) and provide a multidisciplinary and updated consensus for the diagnosis and treatment of GIST patients. We strongly encourage that the managing of these patients should be performed within multidisciplinary teams in reference centers.', 'authors': [{'name': 'Serrano César', 'affiliation': \"Sarcoma Translational Research Group, Vall d'Hebron Institute of Oncology (VHIO), Hospital Universitario Vall d'Hebron, Vall d'Hebron Barcelona Hospital Campus, C/Natzaret, 115-117, 08035, Barcelona, Spain. cserrano@vhio.net.\"}, {'name': 'Álvarez Rosa', 'affiliation': 'Hospital General Universitario Gregorio Marañón, Madrid, Spain.'}, {'name': 'Carrasco Juan Antonio', 'affiliation': 'Hospital Álvaro Cunqueiro-Complejo Hospitalario Universitario de Vigo, Pontevedra, Spain.'}, {'name': 'Marquina Gloria', 'affiliation': 'Hospital Universitario Clínico San Carlos, Madrid, Spain.'}, {'name': 'Martínez-García Jerónimo', 'affiliation': 'Hospital Universitario Virgen de la Arrixaca, Murcia, Spain.'}, {'name': 'Martínez-Marín Virginia', 'affiliation': 'Hospital Universitario la Paz, Madrid, Spain.'}, {'name': 'Sala María Ángeles', 'affiliation': 'Hospital Universitario Basurto, Bilbao, Spain.'}, {'name': 'Sebio Ana', 'affiliation': 'Hospital de la Santa Creu I Sant Pau, Barcelona, Spain.'}, {'name': 'Sevilla Isabel', 'affiliation': 'Hospitales Universitarios Regional y Virgen de la Victoria, Málaga, Spain.'}, {'name': 'Martín-Broto Javier', 'affiliation': 'Hospital Universitario Fundación Jiménez Díaz, Madrid, Spain.'}], 'journal': 'Clinical & translational oncology : official publication of the Federation of Spanish Oncology Societies and of the National Cancer Institute of Mexico', 'pub_date': '2023 Sep', 'pub_year': '2023', 'doi': '10.1007/s12094-023-03177-7', 'doi_url': 'https://doi.org/10.1007/s12094-023-03177-7', 'url': 'https://pubmed.ncbi.nlm.nih.gov/37129716/', 'mesh_terms': ['Humans', 'Gastrointestinal Stromal Tumors', 'Medical Oncology', 'Sarcoma', 'Consensus', 'Receptor Protein-Tyrosine Kinases'], 'publication_types': ['Journal Article', 'Practice Guideline']}, 'url': 'https://eutils.ncbi.nlm.nih.gov/entrez/eutils/efetch.fcgi?db=pubmed&retmode=xml&id=37129716'}"
  },
  {
    "name": "web_search",
    "arguments": {
      "query": "SEOM GEIS 2022 guideline GIST surveillance after adjuvant imatinib",
      "max_results": 5
    },
    "result_preview": "{'status': 'success', 'data': {'status': 'success', 'query': 'SEOM GEIS 2022 guideline GIST surveillance after adjuvant imatinib', 'search_type': 'general', 'total_results': 5, 'results': [{'title': 'PDF SEOM-GEIS clinical guideline for gastrointestinal stromal tumors (2022)', 'url': 'https://www.grupogeis.org/documentos/guias-geis/Pub-Int+_Area-GIST_Guias-GIST-SEOM-GEIS_2023.pdf', 'snippet': 'Imatinib is also the standard treatment for those patients with completely resected metastatic disease, and in GIST patients that develop metastatic disease after the na- lization of adjuvant imatinib.', 'rank': 1}, {'title': 'SEOM-GEIS clinical guideline for gastrointestinal stromal tumors (2022)', 'url': 'https://pubmed.ncbi.nlm.nih.gov/37129716/', 'snippet': 'These guidelines are elaborated by the conjoint effort of the Spanish Society of Medical Oncology (SEOM) and the Spanish Sarcoma Research Group (GEIS) and provide a multidisciplinary and updated consensus for the diagnosis and treatment of GIST patients.', 'rank': 2}, {'title': 'SEOM-GEIS clinical guideline for gastrointestinal stromal tumors (2022)', 'url': 'https://link.springer.com/article/10.1007/s12094-023-03177-7', 'snippet': 'Imatinib is also the standard treatment for those patients with completely resected metastatic disease, and in GIST patients that develop metastatic disease after the finalization of adjuvant imatinib.', 'rank': 3}, {'title': 'SEOM-GEIS clinical guideline for gastrointestinal stromal tumors (2022)', 'url': 'https://www.researchgate.net/publication/370470414_SEOM-GEIS_clinical_guideline_for_gastrointestinal_stromal_tumors_2022', 'snippet': 'Gastrointestinal stromal tumor (GIST) is the most common malignant neoplasm of mesenchymal origin, and a paradigmatic model for a successful rational development of targeted therapies in cancer.', 'rank': 4}, {'title': 'SEOM-GEIS clinical guideline for gastrointestinal stromal tumors (2022)', 'url': 'https://www.academia.edu/105576798/SEOM_GEIS_clinical_guideline_for_gastrointestinal_stromal_tumors_2022_', 'snippet': 'Poveda A, Martinez V, Serrano C, Sevilla I, Lecumberri MJ, de After stopping adjuvant imatinib, a closest follow-up is Beveridge RD, et al. SEOM clinical guideline for gastrointestinal necessary for the following 2 years, when the risk of recur- sarcomas (GIST) (2016).', 'rank': 5}], 'backend_used': 'duckduckgo', 'attempted_backends': ['duckduckgo']}}"
  },
  {
    "name": "web_search",
    "arguments": {
      "query": "site:grupogeis.org GIST surveillance after adjuvant imatinib SEOM GEIS 2022",
      "max_results": 3
    },
    "result_preview": "{'status': 'success', 'data': {'status': 'success', 'query': 'site:grupogeis.org GIST surveillance after adjuvant imatinib SEOM GEIS 2022', 'search_type': 'general', 'total_results': 3, 'results': [{'title': 'SEOM-GEIS clinical guideline for gastrointestinal stromal ...', 'url': 'https://www.grupogeis.org/documentos/guias-geis/Pub-Int+_Area-GIST_Guias-GIST-SEOM-GEIS_2023.pdf', 'snippet': 'These guidelines are elaborated by the conjoint eort of the Spanish Society of Medical Oncology (SEOM) and the Spanish Sarcoma Research Group (GEIS) and provide a multidisciplinary and updated consensus for the diagnosis and treatment of GIST patients.', 'rank': 1}, {'title': '2023 GEIS Guidelines for gastrointestinal stromal tumors', 'url': 'https://grupogeis.org/documentos/guias-geis/2023-GEIS-guidelines-for-GIST.pdf', 'snippet': 'Adjuvant treatment with imatinib 400 mg daily is indicated in GIST patients at high risk of relapse-bearing imatinib-sensitive mutations in KIT or PDGFRA (I,A).', 'rank': 2}, {'title': 'GEIS guidelines for gastrointestinal sarcomas (GIST)', 'url': 'https://grupogeis.org/Guías_GIST_2017.pdf', 'snippet': 'Physicians are encouraged to enroll GIST patients in clinical trials after progression on all current standard treatments for advanced/metastatic disease (imatinib, sunitinib, and regorafenib).', 'rank': 3}], 'backend_used': 'auto', 'attempted_backends': ['duckduckgo', 'auto']}}"
  }
]
\end{minted}

\section{Broader impacts}\label{appendix:broader_impacts}
Our evaluation on TempoMed-Bench shows that even state-of-the-art LLMs lack temporal awareness in medical knowledge, and this limitation persists even when integrating agentic retrieval-augmented generation (RAG) systems. These findings suggest that current LLMs are still far from being reliable assistants for time-specific medical decision-making.

It should be emphasized that strong performance on TempoMed-Bench alone does not guarantee the safe deployment of LLMs in clinical settings. However, failure on this benchmark indicates a critical limitation in temporal medical reasoning that must be addressed before such systems can be trusted in real-world healthcare applications. In this sense, we believe that passing TempoMed-Bench should be considered a necessary—although not sufficient—condition for deploying LLMs in evolving medical environments.

\end{document}